\documentclass[10pt,twocolumn,letterpaper]{article}

\usepackage{cvpr}              

\usepackage{graphicx}
\usepackage{amsmath}
\usepackage{amssymb}
\usepackage{booktabs}
\usepackage{fontawesome}
\usepackage{caption}
\usepackage{appendix}

\usepackage[pagebackref,breaklinks,colorlinks]{hyperref}

\usepackage[capitalize]{cleveref}
\crefname{section}{Sec.}{Secs.}
\Crefname{section}{Section}{Sections}
\Crefname{table}{Table}{Tables}
\crefname{table}{Tab.}{Tabs.}

\def\cvprPaperID{8508} 
\def\confName{CVPR}
\def\confYear{2023}

\begin{document}

\title{3DFill:Reference-guided Image Inpainting by Self-supervised 3D Image Alignment}

\author{Liang Zhao\textsuperscript{1} \and
Xinyuan Zhao\textsuperscript{2} \and
Hailong Ma\textsuperscript{1} \and
Xinyu Zhang\textsuperscript{1} \and
Long Zeng\textsuperscript{1} \and
\textsuperscript{1}Tsinghua University \and
\textsuperscript{2}Huawei Technologies \and
{\tt\small \{zhao-l20,mhl20,zhangxy20\}@mails.tsinghua.edu.cn} \and
{\tt\small zhaoxinyuan1@huawei.com} \and
{\tt\small zenglong@sz.tsinghua.edu.cn}
}


\twocolumn[{
\renewcommand\twocolumn[1][]{#1}
\maketitle
\begin{center}
\captionsetup{type=figure}
    \centering
    \begin{subfigure}{0.19\linewidth}
        \includegraphics[width=1\linewidth]{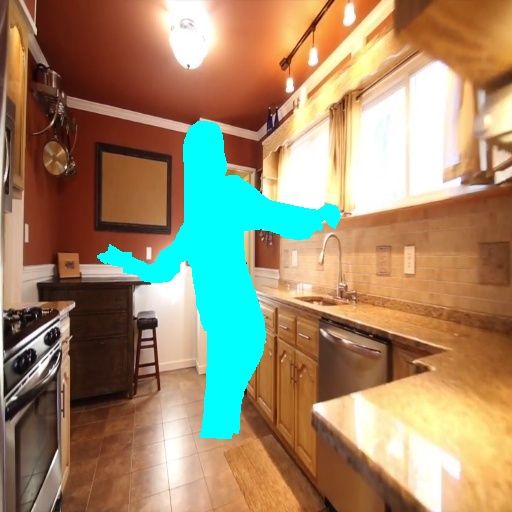}
    \end{subfigure}
    \begin{subfigure}{0.19\linewidth}
        \includegraphics[width=1\linewidth]{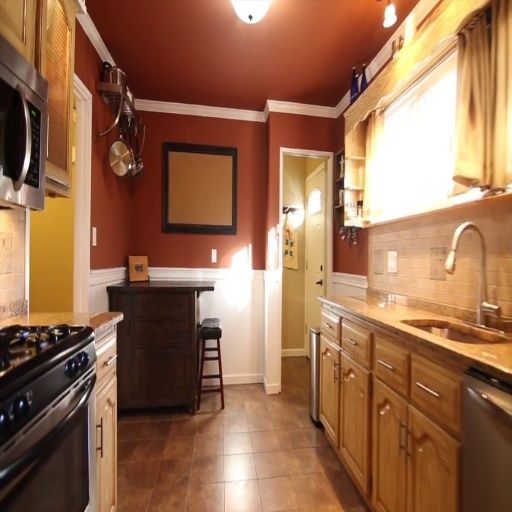}
    \end{subfigure}
    \begin{subfigure}{0.19\linewidth}
        \includegraphics[width=1\linewidth]{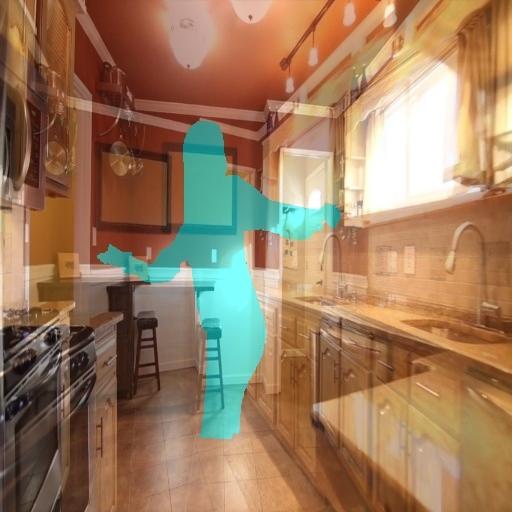}
    \end{subfigure}
    \begin{subfigure}{0.19\linewidth}
        \includegraphics[width=1\linewidth]{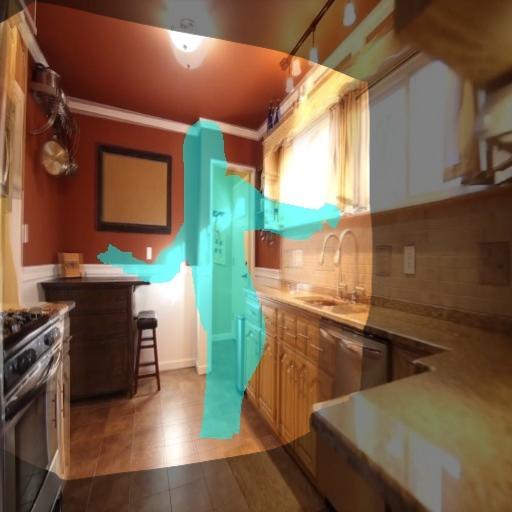}
    \end{subfigure}
    \begin{subfigure}{0.19\linewidth}
        \includegraphics[width=1\linewidth]{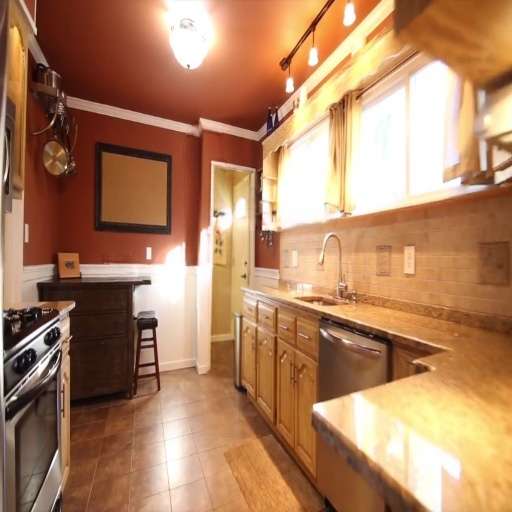}
    \end{subfigure}

    \centering
    \begin{subfigure}{0.19\linewidth}
        \includegraphics[width=1\linewidth]{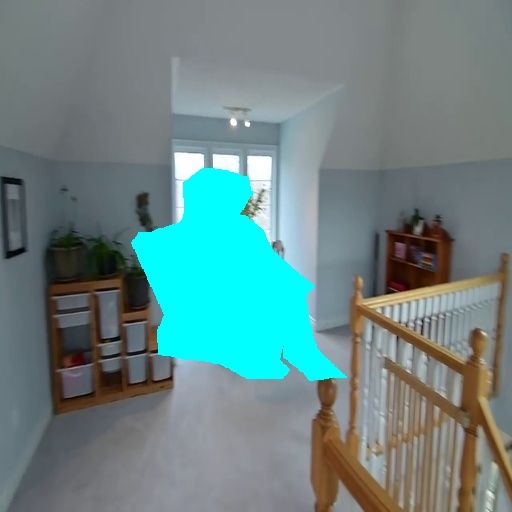}
        \caption{Target}
    \end{subfigure}
    \begin{subfigure}{0.19\linewidth}
        \includegraphics[width=1\linewidth]{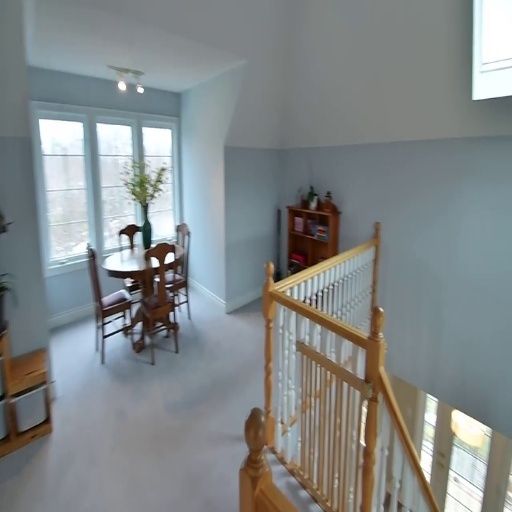}
        \caption{Reference}
    \end{subfigure}
    \begin{subfigure}{0.19\linewidth}
        \includegraphics[width=1\linewidth]{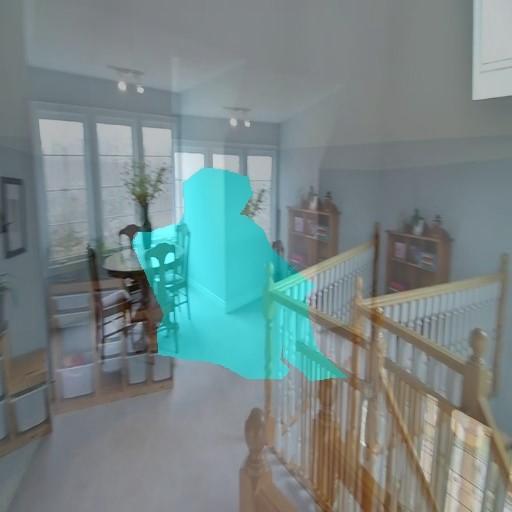}
        \caption{Before alignment}
    \end{subfigure}
    \begin{subfigure}{0.19\linewidth}
        \includegraphics[width=1\linewidth]{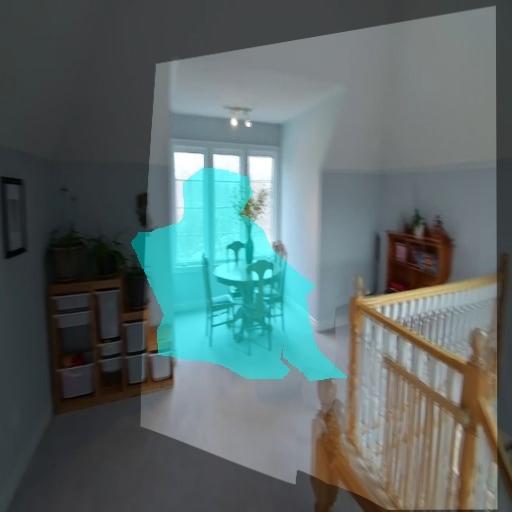}
        \caption{After alignment}
    \end{subfigure}
    \begin{subfigure}{0.19\linewidth}
        \includegraphics[width=1\linewidth]{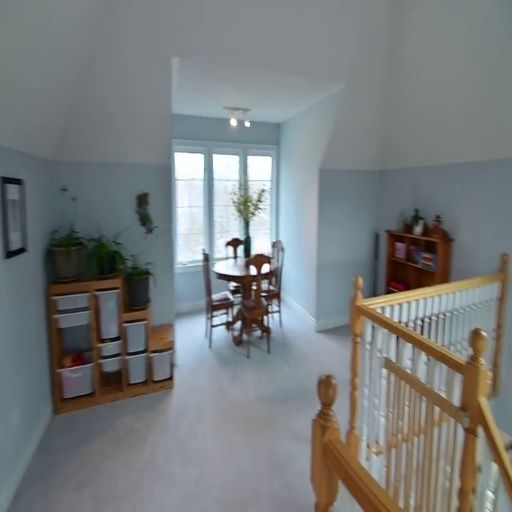}
        \caption{Result}
    \end{subfigure}
    \caption{Results of our reference-guided inpainting algorithm. Our method uses a reference image as prior information to inpaint the masked target image and can solve cases with large holes or complex scenes.}
    \label{fig:big examples}
\end{center}
}]

\begin{abstract}
   Most existing image inpainting algorithms are based on a single view, struggling with large holes or the holes containing complicated scenes.
   Some reference-guided algorithms fill the hole by referring to another viewpoint image and use 2D image alignment.
   Due to the camera imaging process, simple 2D transformation is difficult to achieve a satisfactory result.
   In this paper, we propose 3DFill, a simple and efficient method for reference-guided image inpainting. 
   Given a target image with arbitrary hole regions and a reference image from another viewpoint, the 3DFill first aligns the two images by a two-stage method: 3D projection + 2D transformation, which has better results than 2D image alignment.
   The 3D projection is an overall alignment between images and the 2D transformation is a local alignment focused on the hole region.
   The entire process of image alignment is self-supervised.
   We then fill the hole in the target image with the contents of the aligned image.
   Finally, we use a conditional generation network to refine the filled image to obtain the inpainting result.
   3DFill achieves state-of-the-art performance on image inpainting across a variety of wide view shifts and has a faster inference speed than other inpainting models.
\end{abstract}

\section{Introduction}
\label{sec:intro}

Image inpainting aims to restore missing pixels within a hole region in an image while making the restored image visually realistic.
The hole region could contain occluded objects or corrupted regions of the image.
Most of the existing algorithms focus on filling holes with the background information of a single image, which can be divided into the traditional algorithm and deep learning algorithm.
The traditional algorithm implements hole region filling based on pixel diffusion around the hole region \cite{diffusion_1} or similar pixel matching \cite{patchmatch_1}.
Deep learning algorithms based on neural network \cite{CNN_1, CNN_2_EdgeConnect, CNN_3, CNN_4} leverage knowledge learned from large-scale training data to fill in holes.
Although these algorithms have achieved outstanding inpainting results in some cases, due to the ill-posed nature \cite{ill-nature} of the task itself, it is difficult to solve the case with large hole or the hole region containing complex scenes.

The aforementioned difficulties can be addressed by reference-guided image inpainting \cite{OPN_reference-guided} which has a second image from another viewpoint of the same scene.
The second image which we called the \emph{reference image} exposes the content that can be used to fill in the hole of the image which we called the \emph{target image}.
Although the reference image is easily obtained by moving the camera to a different viewpoint to expose the background or collecting from the Internet \cite{image_from_Internet}, reference-guided image inpainting is less explored due to the challenge of image alignment between the target image and reference image.
Most of the existing reference-guided algorithms use 2D transforms, such as affine transformation and homography transformation \cite{align_2D_homography, TransFill}, to achieve image alignment.
Such methods simplify the problem of aligning images taken by cameras with a different viewpoint, resulting in poor effect.
Although the captured image is 2D, the camera imaging process is a 3D to 2D mapping, which prevents simple 2D transformations from achieving the desired alignment.

To address these issues, we propose 3DFill, an image alignment scheme based on two-stage alignment: 3D projection + 2D transformation, and use the aligned image to complete the task of image inpainting.
First, we use 3D projection to achieve the overall alignment of the two images.
We reverse the camera imaging process to recover the 3D scene corresponding to the 2D target image.
To implement it, we estimate the depth map of the target image and use the camera intrinsic parameter to calculate the corresponding 3D scene coordinates for each pixel.
After recovering the 3D scene, we estimate the 6-DOF transformation matrix between the target image and the reference one, which help us to align the 3D scene from the target viewpoint to the reference one.
We then use the forward process of camera imaging to project the 3D scene onto the 2D image at the reference viewpoint.
The coarse-aligned image can be got by the differentiable bilinear sampling method \cite{3D_projection}.
Second, to improve the alignment accuracy of the hole region, we do a further fine 2D alignment between the coarse-aligned image and the target image.
After getting the fine-aligned image, we fill the hole region of the target image with the contents of the fine-aligned image.
Finally, for a realistic and natural filling effect, we use a simple conditional generation network (CGAN) \cite{DCGAN} to refine the filled image.
In our image alignment training process, we do not need the label of the transformation matrix between the image pairs and only use the image similarity after alignment as the loss function. 
Therefore this process is self-supervised training.
The image alignment model and image refinement model of 3DFill are lightweight which makes the entire algorithm efficient and fast.

In summary, the main contributions of our method are:

\begin{itemize}
\item We propose 3DFill, a reference-guided image inpainting algorithm based on 3D image alignment.
\item We propose a simple, efficient and self-supervised module to complete the image alignment task, which achieves great alignment effect.
\item Experiments on the datasets indicate that our 3DFill is effective to generate realistic contents for image inpainting.
\end{itemize}

\begin{figure*}[t]
    \centering
    \includegraphics[width=1\linewidth]{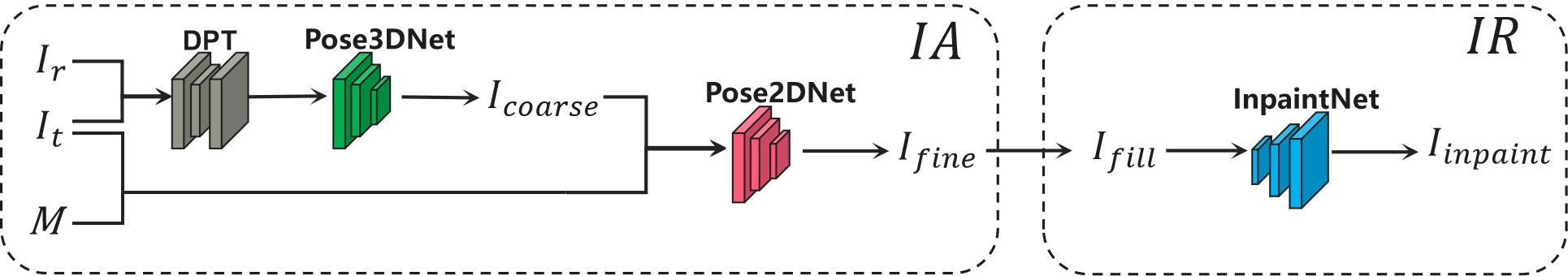}
    \caption{An overview of 3DFill. First, we use the image alignment (IA) module to align the input target image with the reference image. 
    The DPT \cite{DPT} is used to estimate the depth map of the input image to achieve 3D projection.
    The Pose3DNet and Pose2DNet are used to estimate the 3D and 2D pose transformations between the input images, respectively.
    We fill the contents of the aligned image into the hole of the target image, and then use InpaintNet to refine the image by image refinement (IR) module.}
    \label{fig:overview}
\end{figure*}
\section{Related Work}
\label{sec:related}

\textbf{Image inpainting}. 
Inpainting methods can be divided into two categories: traditional methods and deep learning methods.

\emph{Traditional methods:} Some traditional methods \cite{diffusion_1, diffusion_2, diffusion_3} use the technique of pixel diffusion to generate hole regions. 
However, when the hole area becomes larger, the restored effect of this kind of method will become blurred.
Other methods \cite{patchmatch_1, patchmatch_2, patchmatch_3} calculate the similarity of different regions and fill the hole regions with the most similar patches from non-hole regions.
Although these methods can obtain high-quality texture, they will make the structure and semantic information of the filled image unreasonable.

\emph{Deep learning methods:} Most of the deep learning methods use neural network to generate inpainted images.
They can be divided into those with and without prior information.
The input of the method without prior information \cite{noprior_PConv, noprior_PDGAN, noprior_PIC, noprior_LAMA} is only the image to be inpainted and the inpainting result is generated based on the complex model structure and large-scale training data.
When the area of the missing region becomes larger and larger, the method without any prior information will fall into ill-posed nature, which leads to the poor and unreasonable restoration effect.
To solve the problem of insufficient background information, more and more methods begin to leverage the knowledge learned from prior information, such as edges \cite{CNN_2_EdgeConnect}, low-frequency structures \cite{prior_lowstructure}, segmentation masks \cite{prior_segmentation}, and reference images \cite{TransFill, GeoFill}.
Reference image contains the most complete structure and the highest quality textures compared to other prior information.
Our method can be considered as a reference-guided image inpainting method.

\textbf{Image alignment}. 
Image alignment is the task of unifying different images into the same coordinate system.
Usually, the image alignment task first needs to estimate the transformation matrix of the two images, that is, image registration. 
Then warp one image to the another image through the transformation matrix.
Based on different image registration methods, image alignment algorithms can be divided into 2D algorithms and 3D algorithms.

\emph{2D:} Image 2D transformation can be divided into rigid body transformation, affine transformation, and homography transformation.
Traditional methods utilize hand-crafted descriptors \cite{2Dalignment_handcraft_1, 2Dalignment_handcraft_2, 2Dalignment_handcraft_3} or learned local features \cite{2Dalignment_learnedfeature_1, 2Dalignment_learnedfeature_2, 2Dalignment_learnedfeature_3} to estimate the homography matrix.
Deep learning methods directly estimate the relative pose by an end-to-end network, such as OANet \cite{2Dalignment_DL_1}.
According to the principle of camera imaging, images from different viewpoints involve the transformation of the 3D scene and the projection from 3D to 2D.
The transformation of the 3D scene includes 6 DOF, and the projection depends on the intrinsic parameters of the camera including 5 DOF, a total 11 DOF. 
A homography transformation only has 8 DOF which is difficult to achieve the desired image alignment.
TransFill \cite{TransFill} designs a multi-homography strategy to realize 2.5D transformation, however, it is still not a good substitute for 3D transformation.

\emph{3D:} The most classical 3D transformation algorithm is SfM \cite{SfM}, a machine learning algorithm.
Although SfM achieves outstanding performance, it is difficult to achieve a fast registration speed due to the complexity of its calculation.
Deep learning for directly predicting the 3D transformation matrix of two images is rarely studied \cite{GeoFill, MIC-GAN}.
MIC-GAN \cite{MIC-GAN} refers to a self-supervised depth estimation algorithm \cite{3D_projection} to align two images in 3D space, but the depth estimation model trained from scratch limits the accuracy of registration.

\textbf{Reference-guided image inpainting}. 
At present, there is not much research on image inpainting based on reference images.
Most of these \cite{referencguided_2D_1, align_2D_homography} use vanilla homography matrix to complete 2D image alignment before performing image inpainting tasks.
Still, others use SfM \cite{referencguided_SfM_1} or optimized homography \cite{TransFill, GeoFill} matrices to achieve image registration, they are based on various assumptions, such as plane-based \cite{referencguided_SfM_planebased}, and use sophisticated geometric algorithms to fit the image distortion caused by camera viewpoint change.
These methods are not only computationally complex but also difficult to obtain fully aligned images, especially when the camera viewpoint changes largely.
To achieve plausible image inpainting results in this situation, these methods often have to design a complex refining module, such as CST \cite{TransFill}, which further complicates the model.

\section{Method}
\label{sec:method}
\cref{fig:overview} shows an overview of 3DFill. 
We are given an original image $I_{o} \in \mathbb{R}^{W \times H \times 3}$, an associated mask $M \in \mathbb{R}^{W \times H \times 1}$, and a reference image $I_{r} \in \mathbb{R}^{W \times H \times 3}$.
Note that $M$ indicates the hole regions with value 0 and elsewhere with 1.
The original image does not contain the hole region, and the target image containing the hole region is obtained by $I_{t} = M \odot I_{o} $.
Our method is divided into two steps, which correspond to two modules, namely, the image alignment (IA) module and the image refinement (IR) module.
We first use the IA module, which has a two-stage alignment, to complete the image alignment task.
The 3D module uses $I_{t}$ and $I_{r}$ as inputs and outputs the coarse-aligned image $I_{coarse}$.
The 2D module uses $I_{t}$, $I_{coarse}$ and $M$ as inputs and outputs the fine-aligned image $I_{fine}$.
We then fill the contents of the aligned image to the hole area of the target image to obtain the $I_{fill}$.
After obtaining the $I_{fill}$, we use the IR module to implement the image refinement and output the inpainting result $I_{inpaint}$.

\subsection{Image Alignment Module}
\textbf{3D Module.}
To achieve the ideal alignment effect, our algorithm first uses the 3D module for coarse alignment and the following three data are required: intrinsic parameter of the camera, depth map corresponding to the input image, and 3D transformation matrix between the target image and the reference image.
In our 3D module, we use fixed camera intrinsic parameters instead of ground-truth camera intrinsic parameters because we find that our algorithm is robust to camera intrinsic parameters in \cref{sec:Robustness Study}.
We use DPT \cite{DPT}, a state-of-the-art depth estimation model, to help us obtain relatively accurate depth map.
This avoids the difficulty of self-supervised training depth estimation models, allowing us to focus more on the accuracy of 3D projection.
The 3D transformation matrix can be estimated by the Pose3DNet, which is a simple encoder model.

\begin{figure}[t]
    \centering
    \includegraphics[width=1\linewidth]{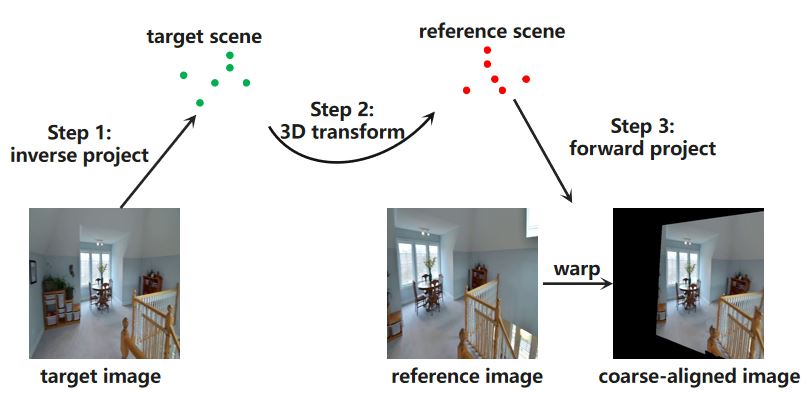}
    \caption{An example of 3D alignment (Green and red dots refer to the 3D points corresponding to each pixel on the image).}
    \label{fig:3D projection}
\end{figure}

\cref{fig:3D projection} shows the process of implementing 3D alignment.

Step 1: we reverse the projection process according to the camera imaging principle to obtain the 3D scene. This process achieves a 2D to 3D projection with the help of the depth map $Z_{t}$ and camera's intrinsic parameters $K$.

Step 2: we convert the scene coordinates from the target viewpoint to the reference viewpoint by 3D transformation.
3D transformation contains 6 degrees of freedom so we use Pose3DNet to predict a vector with 6 elements corresponding to 3 translation variables $t_{x}$, $t_{y}$, $t_{z}$ and 3 rotation variables $\alpha$, $\beta$, $\gamma$.
We use the above variables to obtain the transformation matrix $T_{t \rightarrow r}$.

Step 3: we use the forward process of camera imaging to project the 3D scene onto the 2D image at the reference viewpoint.

So far, we can get the coordinates $P_{r} (u_{r}, v_{r})$ of each pixel $ P_{t} (u_{t}, v_{t})$ on the target image projected on the reference image.
The above projection process can be simplified by:
\begin{equation}
  P_{r}
  =
  K T_{t \rightarrow r} Z_{t} K^{-1} P_{t}
  \label{eq:entire eq}
\end{equation}
$P_{r}$ is a decimal value and we use a differentiable bilinear sampling method \cite{3D_projection} to get the coordinate of each pixel in the coarse-aligned image.
Although the 3D transformation matrix we use is from the target view to the reference view, the actual effect is that we warp the reference image to the target view and get the coarse-aligned image $I_{coarse}$.
We use the $L1$ loss between $I_{coarse}$ and $I_{o}$ to train the Pose3DNet:
\begin{equation}
  \mathcal{L}_{3D}
  =
  \Vert M_{valid} \odot (I_{coarse} - I_{o}) \Vert_1
  \label{eq:3D loss}
\end{equation}
where $M_{valid}$ indicates the area of the valid projection point with value 1 and elsewhere with 0.
The areas where the target image and reference image do not overlap are replaced by black in $I_{coarse}$, as shown in \cref{fig:3D projection}.

\begin{figure}[t]
    \centering
    \begin{subfigure}{0.32\linewidth}
        \includegraphics[width=1\linewidth]{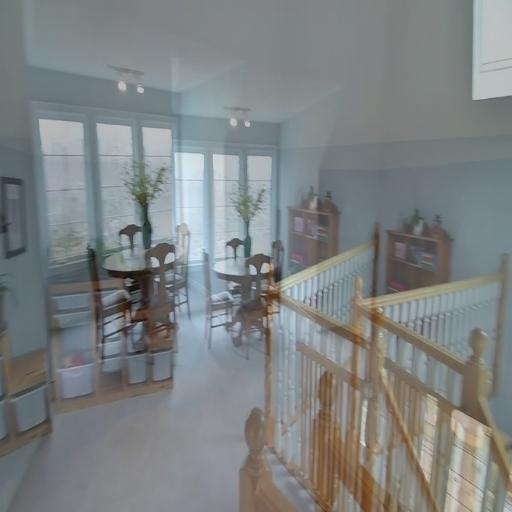}
        \caption{Before alignment}
    \end{subfigure}
    \begin{subfigure}{0.32\linewidth}
        \includegraphics[width=1\linewidth]{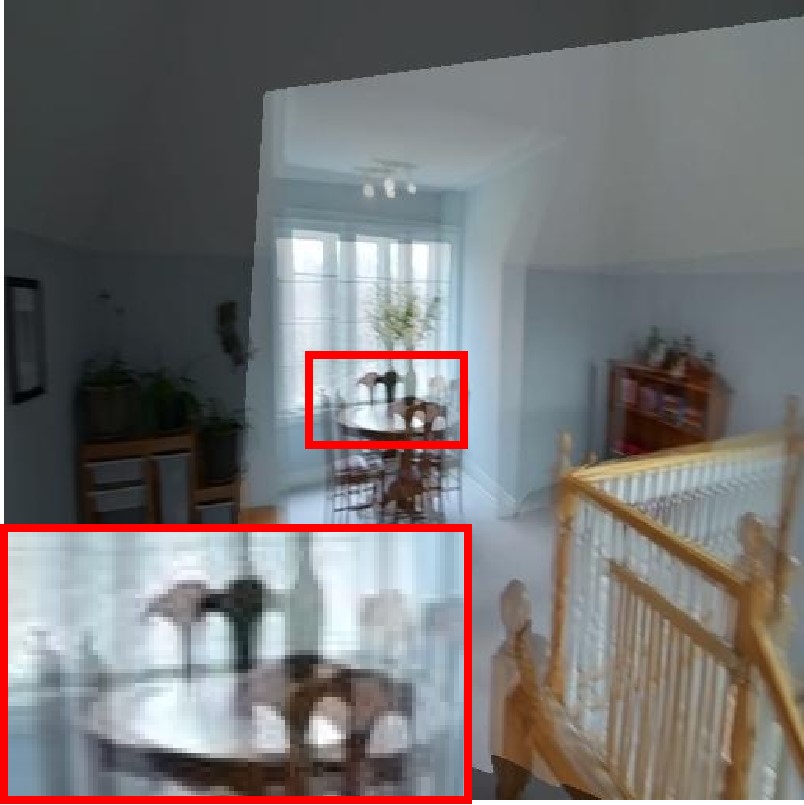}
        \caption{After 3D alignment}
    \end{subfigure}
    \begin{subfigure}{0.32\linewidth}
        \includegraphics[width=1\linewidth]{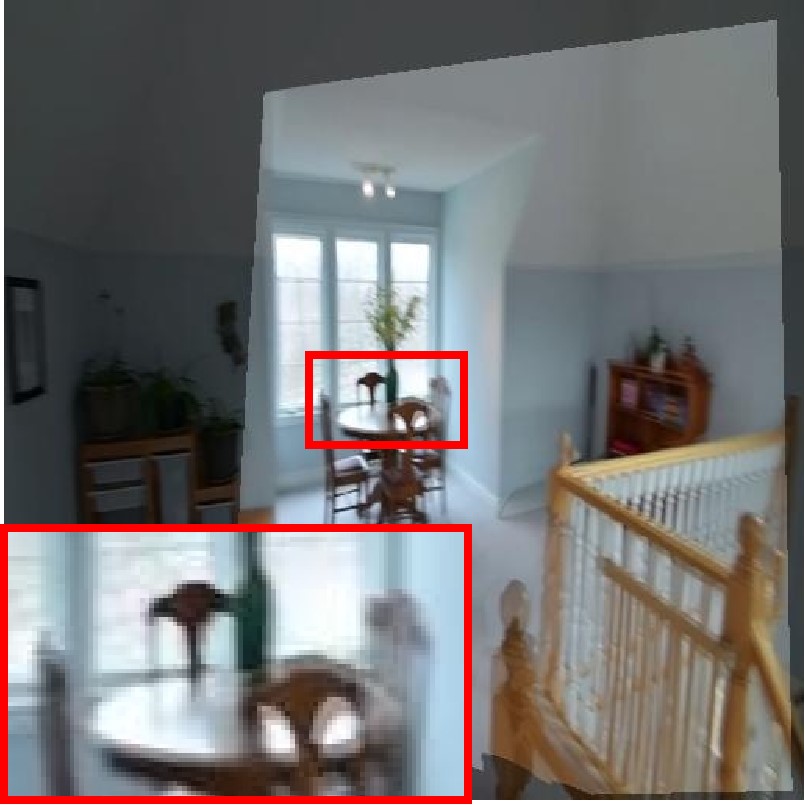}
        \caption{After 2D alignment}
    \end{subfigure}
    \caption{The effect of alignment.}
    \label{fig:the effect of alignment}
\end{figure}

\textbf{2D Module.}
Our 3D projection algorithm completely simulates the process of taking pictures with a camera and can achieve the right alignment when the camera is an ideal camera.
However, the real camera is not ideal, and there may be interference such as lens distortion.
Therefore, our 3D algorithm can achieve great alignment but sometimes is not fully aligned.
To solve this problem, we add a step of 2D alignment after 3D alignment.

As shown in \cref{fig:the effect of alignment}, the reference image and target image are coplanar after 3D alignment but still has some deviations.
Most of these deviations are translations and rotations in the 2D image coordinate system, and a few deviations involve size differences, so we use Pose2DNet to achieve a Scaled Euclidean transformation.
The output of Pose2DNet is a vector with 4 elements corresponding to a 2D rotation angle $\theta$, two translation variables $t_{x}$, $t_{y}$, and a scale variable $s$, the $I_{fine}$ can be obtained by:
\begin{equation}
  \left[\begin{array}{c}
      u_{fine}\\
      v_{fine}\\
      1
      \end{array}
  \right]
  =
  \left[\begin{array}{ccc}
      s\cos{\theta} & -s\sin{\theta} & t_{x}\\
      s\sin{\theta} & s\cos{\theta} & t_{y}\\
      0 & 0 & 1
      \end{array}
  \right]
  \times
  \left[\begin{array}{c}
      u_{coarse}\\
      v_{coarse}\\
      1
      \end{array}
  \right]
  \label{eq:2D transform}
\end{equation}
where $u_{fine}$ and $v_{fine}$ indicate the pixel coordinate of $I_{fine}$, $u_{coarse}$ and $v_{coarse}$ indicate the pixel coordinate of $I_{coarse}$.

To achieve better image filling, our 2D alignment algorithm focuses on the alignment accuracy around the hole region.
We dilate the original mask $M$ so that it can cover the area around the hole, we define the dilated mask as $M_{big}$, where the area around the hole region with value 0 and elsewhere with 1.
The loss of 2D module is:
\begin{equation}
  \mathcal{L}_{2D}
  =
  \Vert  (1 - M_{big}) \odot M_{valid} \odot (I_{fine} - I_{o}) \Vert_1
  \label{eq:2D loss}
\end{equation}
which means that we only calculate the loss around the hole region and it forces the Pose2DNet to pay more attention to the alignment accuracy of the hole region.

All of our image alignment algorithms use self-supervised training, which means that our model can be generalized to various image datasets with reference images regardless of whether there are image pose labels.

\begin{figure}[t]
    \centering
    \includegraphics[width=1\linewidth]{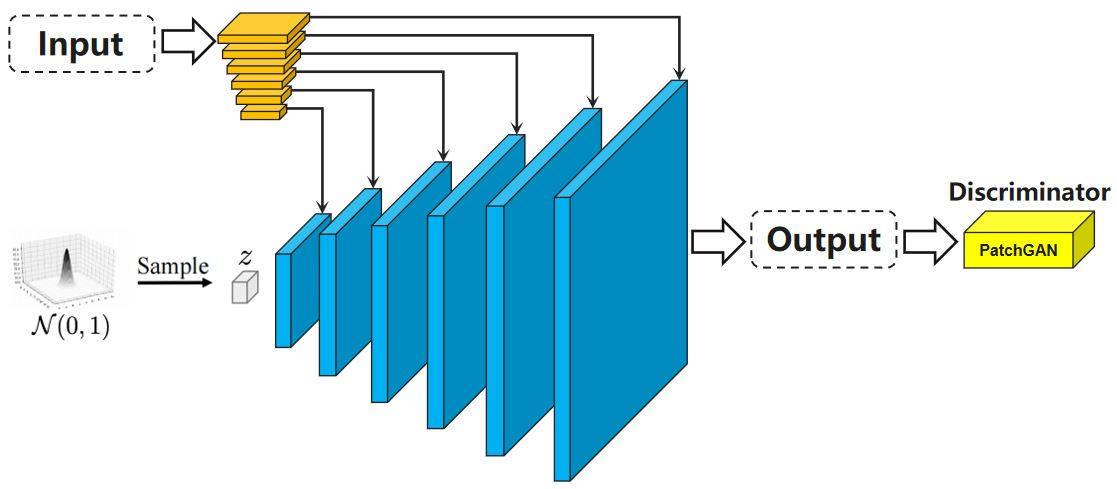}
    \caption{Structure of the InpaintNet.}
    \label{fig:structure of the InpaintNet}
\end{figure}

\subsection{Image Refinement Module}
After obtaining the $I_{fine}$, we fill it to the hole region of the target image to generate a filled image $I_{fill}$:
\begin{equation}
  I_{fill}
  =
  I_{t} \odot M + I_{fine} \odot (1 - M)
  \label{eq:Ifill-fill}
\end{equation}
$I_{fill}$ roughly restores the missing area on the target image, however, there are still problems such as slight deviation in spatial position and inconsistency in brightness, which makes the whole picture look incoherent.

Our Image Refinement (IR) module uses a conditional generation network (CGAN) to improve the above issues.
This CGAN in our algorithm is called as InpaintNet and its structure is shown in \cref{fig:structure of the InpaintNet}.
The inputs of InpaintNet are $I_{fill}$ and $M$, the model extracts features of different levels of the input images and inputs these features as conditional information or prior information into different levels of the generation network.
The generation network samples a latent vector $z$ from a standard Gaussian distribution and modulates $z$ with the condition information to generate an image.
To improve the sharpness of the generated images $I_{g}$, we use PatchGAN \cite{patchGAN} as a discriminator for adversarial training.
The loss function of the generation network is:
\begin{equation}
  \mathcal{L}_{inpaint}
  =
  \lambda_{1} \Vert (I_{g} - I_{o}) \Vert_1 + \lambda_{2} VGG(I_{g}, I_{o}) + \lambda_{3} \mathcal{L}_{adv}
  \label{eq:inpaint loss}
\end{equation}
Here, the VGG loss is the perceptual loss matching features of 5 layers of a pre-trained VGG19 \cite{VGG19},
$\mathcal{L}_{adv}$ is the adversarial loss calculated by PatchGAN.
Finally, we fill the $I_{g}$ into the hole region of the target image to get the final inpainting result $I_{inpaint}$:
\begin{equation}
  I_{inpaint}
  =
  I_{t} \odot M + I_{g} \odot (1 - M)
  \label{eq:Iinpaint-fill}
\end{equation}

\begin{figure*}[t]
    \centering
    \begin{subfigure}{0.135\linewidth}
        \includegraphics[width=1\linewidth]{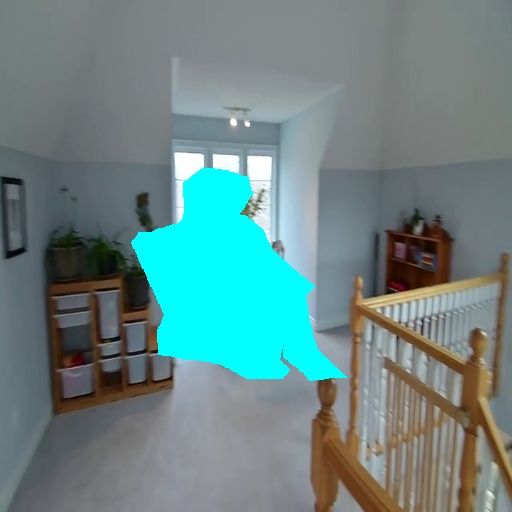}
    \end{subfigure}
    \begin{subfigure}{0.135\linewidth}
        \includegraphics[width=1\linewidth]{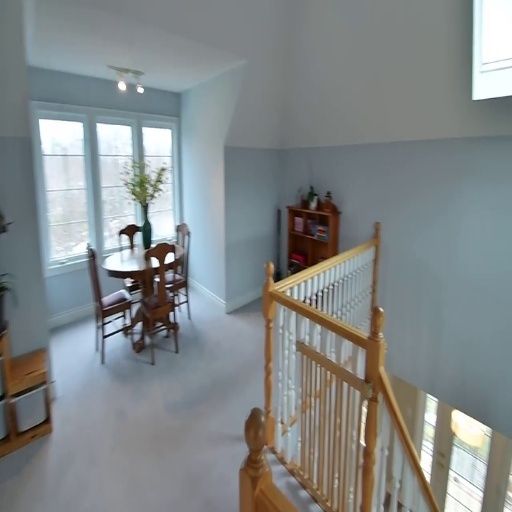}
    \end{subfigure}
    \hfill
    \begin{subfigure}{0.135\linewidth}
        \includegraphics[width=1\linewidth]{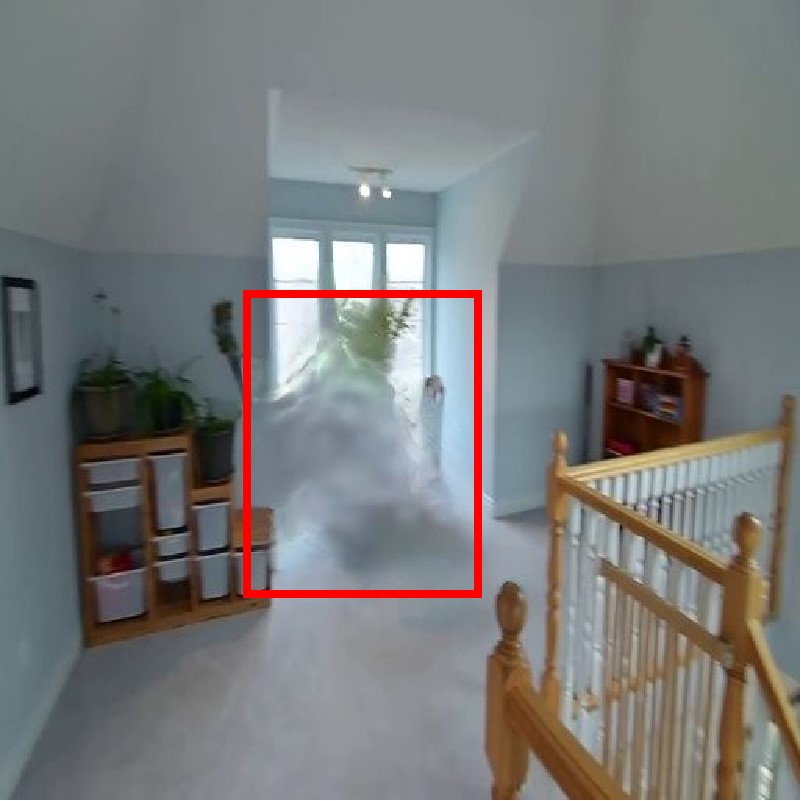}
    \end{subfigure}
    \begin{subfigure}{0.135\linewidth}
        \includegraphics[width=1\linewidth]{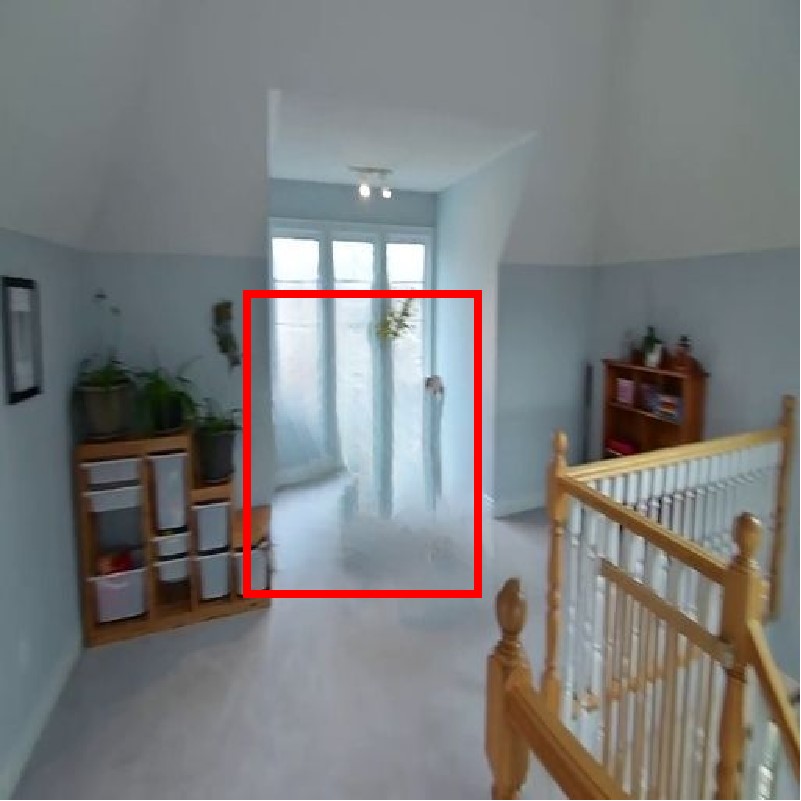}
    \end{subfigure}
    \begin{subfigure}{0.135\linewidth}
        \includegraphics[width=1\linewidth]{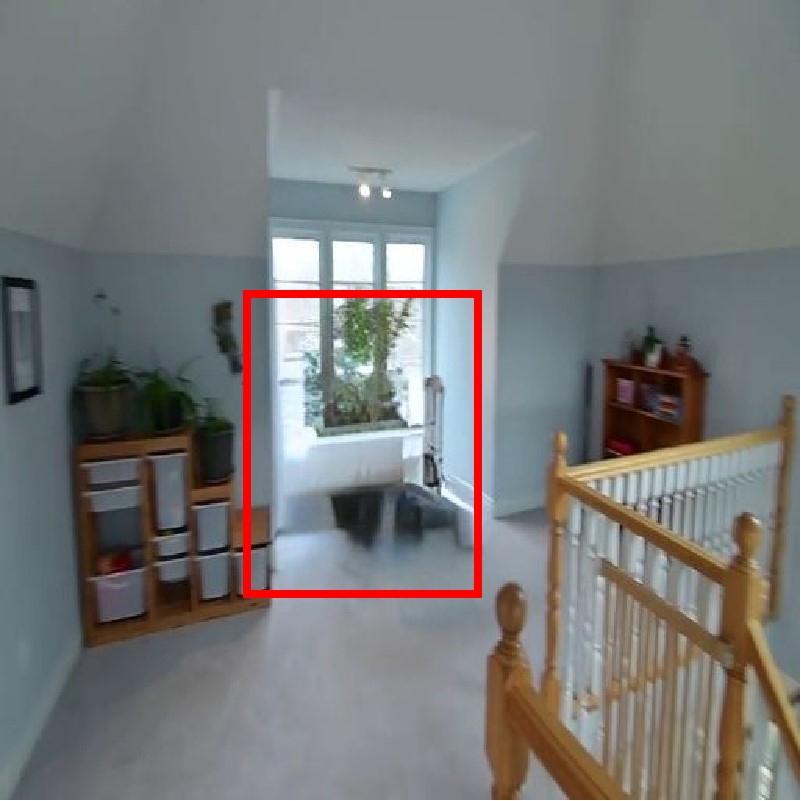}
    \end{subfigure}
    \begin{subfigure}{0.135\linewidth}
        \includegraphics[width=1\linewidth]{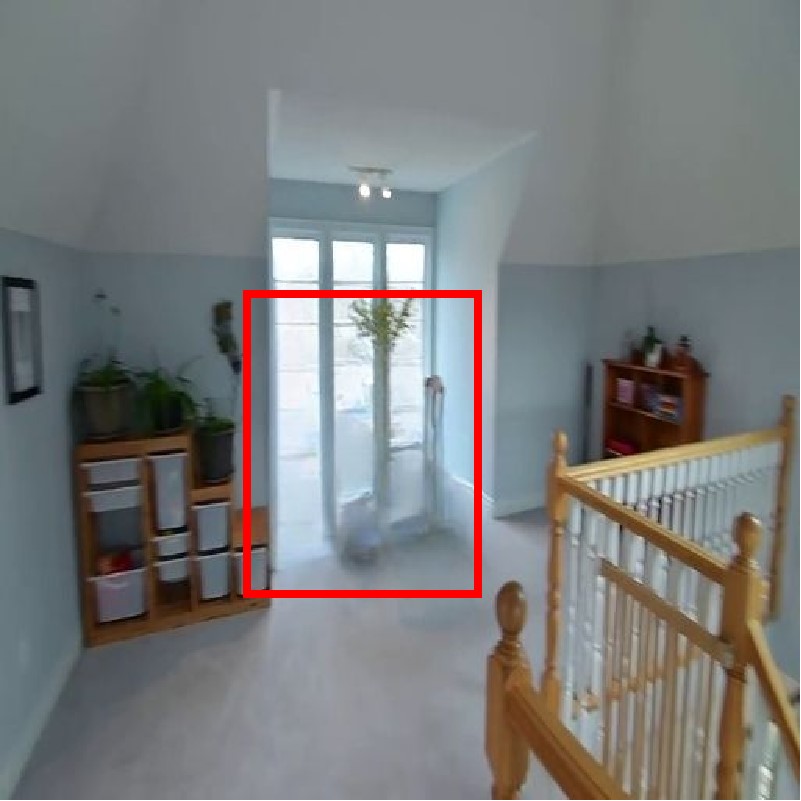}
    \end{subfigure}
    \begin{subfigure}{0.135\linewidth}
        \includegraphics[width=1\linewidth]{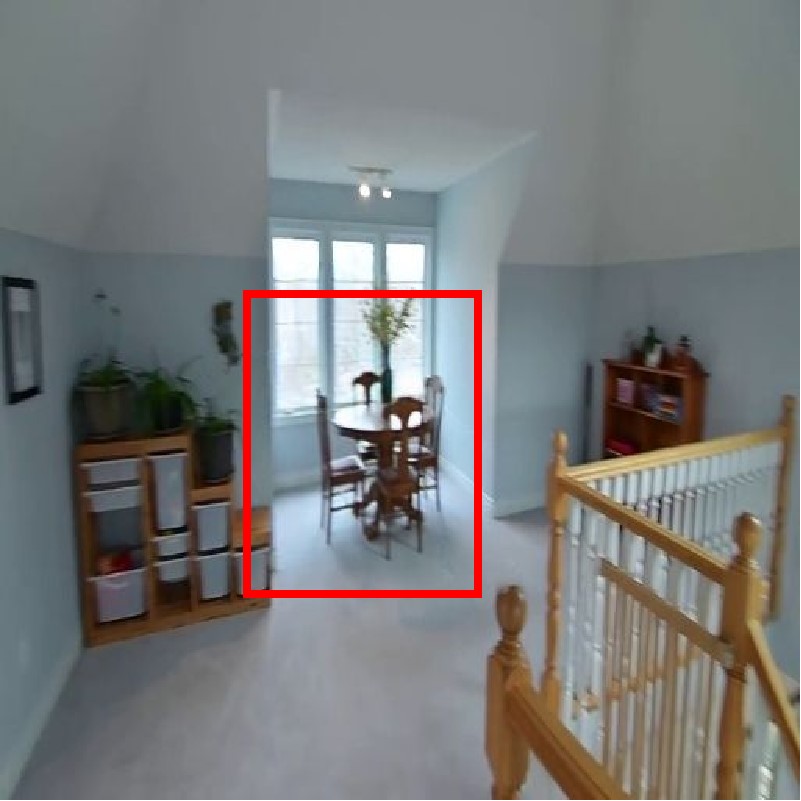}
    \end{subfigure}

    \centering
    \begin{subfigure}{0.135\linewidth}
        \includegraphics[width=1\linewidth]{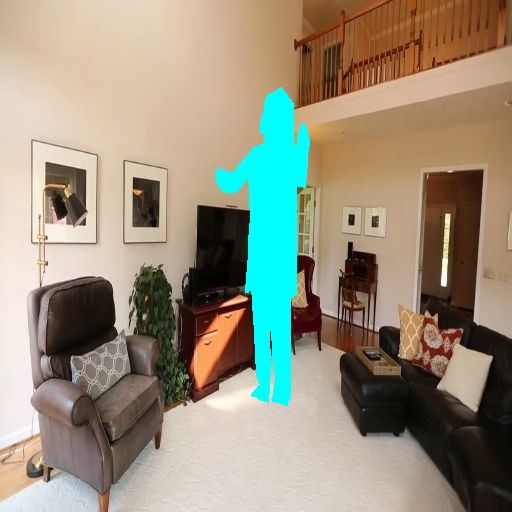}
    \end{subfigure}
    \begin{subfigure}{0.135\linewidth}
        \includegraphics[width=1\linewidth]{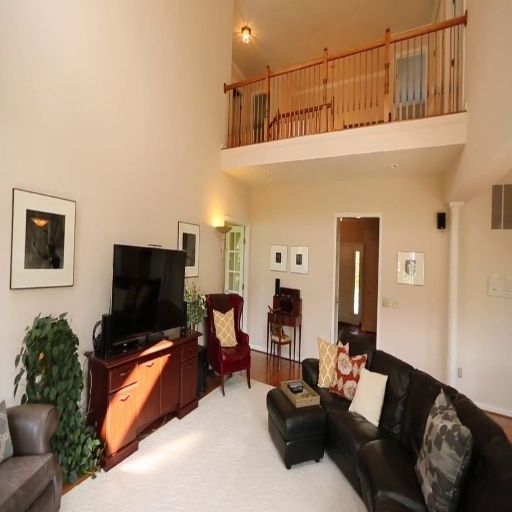}
    \end{subfigure}
    \hfill
    \begin{subfigure}{0.135\linewidth}
        \includegraphics[width=1\linewidth]{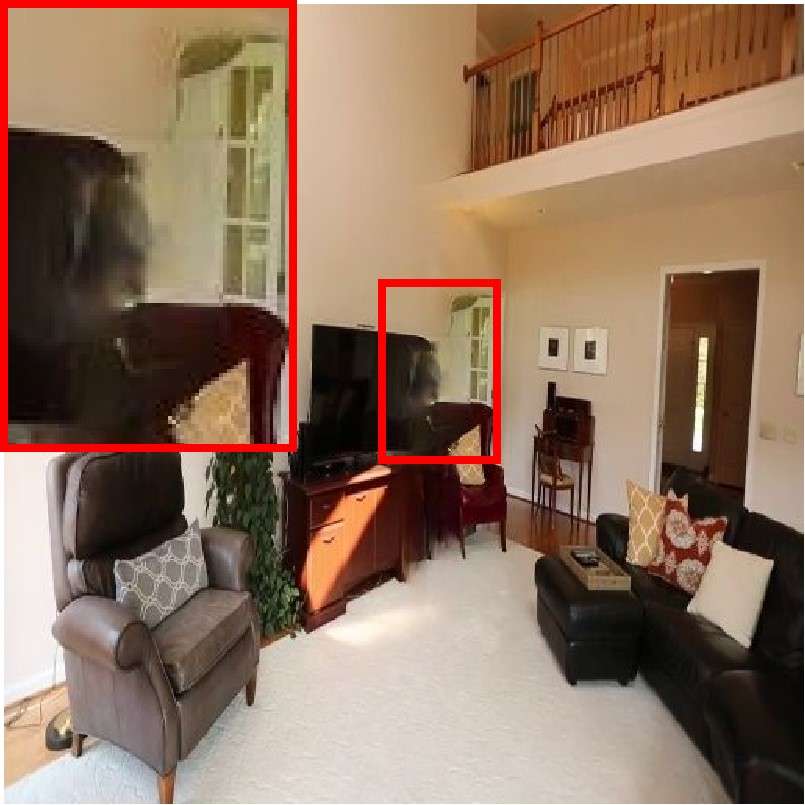}
    \end{subfigure}
    \begin{subfigure}{0.135\linewidth}
        \includegraphics[width=1\linewidth]{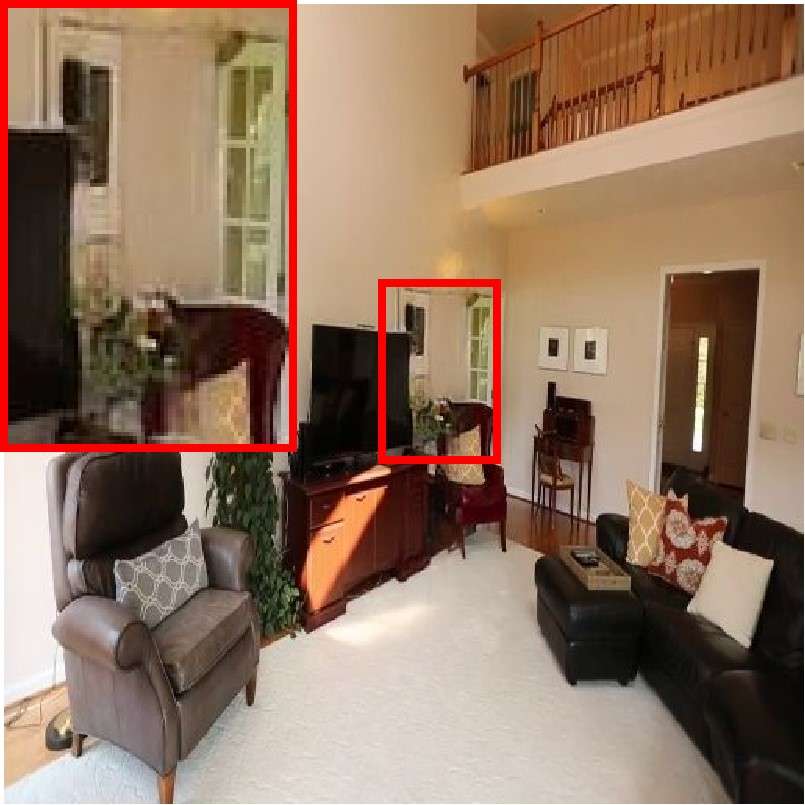}
    \end{subfigure}
    \begin{subfigure}{0.135\linewidth}
        \includegraphics[width=1\linewidth]{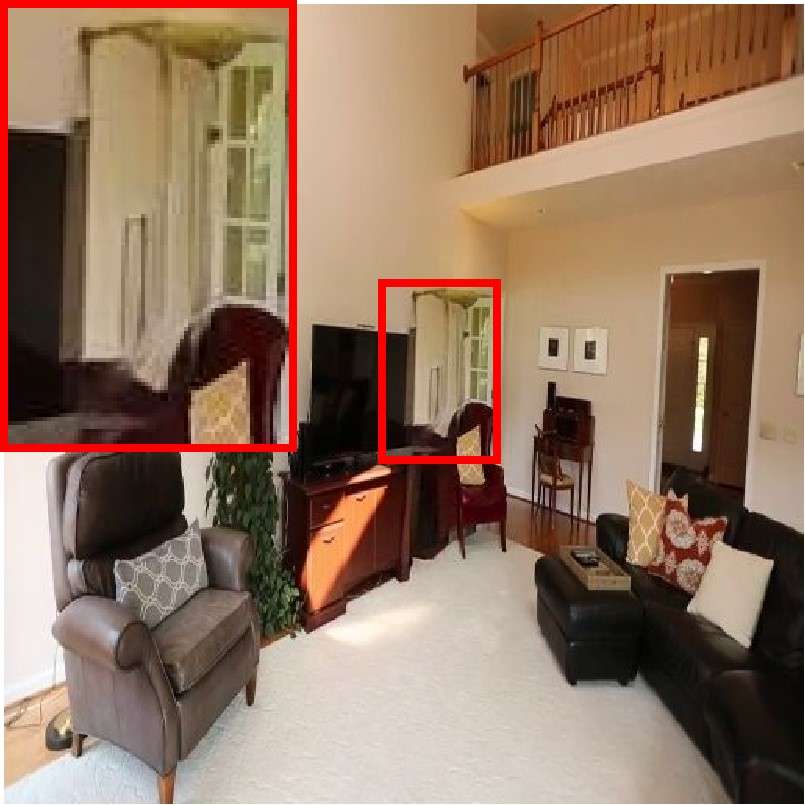}
    \end{subfigure}
    \begin{subfigure}{0.135\linewidth}
        \includegraphics[width=1\linewidth]{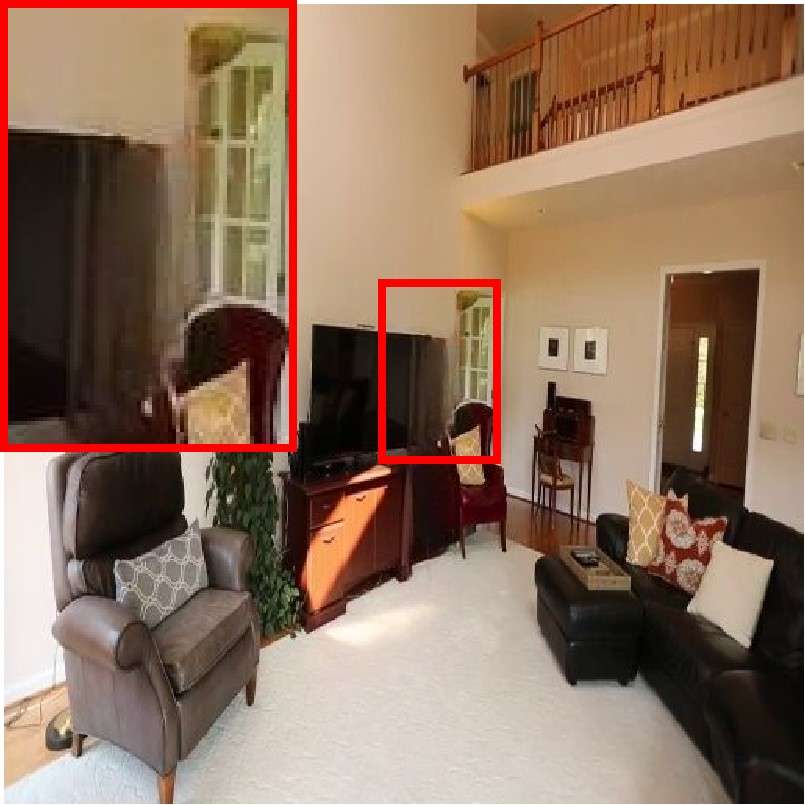}
    \end{subfigure}
    \begin{subfigure}{0.135\linewidth}
        \includegraphics[width=1\linewidth]{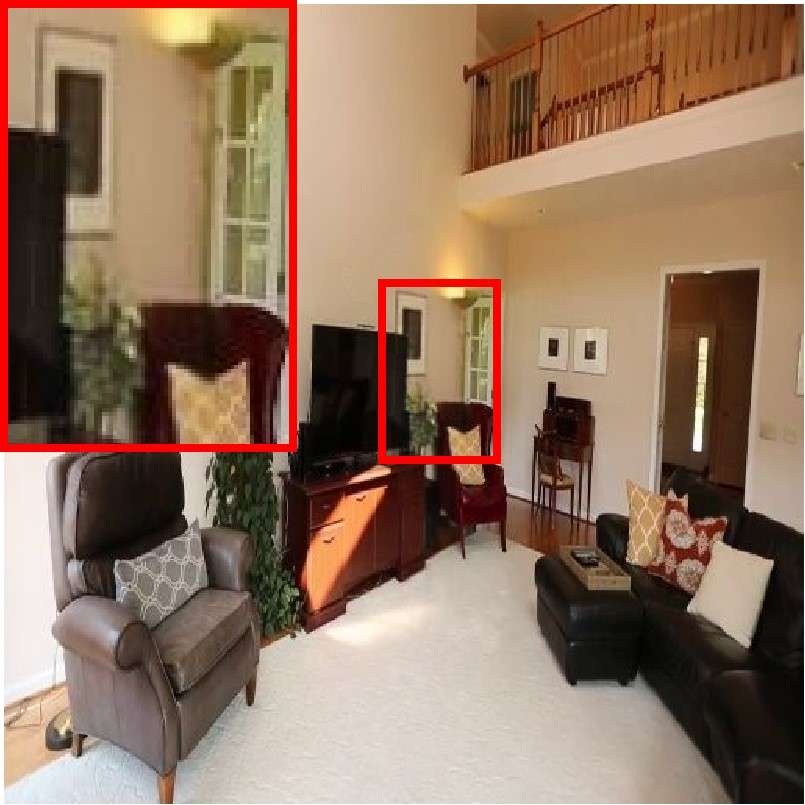}
    \end{subfigure}
    
    \centering
    \begin{subfigure}{0.135\linewidth}
        \includegraphics[width=1\linewidth]{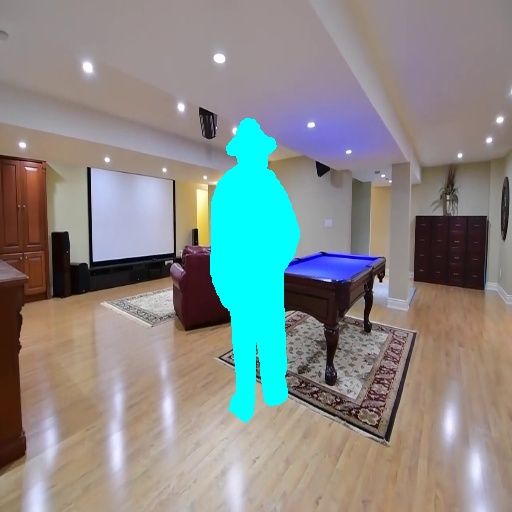}
    \end{subfigure}
    \begin{subfigure}{0.135\linewidth}
        \includegraphics[width=1\linewidth]{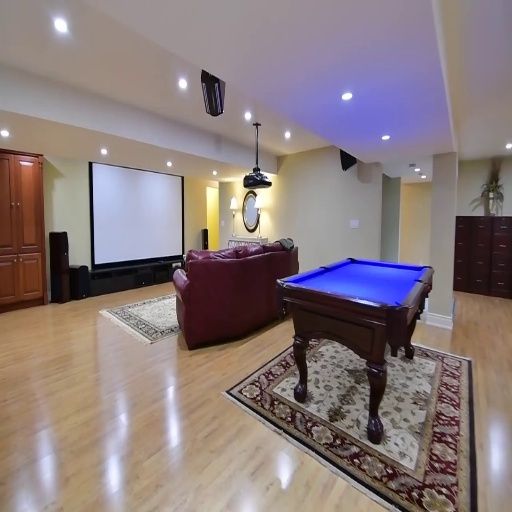}
    \end{subfigure}
    \hfill
    \begin{subfigure}{0.135\linewidth}
        \includegraphics[width=1\linewidth]{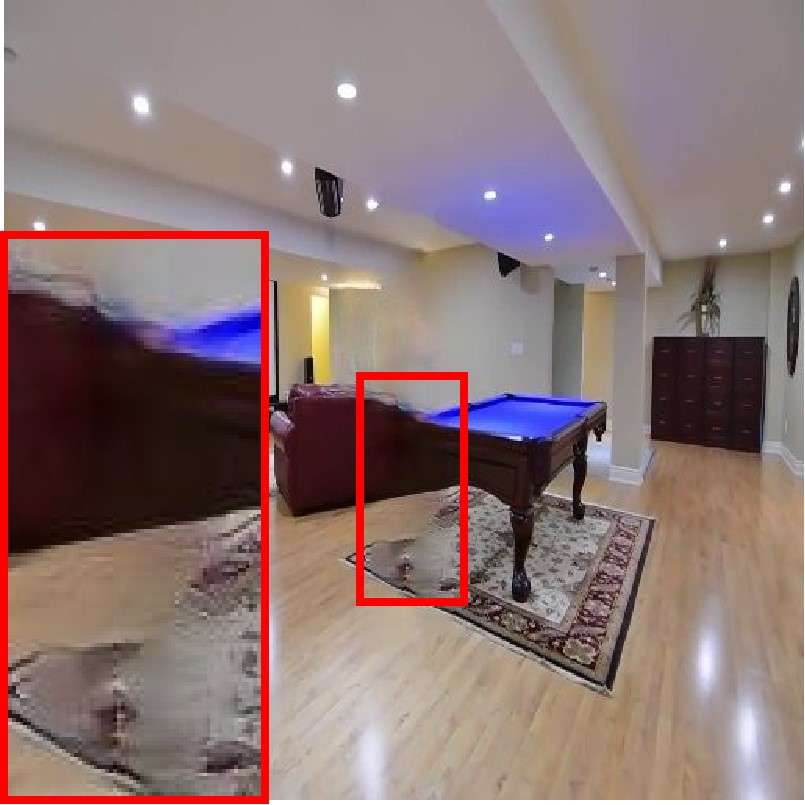}
    \end{subfigure}
    \begin{subfigure}{0.135\linewidth}
        \includegraphics[width=1\linewidth]{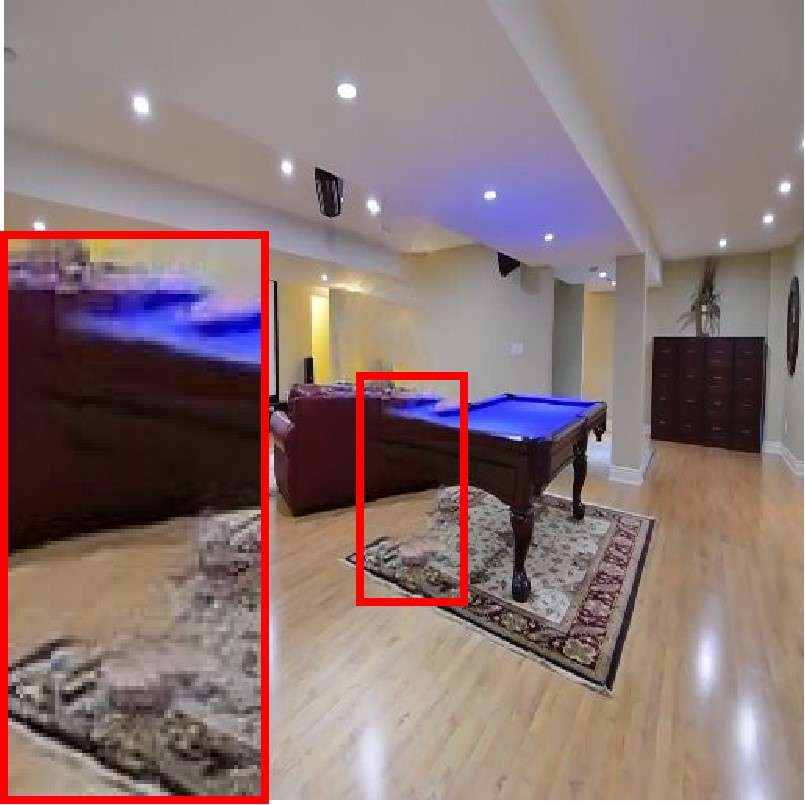}
    \end{subfigure}
    \begin{subfigure}{0.135\linewidth}
        \includegraphics[width=1\linewidth]{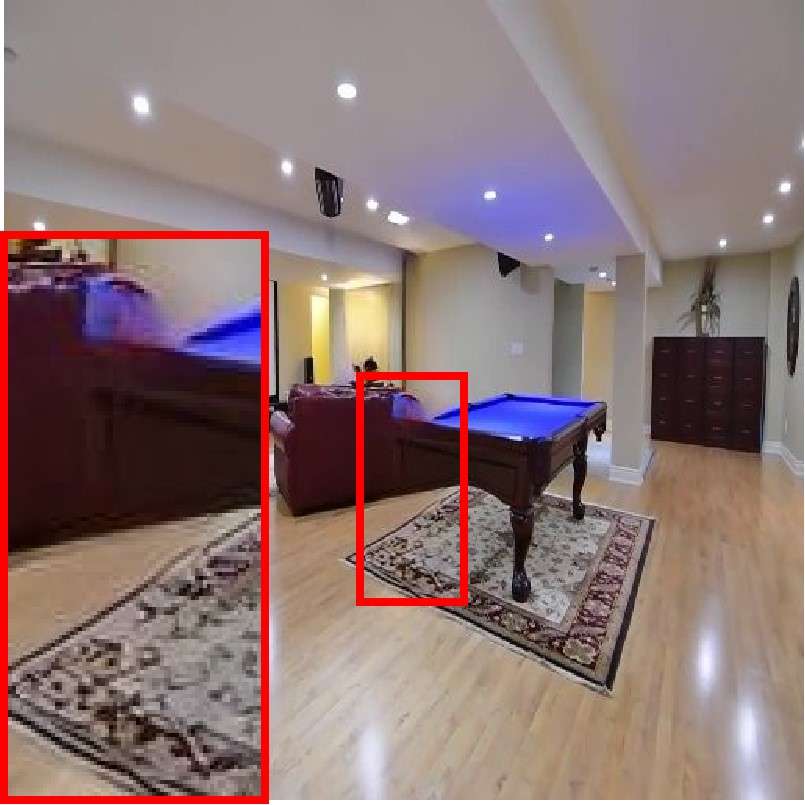}
    \end{subfigure}
    \begin{subfigure}{0.135\linewidth}
        \includegraphics[width=1\linewidth]{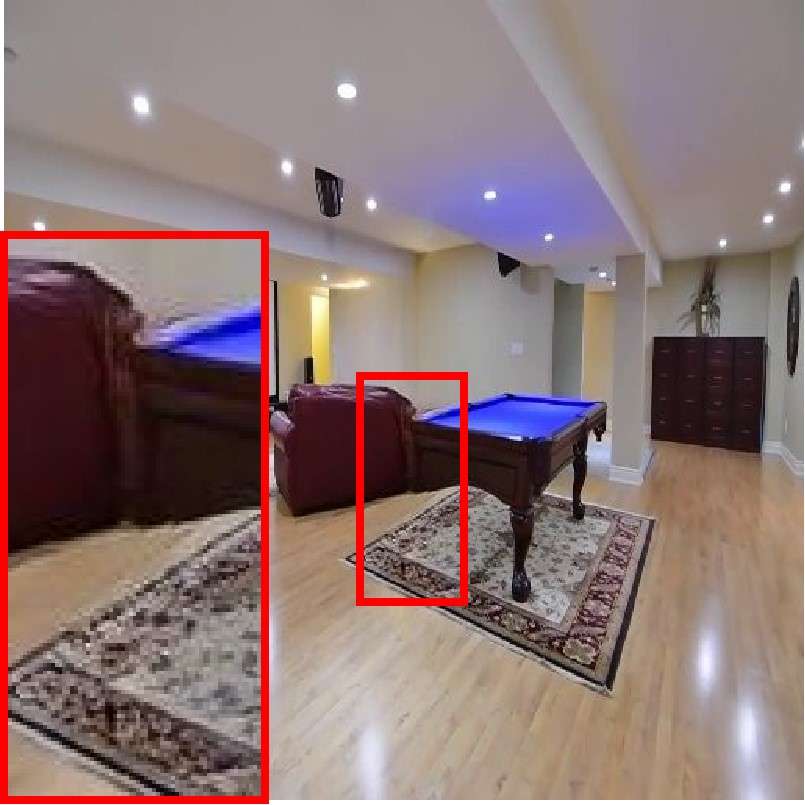}
    \end{subfigure}
    \begin{subfigure}{0.135\linewidth}
        \includegraphics[width=1\linewidth]{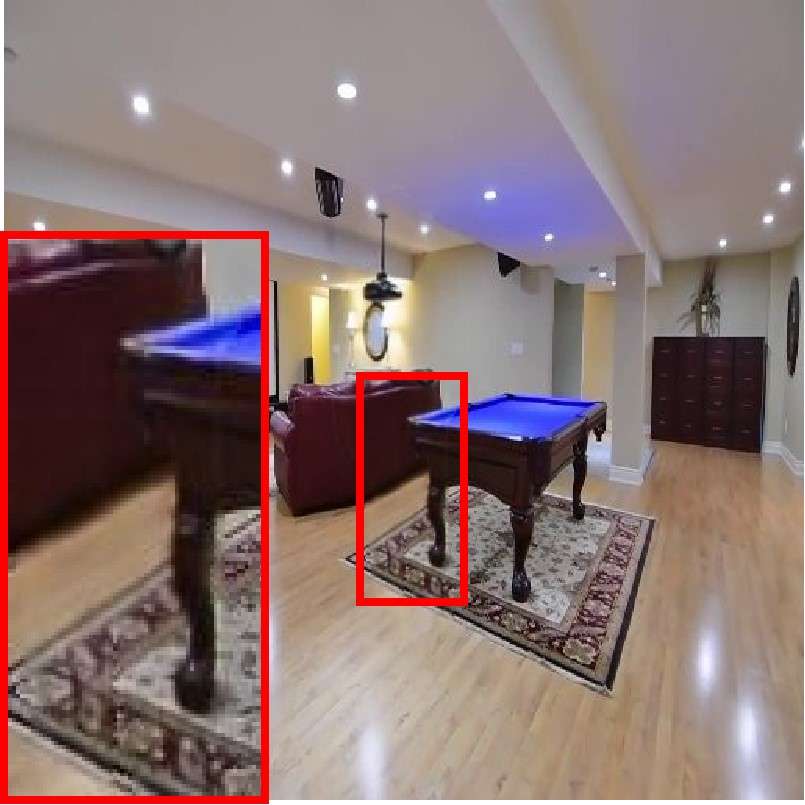}
    \end{subfigure}

    \centering
    \begin{subfigure}{0.135\linewidth}
        \includegraphics[width=1\linewidth]{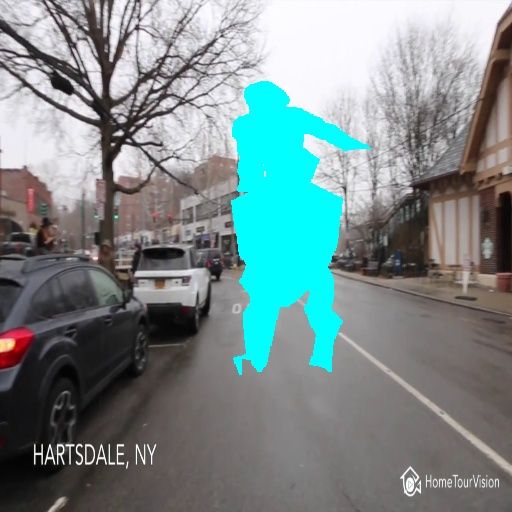}
    \end{subfigure}
    \begin{subfigure}{0.135\linewidth}
        \includegraphics[width=1\linewidth]{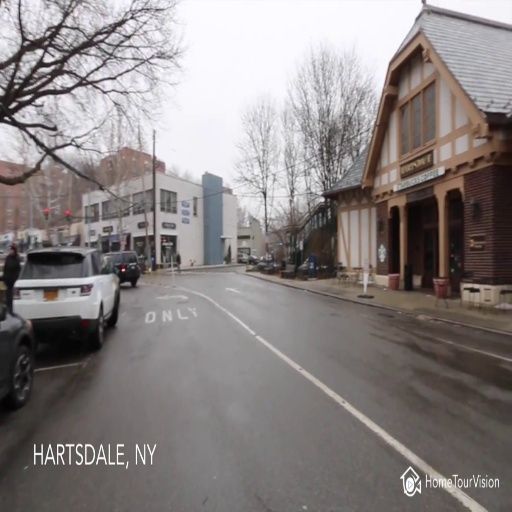}
    \end{subfigure}
    \hfill
    \begin{subfigure}{0.135\linewidth}
        \includegraphics[width=1\linewidth]{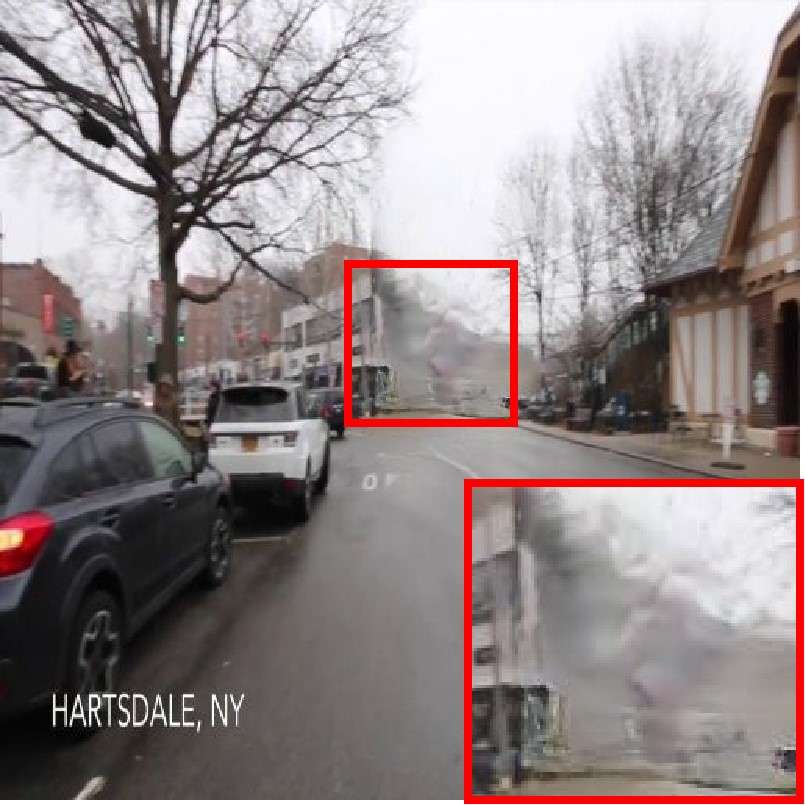}
    \end{subfigure}
    \begin{subfigure}{0.135\linewidth}
        \includegraphics[width=1\linewidth]{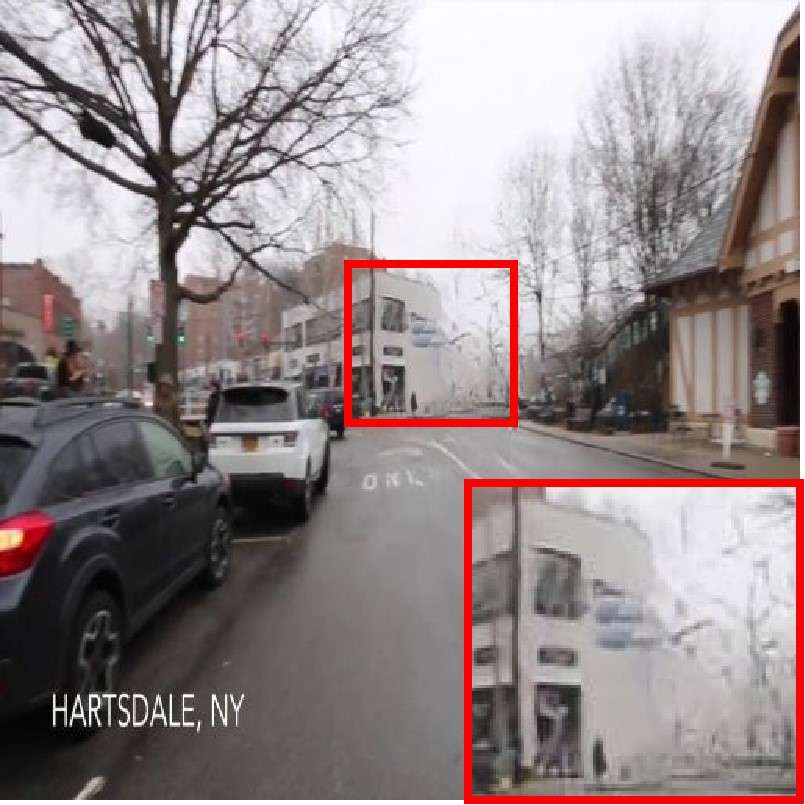}
    \end{subfigure}
    \begin{subfigure}{0.135\linewidth}
        \includegraphics[width=1\linewidth]{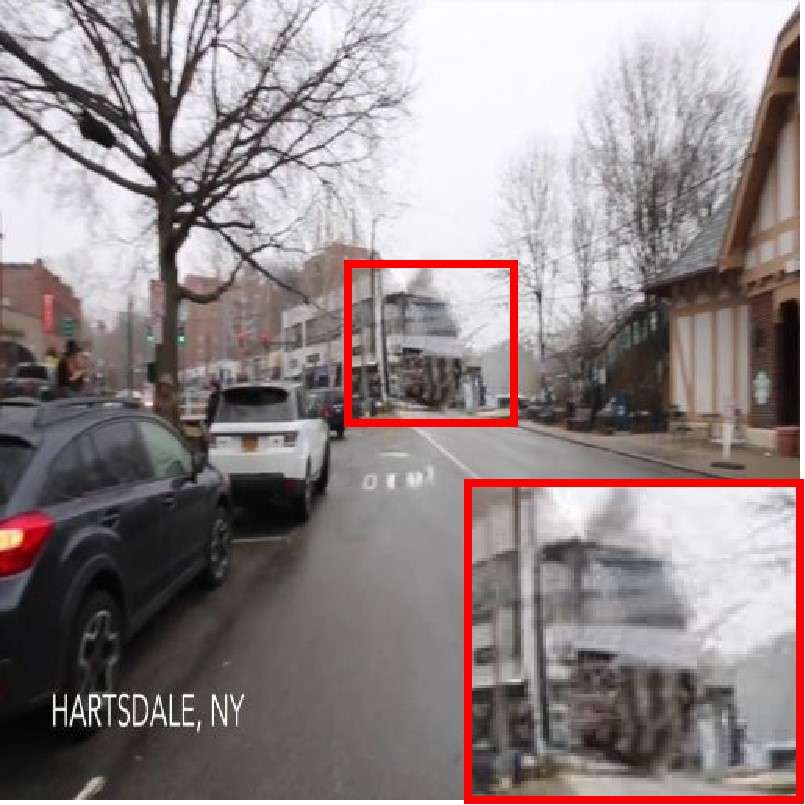}
    \end{subfigure}
    \begin{subfigure}{0.135\linewidth}
        \includegraphics[width=1\linewidth]{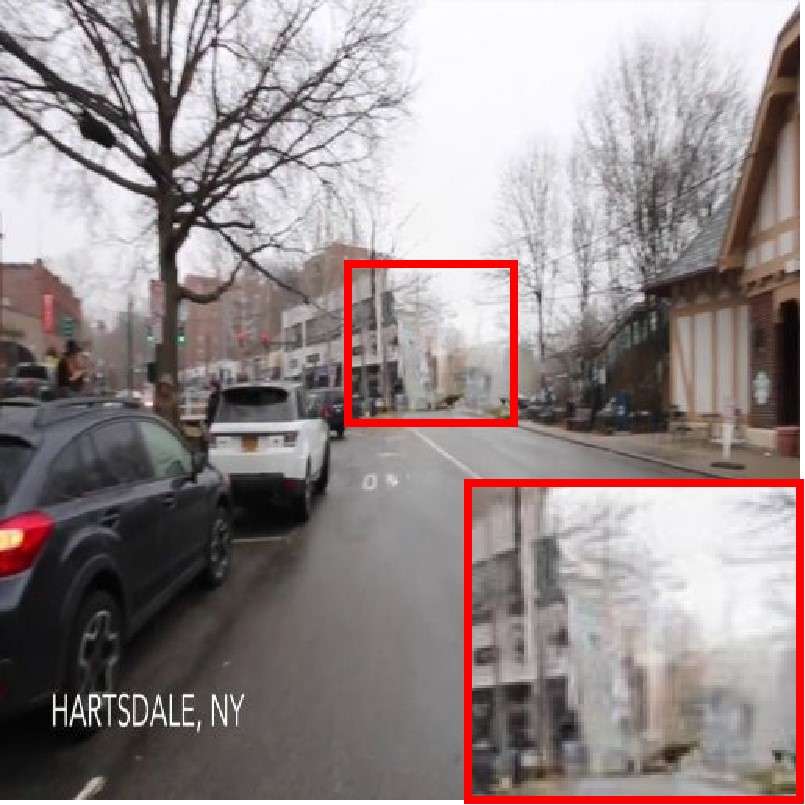}
    \end{subfigure}
    \begin{subfigure}{0.135\linewidth}
        \includegraphics[width=1\linewidth]{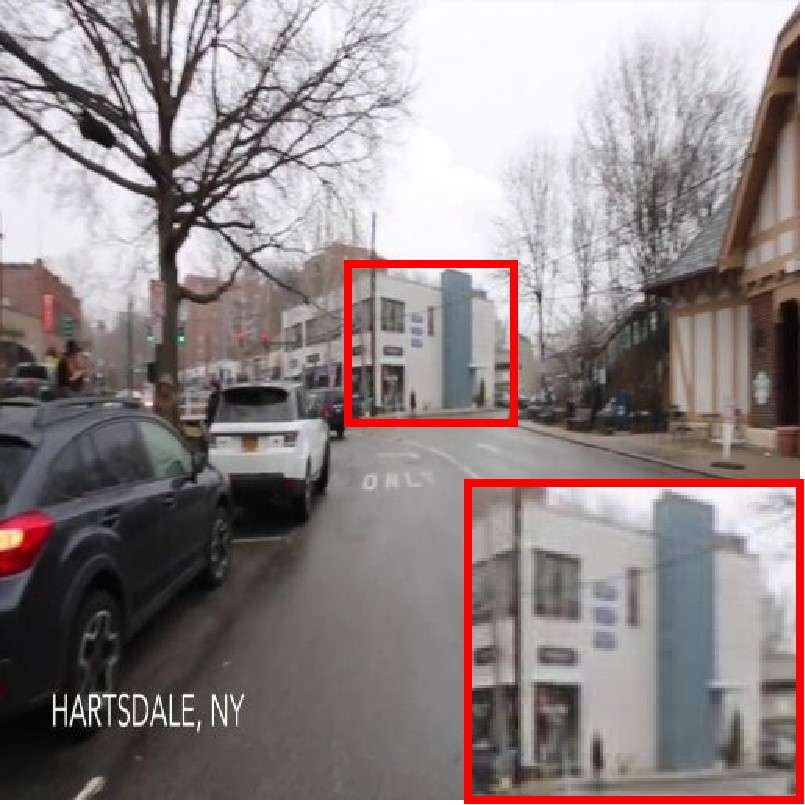}
    \end{subfigure}

    \centering
    \begin{subfigure}{0.135\linewidth}
        \includegraphics[width=1\linewidth]{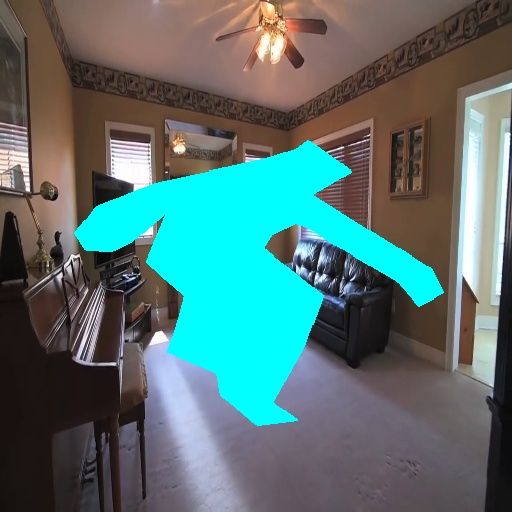}
    \end{subfigure}
    \begin{subfigure}{0.135\linewidth}
        \includegraphics[width=1\linewidth]{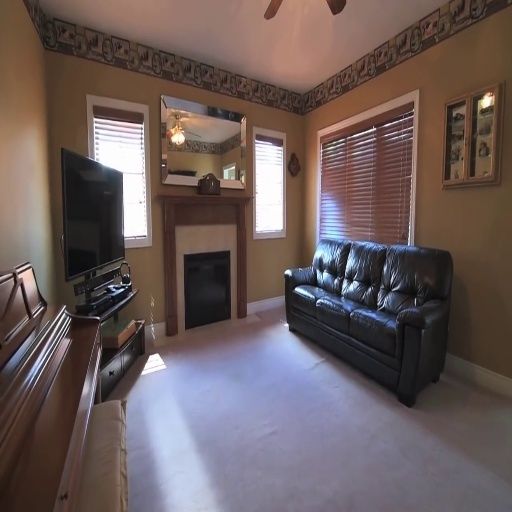}
    \end{subfigure}
    \hfill
    \begin{subfigure}{0.135\linewidth}
        \includegraphics[width=1\linewidth]{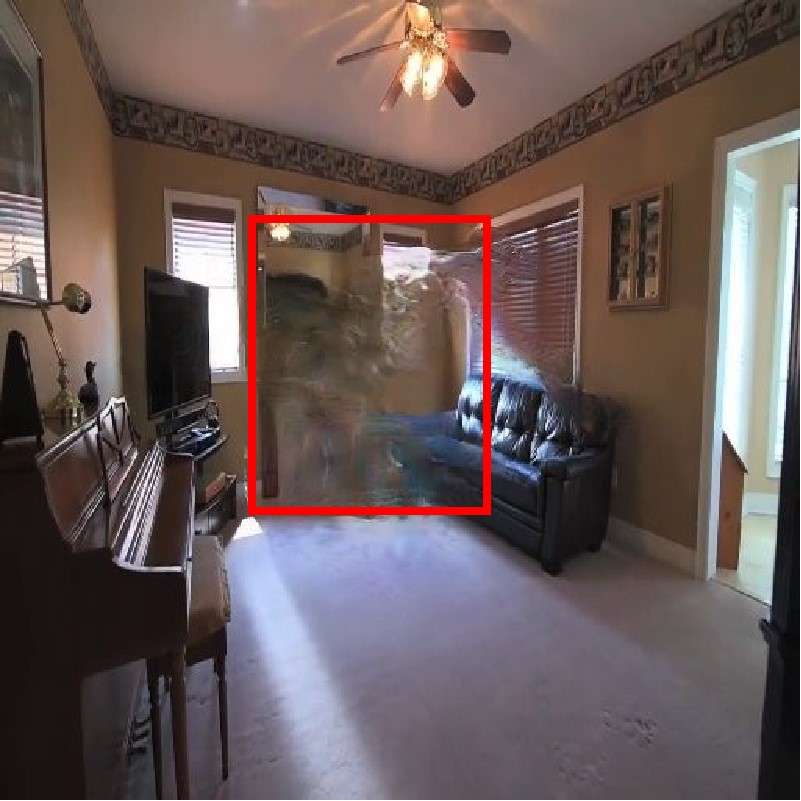}
    \end{subfigure}
    \begin{subfigure}{0.135\linewidth}
        \includegraphics[width=1\linewidth]{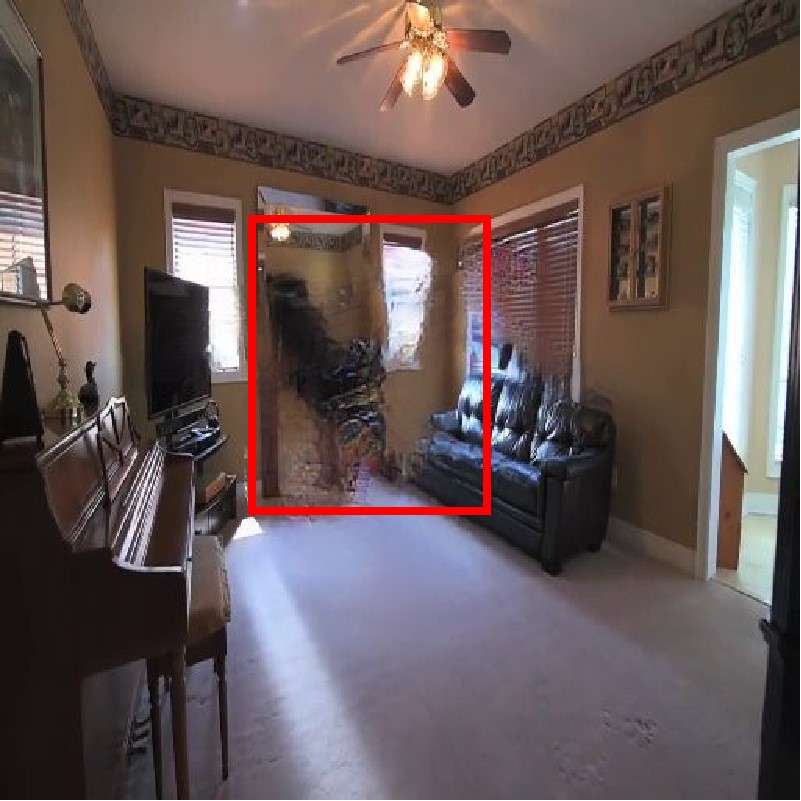}
    \end{subfigure}
    \begin{subfigure}{0.135\linewidth}
        \includegraphics[width=1\linewidth]{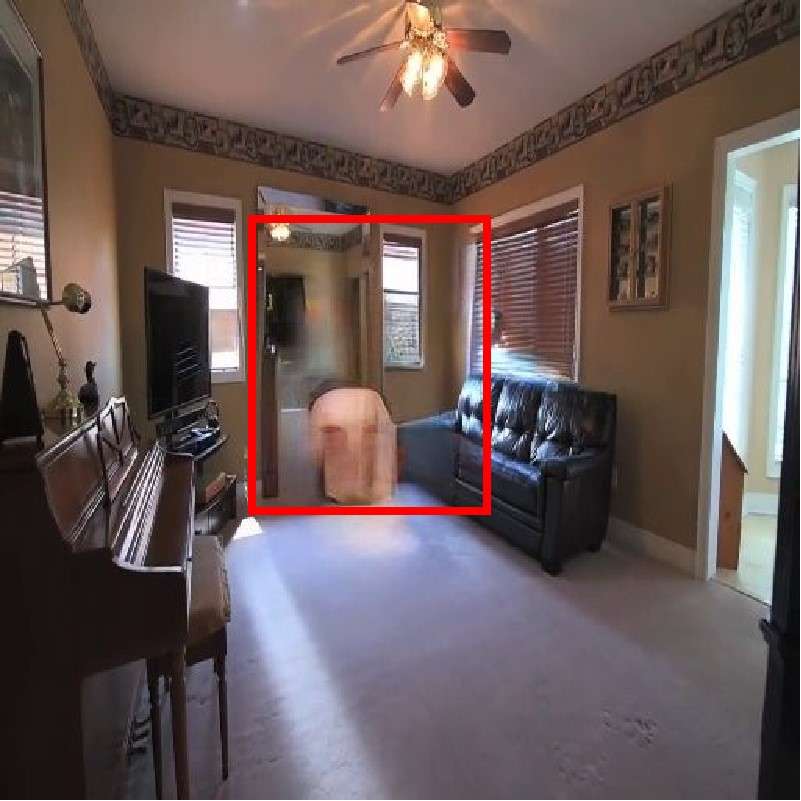}
    \end{subfigure}
    \begin{subfigure}{0.135\linewidth}
        \includegraphics[width=1\linewidth]{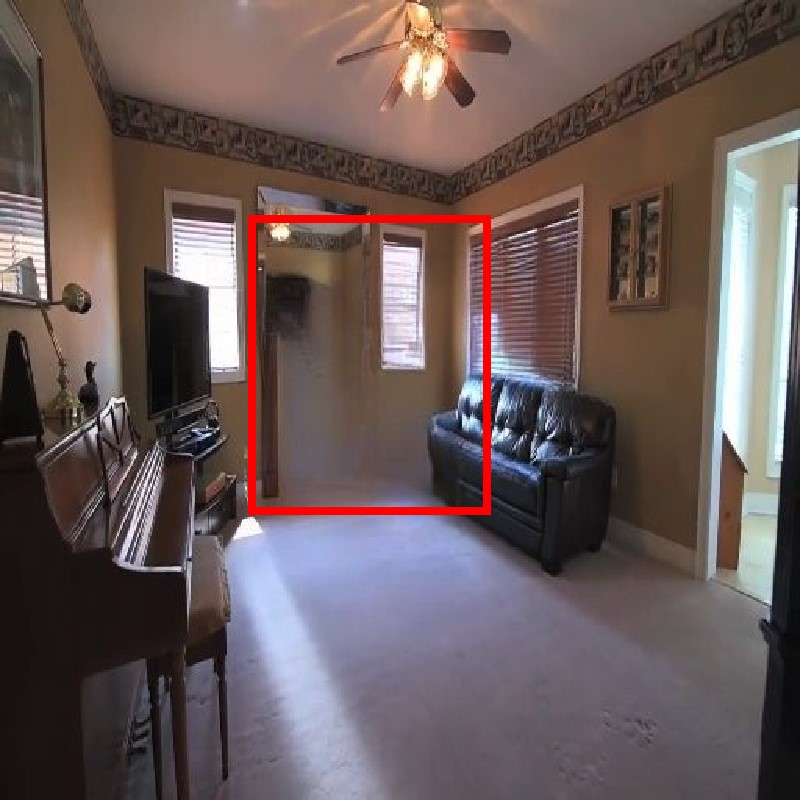}
    \end{subfigure}
    \begin{subfigure}{0.135\linewidth}
        \includegraphics[width=1\linewidth]{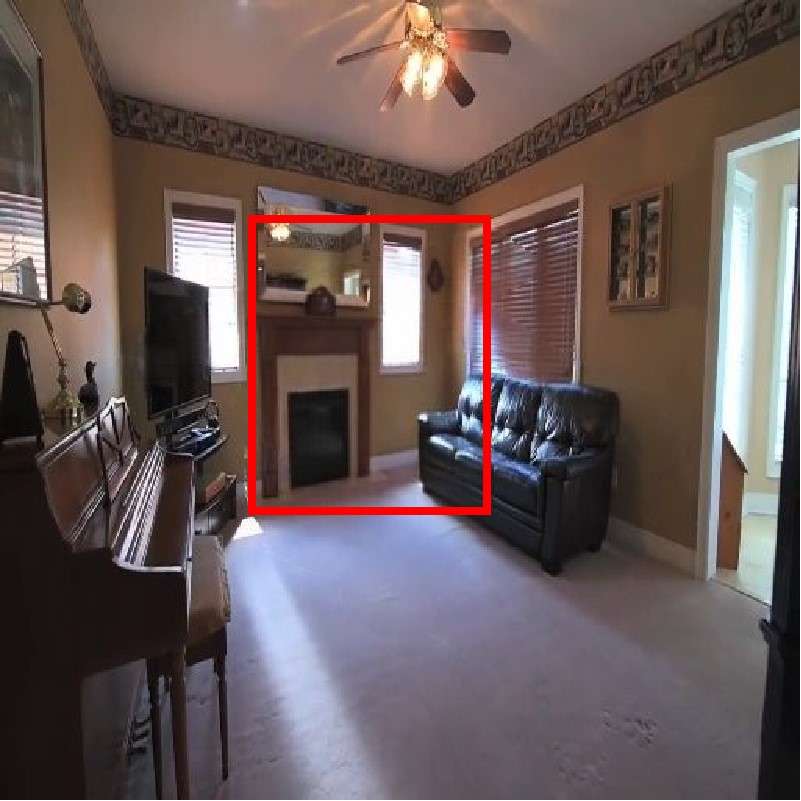}
    \end{subfigure}

    \centering
    \begin{subfigure}{0.135\linewidth}
        \includegraphics[width=1\linewidth]{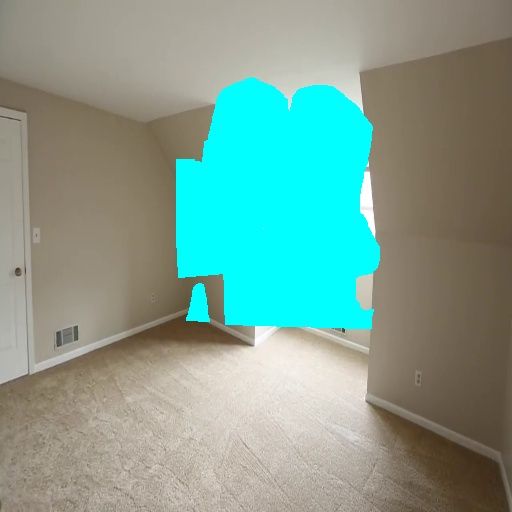}
        \caption{Target}
    \end{subfigure}
    \begin{subfigure}{0.135\linewidth}
        \includegraphics[width=1\linewidth]{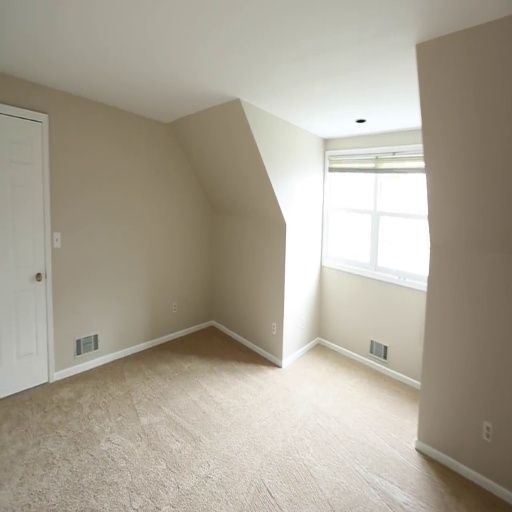}
        \caption{Reference}
    \end{subfigure}
    \hfill
    \begin{subfigure}{0.135\linewidth}
        \includegraphics[width=1\linewidth]{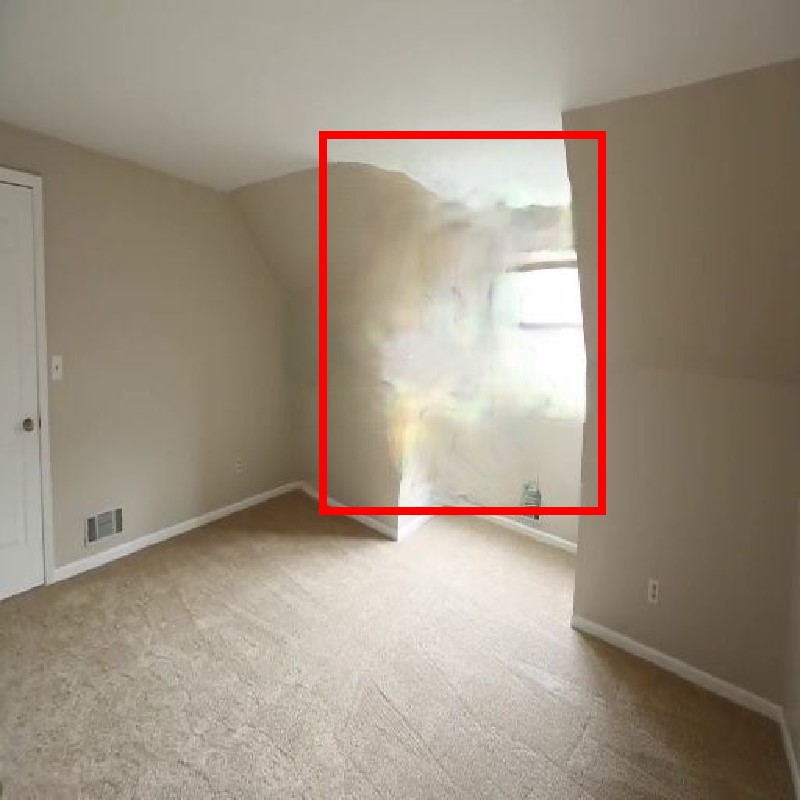}
        \caption{STTN}
    \end{subfigure}
    \begin{subfigure}{0.135\linewidth}
        \includegraphics[width=1\linewidth]{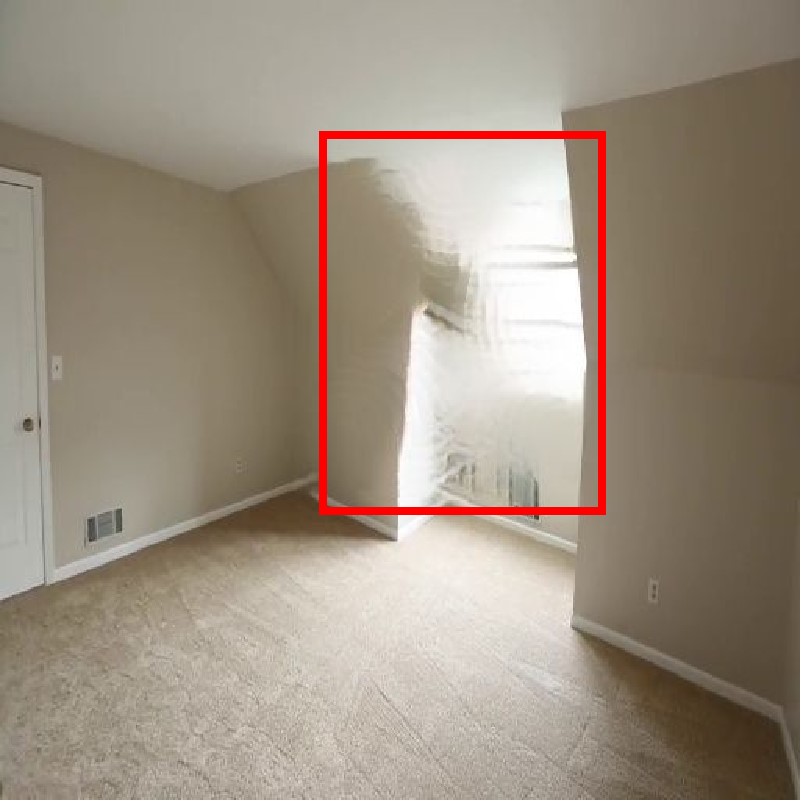}
        \caption{OPN}
    \end{subfigure}
    \begin{subfigure}{0.135\linewidth}
        \includegraphics[width=1\linewidth]{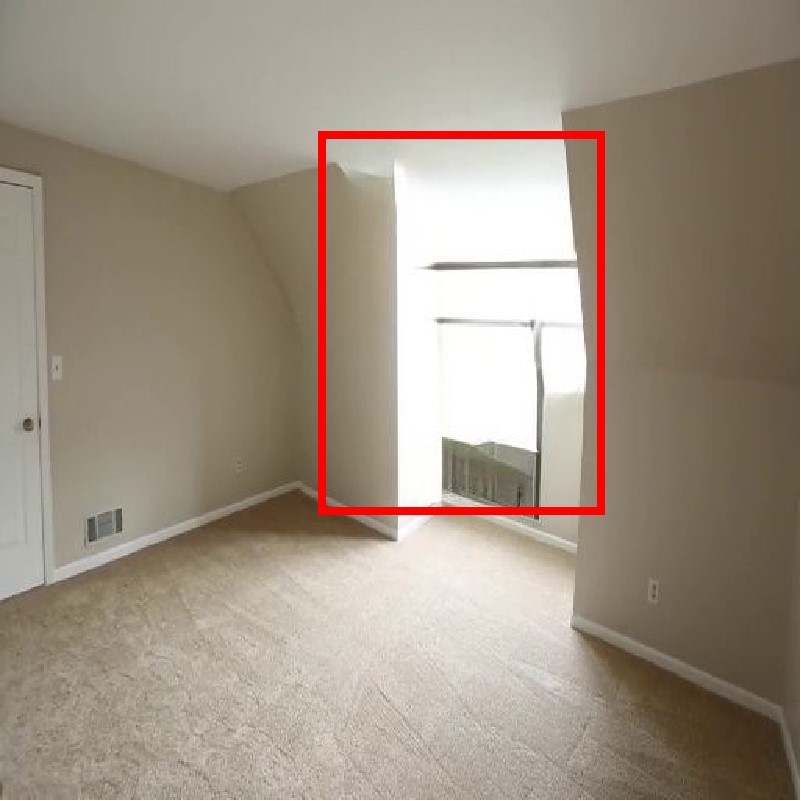}
        \caption{MAT}
    \end{subfigure}
    \begin{subfigure}{0.135\linewidth}
        \includegraphics[width=1\linewidth]{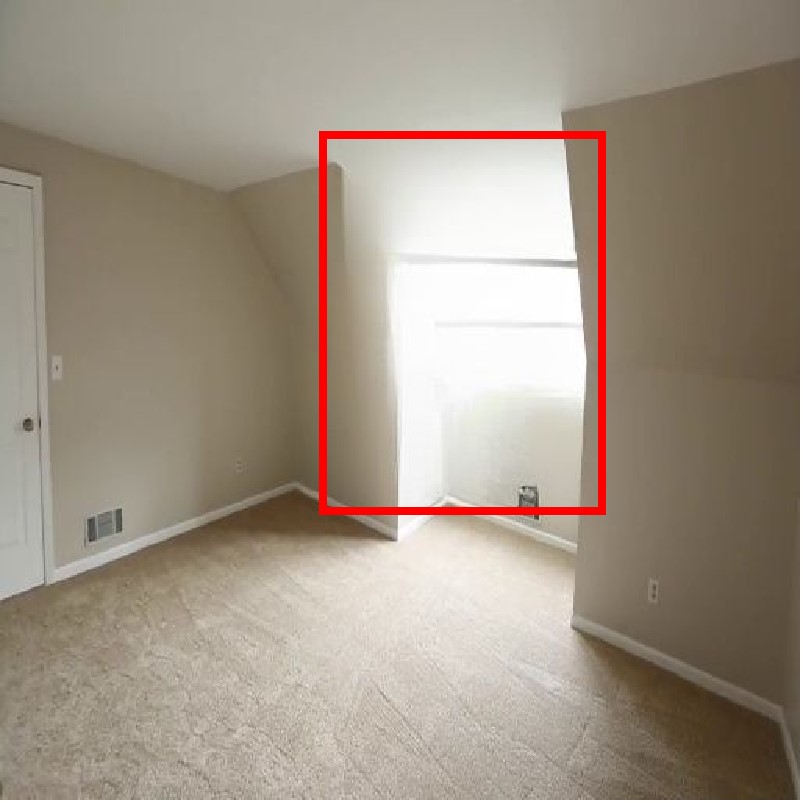}
        \caption{LaMa}
    \end{subfigure}
    \begin{subfigure}{0.135\linewidth}
        \includegraphics[width=1\linewidth]{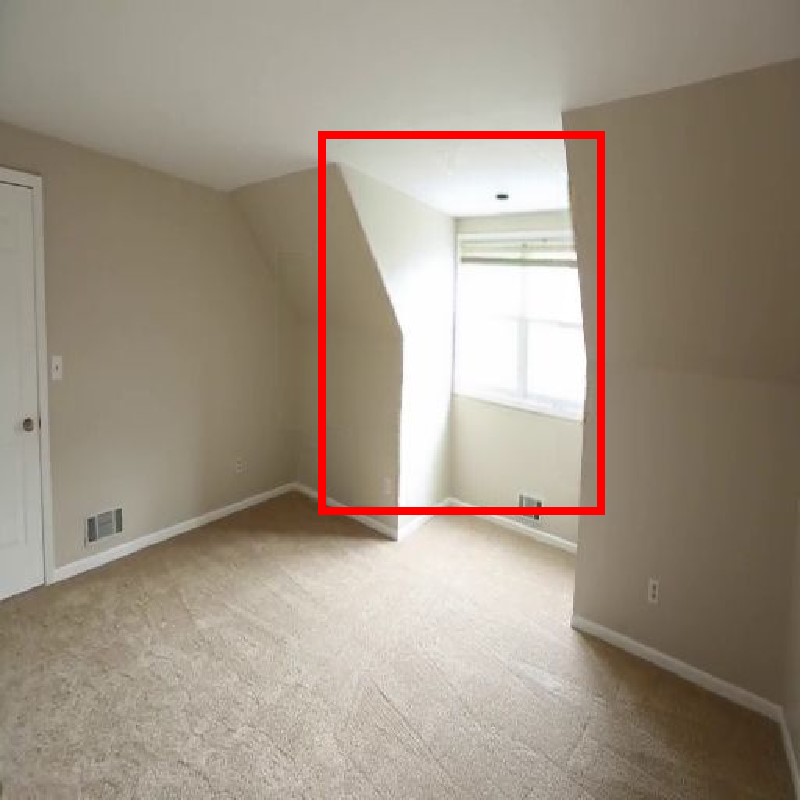}
        \caption{3DFill}
    \end{subfigure}
    \caption{Comparison with baselines on RealEstate10K. For better visualization, please zoom in to see the details.}
    \label{fig:qualitative comparison}
\end{figure*}
\section{Experiments}
\label{sec:experiments}
\textbf{Datasets.}
We trained and evaluated the 3DFill on the RealEstate10K dataset \cite{RealEstate}, which contains a series of videos from YouTube taken by moving cameras in various scenarios.
For each scene, we sampled 3 images with a displacement of 25 frames apart, we call this as original dataset, which contains 202K scenes.
We used the trained IA module to test each scene and got the valid map $M_{valid}$.
We found that when the valid area in $M_{valid}$ is between 60\% $\sim$ 80\%, the range of viewpoint changes in the scene was reasonable and acceptable, in other cases, the viewpoint changes were too large or too small.
Excessive viewpoint changes result in too little overlap between images, which is not the case for reference-guided image inpainting method to deal with.
Cases with slight viewpoint changes are too simple for training so we abandoned them, but experiments show that 3DFill can handle such cases.
We used this as a standard to clean the original dataset and got the final dataset, which contains 33.5K scenes, the training dataset has 32.5K scenes and the test dataset has 1K scenes.

For every scene, we set the first image as the original image, the second image as the small-view reference image, and the third image as the large-view reference image.
That means we had 65K image pairs for training.
We divided the test dataset into the \emph{small-view} dataset and \emph{large-view} dataset, consisting of the small-view reference image and the large-view reference image, respectively.
This division helps test the alignment ability of the model to varying degrees of viewpoint change.

For ease of processing, we resized each image to a resolution of 512 $\times$ 512.
We generated the mask of the human collected from the COCO dataset \cite{COCO}.
We divide the test dataset into two categories based on the area of the mask: \emph{small-mask} with the mask area accounting for 5\% $\sim$ 15\% of the image and \emph{large-mask} with the mask area accounting for 15\% $\sim$ 30\% of the image.

\textbf{Implementation Details.}
We applied the pre-trained DPT \cite{DPT} to estimate the depth map of input images and its model parameters were frozen during training.
In IA module, we first trained the Pose3DNet for 120 epochs and then trained Pose2DNet for 30 epochs.
Both of these models were optimized by Adam \cite{Adam} with $\beta_{1} = 0.9$, $\beta_{2} = 0.999$, and a learning rate of $1\times10^{-4}$.
In IR module, we trained the InpaintNet and PatchGAN together for 50 epochs and their optimizer were Adam with $\beta_{1} = 0.9$ and $\beta_{2} = 0.999$.
The learning rate of InpaintNet is $1\times10^{-3}$ and the learning rate of PatchGAN is $1\times10^{-4}$.
As for the loss function, we set $\lambda_{1} = 1$, $\lambda_{2} = 200$, and $\lambda_{3} = 0.01$ in \cref{eq:inpaint loss}.

\textbf{Compared Methods.}
We choose several representative image inpainting algorithms with or without prior information for comparison.
We regard the video inpainting algorithm as a kind of image inpainting algorithm that takes adjacent frames as reference images.

\noindent \textbf{FGVC} \cite{FGVC}: Flow-edge Guided Video Completion is a video completion algorithm which first extracts and completes motion edges, and then uses them to guide piecewise-smooth flow completion with sharp edges.

\noindent \textbf{STTN} \cite{STTN}: Spatial-Temporal Transformer Network fills missing regions in all input video frames by the proposed multi-scale patch-based attention modules.

\noindent \textbf{OPN} \cite{OPN_reference-guided}: given a set of reference images, Onion-Peel Network progressively fills the hole from the hole boundary enabling it to exploit richer contextual information for the missing regions.

\noindent \textbf{EdgeConnect} \cite{CNN_2_EdgeConnect}: EdgeConnect is a two-stage image inpainting method using hallucinated edges as a prior.

\noindent \textbf{MAT} \cite{MAT}: Mask-Aware Transformer is a transformer-based model for large mask inpainting with high fidelity and diversity.

\noindent \textbf{LaMa} \cite{noprior_LAMA}: LaMa is a state-of-the-art single-image inpainting method which uses fast Fourier convolutions (FFCs) achieving excellent performance in a range of datasets.

\begin{table*}
  \centering
  \resizebox{\linewidth}{!}{
  \begin{tabular}{c||c|c|c||c|c|c||c|c|c||c|c|c}
    \hline
     & \multicolumn{6}{|c||}{Small-mask} &  \multicolumn{6}{|c}{Large-mask} \\
    \hline
     & \multicolumn{3}{|c||}{Small-view} &  \multicolumn{3}{|c||}{Large-view}  & \multicolumn{3}{|c||}{Small-view} &  \multicolumn{3}{|c}{Large-view} \\
    \hline
    Model & PSNR & SSIM & LPIPS & PSNR & SSIM & LPIPS & PSNR & SSIM & LPIPS & PSNR & SSIM & LPIPS \\
    \hline\hline
    FGVC & 26.51 & 0.9475 & 0.0465 & 26.90 & 0.9468 & 0.0437 & 21.51 & 0.8792 & 0.1071 & 21.67 & 0.8771 & 0.1043 \\
    \hline
    STTN & 29.63 & 0.9628 & 0.0376 & 29.49 & 0.9625 & 0.0380 & 24.42 & 0.9113 & 0.0914 & 24.28 & 0.9108 & 0.0923 \\
    \hline
    OPN & 29.62 & 0.9630 & 0.0312 & 29.52 & 0.9616 & 0.0319 & 24.12 & 0.9104 & 0.0775 & 23.89 & 0.9068 & 0.0815 \\
    \hline
    EdgeConnect & 28.02 & 0.9570 & 0.0424 & 28.02 & 0.9570 & 0.0424 & 23.23 & 0.9058 & 0.0927 & 23.23 & 0.9058 & 0.0927 \\
    \hline
    MAT & 29.14 & 0.9636 & 0.0290 & 29.14 & 0.9636 & 0.0290 & 23.77 & 0.9132 & 0.0708 & 23.77 & 0.9132 & 0.0708 \\
    \hline
    LaMa & 30.46 & 0.9685 & 0.0259 & 30.46 & 0.9685 & 0.0259 & 25.33 & 0.9263 & 0.0639 & 25.33 & \textbf{0.9263} & 0.0639 \\
    \hline
    Ours & \textbf{34.97} & \textbf{0.9773} & \textbf{0.0146} & \textbf{32.70} & \textbf{0.9690} & \textbf{0.0227} & \textbf{29.60} & \textbf{0.9404} & \textbf{0.0388} & \textbf{27.51} & 0.9227 & \textbf{0.0633} \\
    \hline
  \end{tabular}}
  \caption{Quantitative comparisons on the RealEstate10K dataset.}
  \label{tab:quantitative comparisons}
\end{table*}

\begin{table}
  \centering
  \resizebox{!}{!}{
  \begin{tabular}{c||c|c}
    \hline
    Model & Speed/FPS & Params/M \\
    \hline\hline
    FGVC & 0.50 & 5.26 \\
    \hline
    STTN & 1.62 & 16.57 \\
    \hline
    OPN & 1.73 & 3.42 \\
    \hline
    EdgeConnect & 3.16 & 67.11 \\
    \hline
    MAT & 4.52 & 61.56 \\
    \hline
    LaMa & 3.76 & 50.98 \\
    \hline
    Ours & \textbf{5.46} & 42.11 \\
    \hline
  \end{tabular}}
  \caption{Quantitative comparison of inference speed and model size.}
  \label{tab:time comparisons}
\end{table}

\subsection{Qualitative Comparison}
\cref{fig:qualitative comparison} shows the qualitative comparison results on RealEstate10K.
STTN and OPN can restore the approximate structure with the help of a small-view reference image.
However, these video inpainting algorithms are based on simple geometric transformations to align adjacent frames, as the viewpoint of the reference image changes more and more violently, it is more difficult to extract meaningful geometric information and resulting in blurred and confusing images.
OPN's results are slightly better than STTN because OPN can perform image inpainting with fewer reference images, whereas STTN relies on coherent multiple adjacent frames.
We found that when the scenario is relatively simple, MAT and LaMa can achieve good inpainting results, but when the mask blocks a complex scene, such as objects at different depths, the results of the single-image inpainting algorithm are not satisfactory, even for SOTA algorithms.
MAT often generates high-quality but strange scenes like structures that don't exist in nature.
LaMa prefers to generate simple structures to fill the hole region, this works in some cases, but the structure rationality of the whole image will be destroyed when the background is chaotic and complex.
Not to mention that LaMa-generated results become blurry when the mask becomes larger.
Compared with the above algorithms, the results generated by 3DFill are more credible and structurally consistent with the surrounding scenario.
3DFill makes full use of the geometrical information of the reference image, so that no obvious structural discontinuity appears in the inpainting result, making the image look natural and faithful to the target image.

\subsection{Quantitative Comparison}
\cref{tab:quantitative comparisons} shows the quantitative comparison on the RealEstate10K dataset.
We divided the test dataset into 4 types according to the mask size (\emph{Small-mask} and \emph{Large-mask}) and viewpoint variation of the reference image (\emph{Small-view} and \emph{Large-view}).
For the single-image inpainting algorithms---EdgeConnect, MAT and LaMa, the results are the same in \emph{Small-view} and \emph{Large-view}.
For the evaluation metrics, we use the PSNR, SSIM and LPIPS \cite{LPIPS} scores based on AlexNet \cite{AlexNet}.
The quantitative results showed that FGVC failed to restore the target images.
Although STTN and OPN achieved better results than FGVC, they still can't compete with the single-image SOTA algorithm.
MAT and LaMa can get better results in the case of \emph{Small-mask}.
Our 3DFill outperforms existing methods under various situations.

Considering the practicality of the algorithm, we compare the inference speed and model size of the above methods, the results are shown in \cref{tab:time comparisons}.
Although video inpainting models have a smaller number of parameters, their inference time is slower because they require multiple cycles to compute the features of all adjacent frames.
Our model has fewer parameters than single-image inpainting algorithms and has the fastest inference speed of more than 5 FPS.
This shows that our method does not depend on the complex design of the neural network, but relies on high-precision simulation of geometric relationship between images to achieve high-quality image inpainting results, which is more consistent with the laws of physics.

\begin{figure*}[t]
    \centering
    \begin{subfigure}{0.135\linewidth}
        \includegraphics[width=1\linewidth]{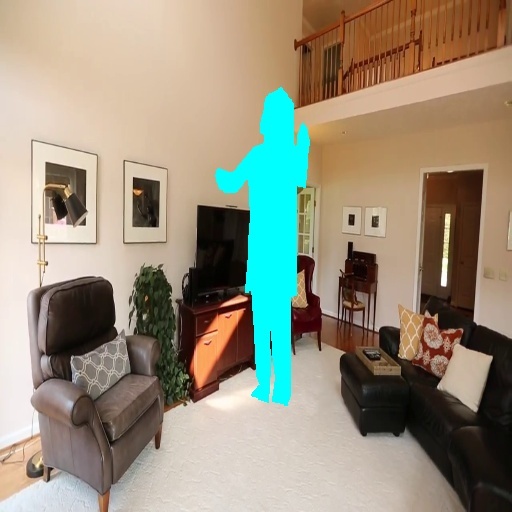}
    \end{subfigure}
    \begin{subfigure}{0.135\linewidth}
        \includegraphics[width=1\linewidth]{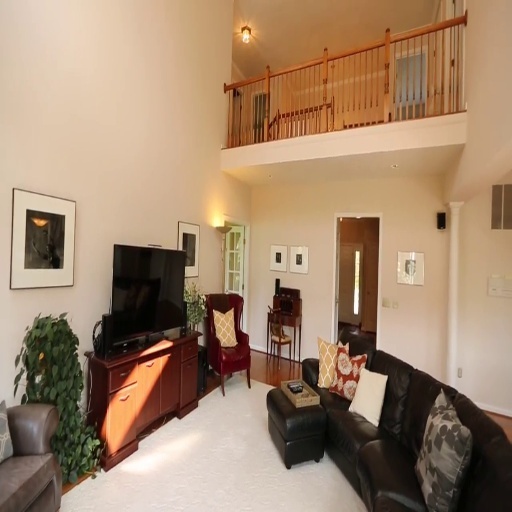}
    \end{subfigure}
    \hfill
    \begin{subfigure}{0.135\linewidth}
        \includegraphics[width=1\linewidth]{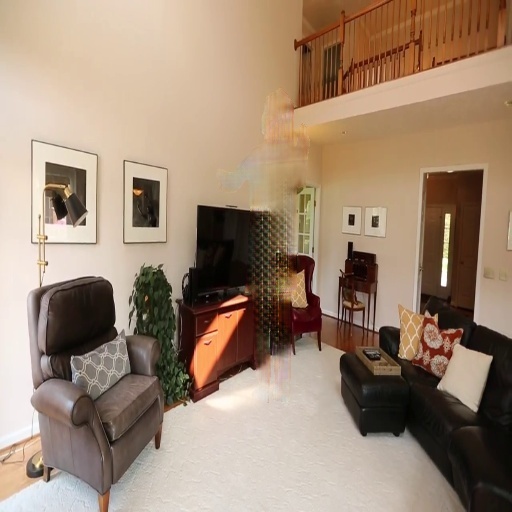}
    \end{subfigure}
    \begin{subfigure}{0.135\linewidth}
        \includegraphics[width=1\linewidth]{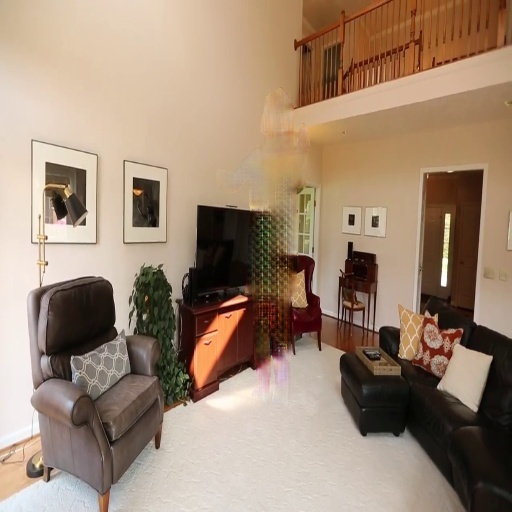}
    \end{subfigure}
    \begin{subfigure}{0.135\linewidth}
        \includegraphics[width=1\linewidth]{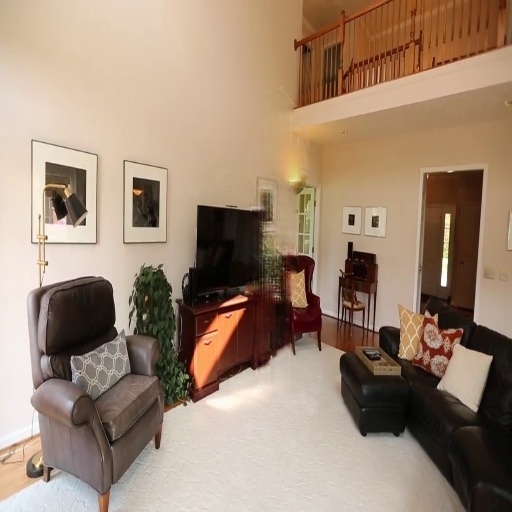}
    \end{subfigure}
    \begin{subfigure}{0.135\linewidth}
        \includegraphics[width=1\linewidth]{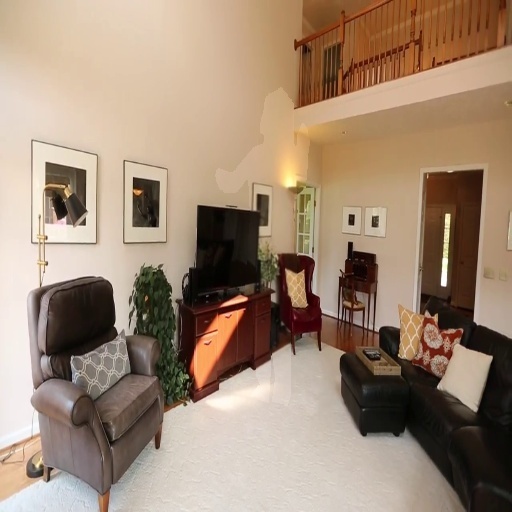}
    \end{subfigure}
    \begin{subfigure}{0.135\linewidth}
        \includegraphics[width=1\linewidth]{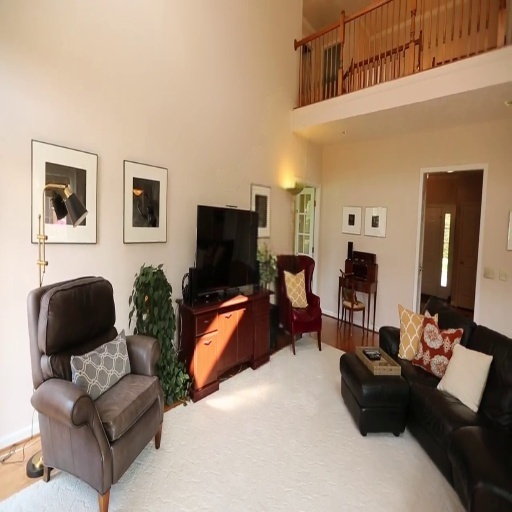}
    \end{subfigure}

    \centering
    \begin{subfigure}{0.135\linewidth}
        \includegraphics[width=1\linewidth]{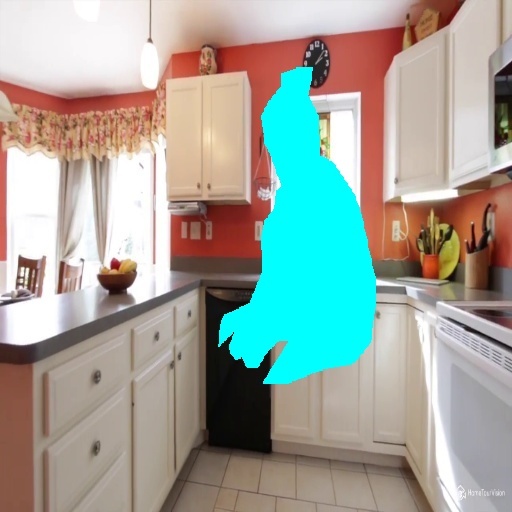}
        \caption{Target}
    \end{subfigure}
    \begin{subfigure}{0.135\linewidth}
        \includegraphics[width=1\linewidth]{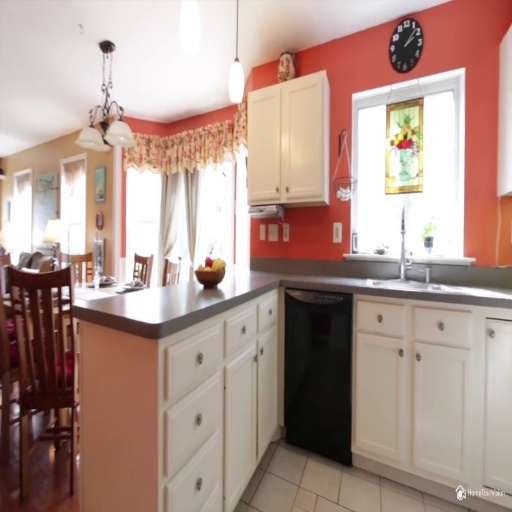}
        \caption{Reference}
    \end{subfigure}
    \hfill
    \begin{subfigure}{0.135\linewidth}
        \includegraphics[width=1\linewidth]{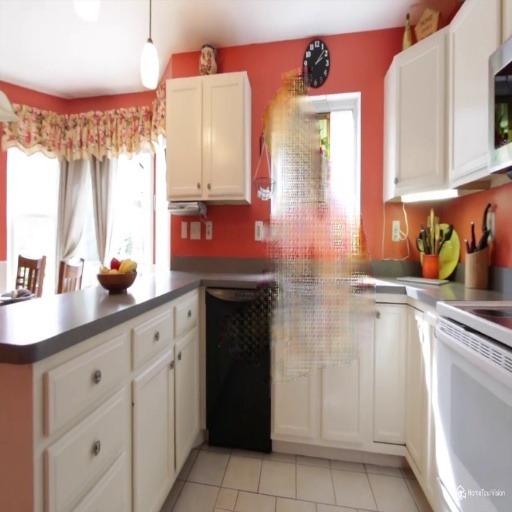}
        \caption{w/o IA}
    \end{subfigure}
    \begin{subfigure}{0.135\linewidth}
        \includegraphics[width=1\linewidth]{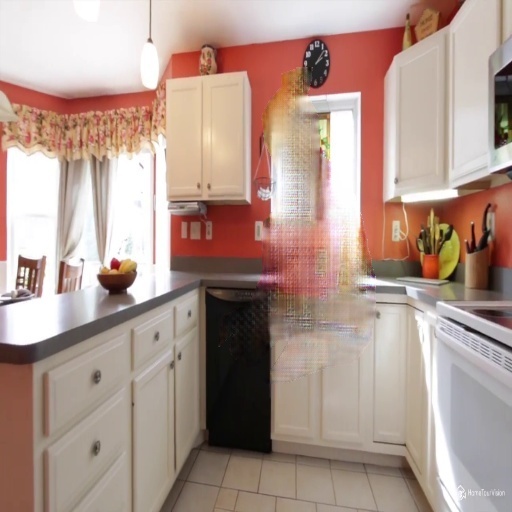}
        \caption{w/o 3D}
    \end{subfigure}
    \begin{subfigure}{0.135\linewidth}
        \includegraphics[width=1\linewidth]{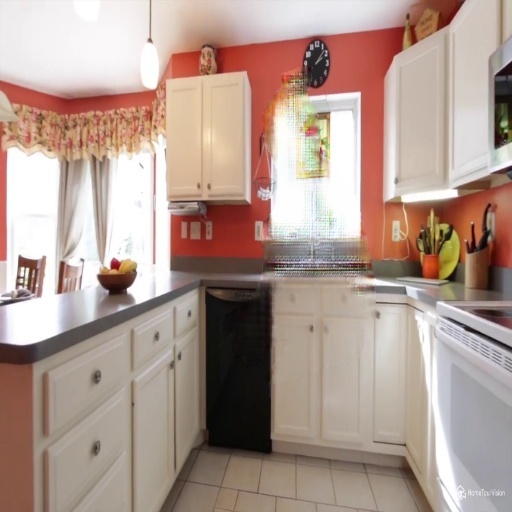}
        \caption{w/o 2D}
    \end{subfigure}
    \begin{subfigure}{0.135\linewidth}
        \includegraphics[width=1\linewidth]{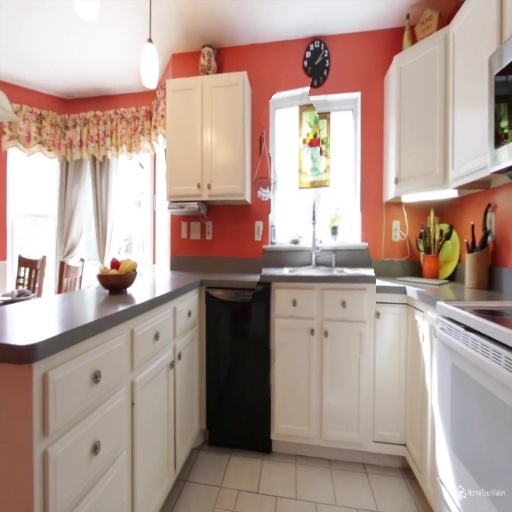}
        \caption{w/o IR}
    \end{subfigure}
    \begin{subfigure}{0.135\linewidth}
        \centering
        \includegraphics[width=1\linewidth]{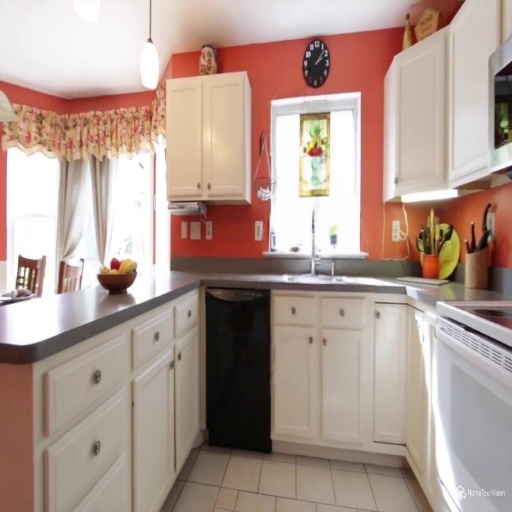}
        \caption{Ours}
    \end{subfigure}
    \caption{Visual results of ablation study.}
    \label{fig:ablation study}
\end{figure*}

\subsection{Ablation Study}
To evaluate the effects of modules in our method, we design the following ablations (IA module consists of 3D alignment module and 2D alignment module):

\begin{enumerate}
\item Removing the IA module (w/o IA);
\item Removing the 3D alignment module (w/o 3D);
\item Removing the 2D alignment module (w/o 2D);
\item Removing the IR module (w/o IR);.
\end{enumerate}

As shown in \cref{tab:ablation study} and \cref{fig:ablation study}, the IA module is the core of our method, without which we can hardly achieve the desired inpainting results.
In the IA module, 3D alignment plays a major role because it roughly aligns two images. 2D alignment is an auxiliary alignment.
The viewpoint change of the reference image has a significant impact on both 3D and 2D alignment.
The IR module has little effect on image inpainting for \emph{small-view}. 
However, when the viewpoint of the reference image becomes larger, the IA module alone cannot achieve the ideal alignment effect. 
In this case, the IR module plays an important role in refining the images.

\begin{table}
  \centering
  \resizebox{\linewidth}{!}{
  \begin{tabular}{c||c|c|c||c|c|c}
    \hline
      & \multicolumn{3}{|c||}{Small-view} &  \multicolumn{3}{|c}{Large-view} \\
    \hline
      & PSNR & SSIM & LPIPS & PSNR & SSIM & LPIPS \\
    \hline\hline
    w/o IA & 29.82 & 0.9572 & 0.0409 & 29.23 & 0.9540 & 0.0478 \\
    \hline
    w/o 3D & 31.68 & 0.9648	& 0.0291 & 29.91 & 0.9581 & 0.0407 \\
    \hline
    w/o 2D & 33.26 & 0.9709 & 0.0174 & 31.48 & 0.9640 & 0.0277 \\
    \hline
    w/o IR & 33.72 & 0.9752 & 0.0152 & 30.08 & 0.9623 & 0.0267 \\
    \hline
    Ours & \textbf{34.97} & \textbf{0.9773}	& \textbf{0.0146} & \textbf{32.70} & \textbf{0.9690} & \textbf{0.0227} \\
    \hline
  \end{tabular}}
  \caption{Ablation study on each module.}
  \label{tab:ablation study}
\end{table}

\subsection{Robustness Study}
\label{sec:Robustness Study}
3DFill uses fixed camera intrinsic parameters $K$ to achieve 3D projection because we can't get ground-truth camera parameters in most cases.
The camera intrinsic parameter contains 5 variables: the focal length $f$, the pixel length $d_{x}$ and $d_{y}$, the pixel coordinates of the origin of the image coordinate system $u_{0}$ and $v_{0}$.
The parameter matrix $K$ is:
\[
	\begin{bmatrix}
      \frac{f}{d_{x}} & 0 & u_{0}\\
      0 & \frac{f}{d_{y}} & v_{0}\\
      0 & 0 & 1
	\end{bmatrix}
	\]
The origin of the image coordinate system is in the center of the image so we can set the $u_{0}$ and $v_{0}$ be half the resolution of the image.
Generally, the pixel lengths in the $x$ and $y$ directions are the same, so $d_{x}$ and $d_{y}$ can be standardized to take pixels as units, and the proportional relationship between pixels and the real length is transferred to $f$.
So the only important variable is the focal length $f$.
In this paper we set the focal length of all images to 750 pixels which is a common number.
We test the robustness of 3DFill to focal length as shown in \cref{tab:robustness experiment}.
The experimental results show that different focal lengths have little effect on 3DFill.
We believe that the robustness of 3DFill to the focal length is that the intrinsic parameters of the camera are used in the forward projection and the inverse projection respectively during the projection process, as shown in \cref{eq:entire eq}, so that the scale variation of the 3D scene caused by different intrinsic parameters is largely offset.
This can be considered as an advantage of 3D projection.
\begin{table}
  \centering
  \resizebox{\linewidth}{!}{
  \begin{tabular}{c||c|c|c||c|c|c}
    \hline
      & \multicolumn{3}{|c||}{Small-view} &  \multicolumn{3}{|c}{Large-view} \\
    \hline
    Focal Length & PSNR & SSIM & LPIPS & PSNR & SSIM & LPIPS \\
    \hline\hline
    1050 & 34.40 & 0.9749 & 0.0160 & 32.03 & 0.9663 & 0.0265 \\
    \hline
    900 & 34.64 & 0.9761 & 0.0154 & 32.29 & 0.9674 & 0.0249 \\
    \hline
    750 & \textbf{34.97} & \textbf{0.9773} & \textbf{0.0146} & \textbf{32.70} & \textbf{0.9690} & \textbf{0.0227} \\
    \hline
    600 & 34.84 & 0.9769 & 0.0149 & 32.44 & 0.9681 & 0.0242 \\
    \hline
    450 & 34.19 & 0.9743 & 0.0169 & 31.72 & 0.9652 & 0.0283 \\
    \hline
  \end{tabular}}
  \caption{Robustness study with different focal lengths (pixels).}
  \label{tab:robustness experiment}
\end{table}
\section{Conclusion}
\label{sec:conclusion}
We propose 3DFill---a novel reference-guided image inpainting method.
To maximize the use of information from the reference image, 3DFill utilizes a two-stage image alignment module to align the reference image with high accuracy and utilizes a simple image refinement module to further refine the alignment result.
The inpainting results outperform state-of-the-art single-image and reference-guided image inpainting methods, especially when the hole is large or contains complicated scenes.

3DFill still has some limitations.
First, the image alignment algorithm is not suited for image pairs with extreme viewpoint changes.
Second, 3DFill uses the pre-trained DPT for depth estimation, which makes the accuracy of 3D projection dependent to some extent on the performance of DPT.
Third, due to perspective, errors in depth estimation and lens distortion, the alignment effect of the image edge area is worse than that of the center area.
When the hole region involve image edges and contain complex scenes, the image inpainting result is poor.

{\small
\bibliographystyle{ieee_fullname}
\bibliography{egbib}
}

\clearpage
\appendix

{\centering\section*{Appendices}}
\section{3D Alignment in Detail}
\label{sec:3D alignment}
In this section, we introduce the 3D alignment algorithm in detail.
\cref{fig:3D projection} shows the process of implementing 3D alignment.

Step 1: we reverse the projection process according to the camera imaging principle to obtain the 3D scene.
The coordinates of the 3D scene corresponding to each pixel on the target image can be calculated by:
\begin{equation}
  \left[\begin{array}{c}
      X_{t}\\
      Y_{t}\\
      Z_{t}
      \end{array}
  \right]
  =  Z_{t} \cdot 
  \left[\begin{array}{ccc}
      \frac{f}{d_{x}} & 0 & u_{0}\\
      0 & \frac{f}{d_{y}} & v_{0}\\
      0 & 0 & 1
      \end{array}
  \right]^{-1}
  \times
  \left[\begin{array}{c}
      u_{t}\\
      v_{t}\\
      1
      \end{array}
  \right]
  \label{eq:inverse projection}
\end{equation}
$u_{t},v_{t}$ indicates the pixel coordinates on the target image.
$X_{t},Y_{t}, Z_{t}$ are the coordinates of the 3D scene from the target viewpoint and $Z_{t}$ is the depth estimate of the DPT.
The first matrix on the right side of the equation is the inverse of the camera intrinsic matrix, where $f$ is the focal length, $d_{x}$ and $d_{y}$ indicate the pixel lengths in the $x$ and $y$ directions of the image respectively, $u_{0}$ and $v_{0}$ indicates the coordinates of the origin of the image coordinate system in the pixel coordinate system.
Briefly speaking, this process achieves a 2D to 3D projection with the help of the depth map and camera's intrinsic parameters.

Step 2: we convert the scene coordinates from the target viewpoint to the reference viewpoint by 3D transformation.
3D transformation contains 6 degrees of freedom so we use Pose3DNet to predict a vector with 6 elements corresponding to 3 translation variables $t_{x}$, $t_{y}$, $t_{z}$ and 3 rotation variables (Euler angles) $\alpha$, $\beta$, $\gamma$.
We compose the translation variables into the vector $t$ and convert three Euler angles to a rotation matrix $R$.
We can get the coordinates in the reference viewpoint by:
\begin{equation}
    \left[\begin{array}{c}
        X_{r}\\
        Y_{r}\\
        Z_{r}\\
        1
        \end{array}
    \right]
    =
    \left[\begin{array}{cc}
        R & t\\
        0^{T} & 1
        \end{array}
    \right]
    \times
    \left[\begin{array}{c}
        X_{t}\\
        Y_{t}\\
        Z_{t}\\
        1
        \end{array}
    \right]
    \label{eq:3D transform}
\end{equation}

Step 3: we use the forward process of camera imaging to project the 3D scene onto the 2D image at the reference viewpoint by:
\begin{equation}
  \left[\begin{array}{c}
      u_{r}\\
      v_{r}\\
      1
      \end{array}
  \right]
  =
  \frac{1}{Z_{r}} \cdot 
  \left[\begin{array}{ccc}
      \frac{f}{d_{x}} & 0 & u_{0}\\
      0 & \frac{f}{d_{y}} & v_{0}\\
      0 & 0 & 1
      \end{array}
  \right]
  \times
  \left[\begin{array}{c}
      X_{r}\\
      Y_{r}\\
      Z_{r}
      \end{array}
  \right]
  \label{eq:forward projection}
\end{equation}
So far, we can get the coordinates $P_{r} (u_{r}, v_{r})$ of each pixel $ P_{t} (u_{t}, v_{t})$ on the target image projected on the reference image.
The above projection process can be simplified by:
\begin{equation}
  P_{r}
  =
  \frac{Z_{t}}{Z_{r}} K T_{t \rightarrow r} K^{-1} P_{t}
  \label{eq:new entire eq}
\end{equation}
$K$ is the intrinsic camera matrix, $T_{t \rightarrow r}$ is the camera transformation matrix.

\section{Model Architecture}
\label{sec:architecture}
\begin{figure*}[t]
    \centering
    \includegraphics[width=1\linewidth]{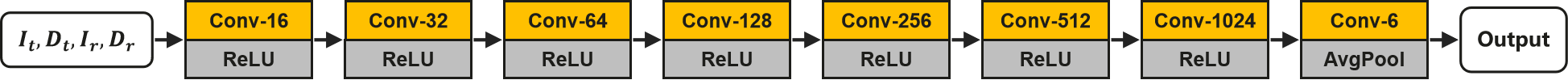}
    \caption{Model architecture of Pose3DNet. $D_{t}$ and $D_{r}$ are the depth maps of $I_{t}$ and $I_{r}$ estimated by DPT.}
    \label{fig:Pose3DNet architecture}
\end{figure*}

\begin{figure*}[t]
    \centering
    \includegraphics[width=1\linewidth]{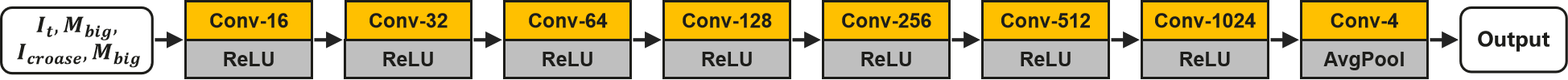}
    \caption{Model architecture of Pose2DNet.}
    \label{fig:Pose2DNet architecture}
\end{figure*}

\begin{figure*}[t]
    \centering
    \includegraphics[width=1\linewidth]{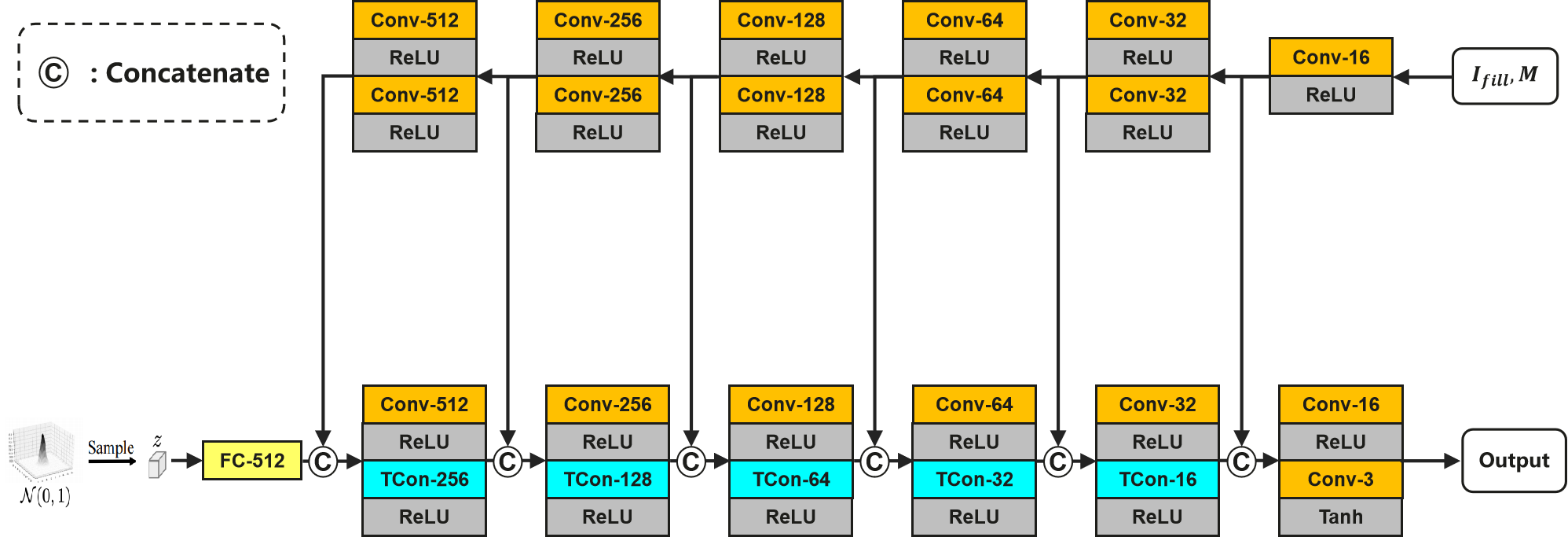}
    \caption{Model architecture of InpaintNet.}
    \label{fig:InpaintNet architecture}
\end{figure*}

In this section, we show the detailed architecture of the models used by 3DFill.
Both the Pose3DNet and Pose2DNet are encoders as shown in \cref{fig:Pose3DNet architecture}, \cref{fig:Pose2DNet architecture}, and the InpaintNet is a CGAN as shown in \cref{fig:InpaintNet architecture}.

\section{Additional Visual Results}
\label{sec:results}
\begin{figure*}[t]
    \centering
    \begin{subfigure}{0.2\linewidth}
        \includegraphics[width=1\linewidth]{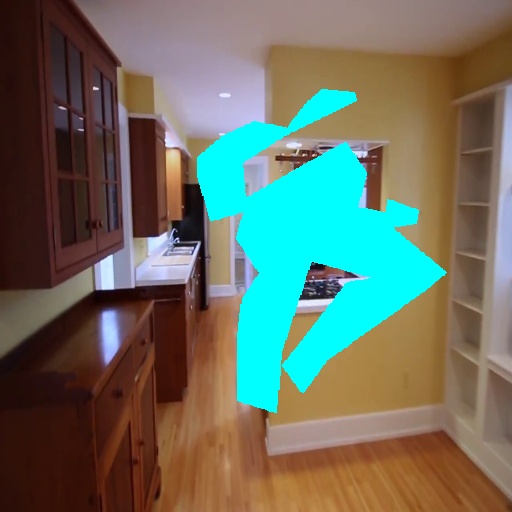}
    \end{subfigure}
    \begin{subfigure}{0.2\linewidth}
        \includegraphics[width=1\linewidth]{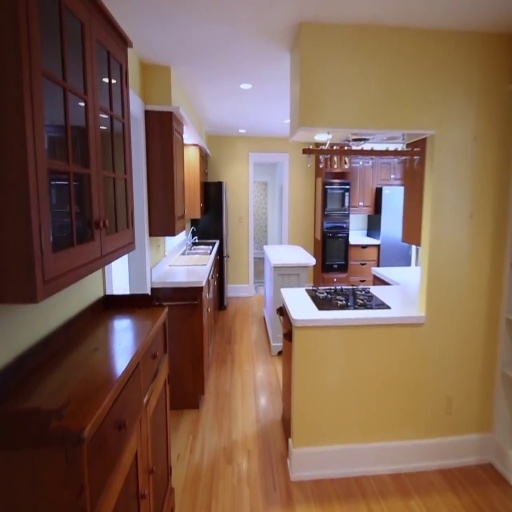}
    \end{subfigure}
    \begin{subfigure}{0.2\linewidth}
        \includegraphics[width=1\linewidth]{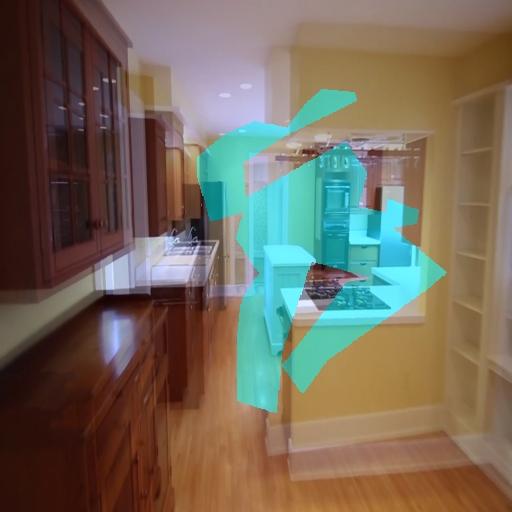}
    \end{subfigure}
    \begin{subfigure}{0.2\linewidth}
        \includegraphics[width=1\linewidth]{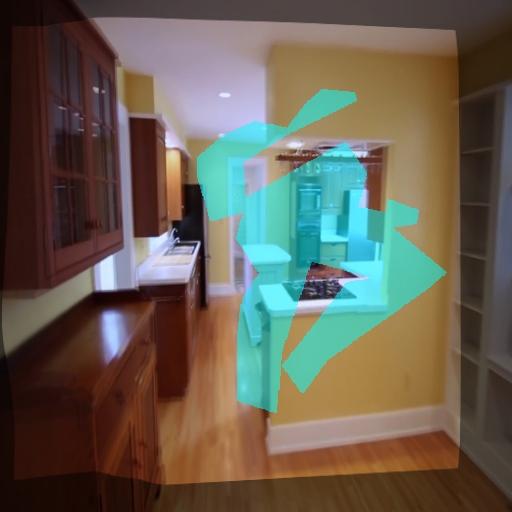}
    \end{subfigure}

    \centering
    \begin{subfigure}{0.2\linewidth}
        \includegraphics[width=1\linewidth]{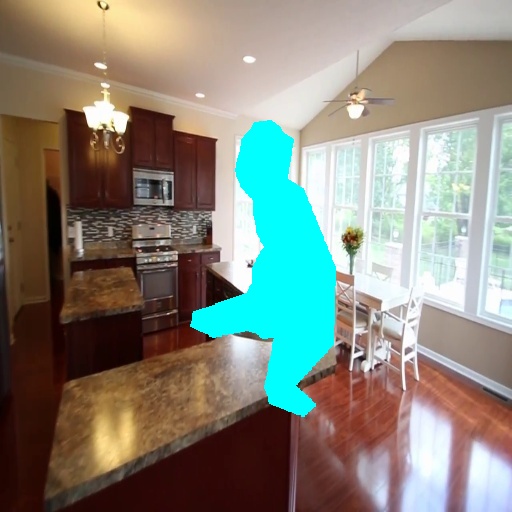}
    \end{subfigure}
    \begin{subfigure}{0.2\linewidth}
        \includegraphics[width=1\linewidth]{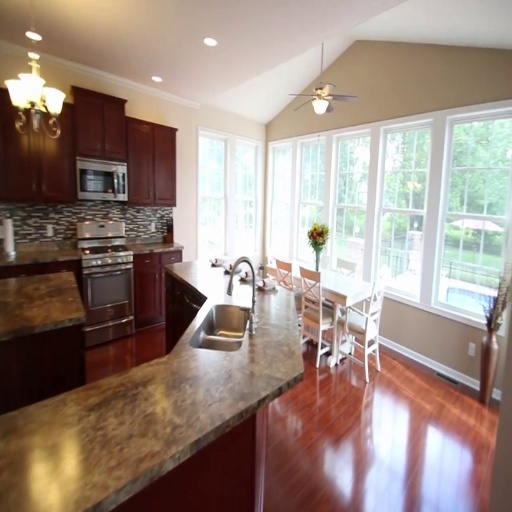}
    \end{subfigure}
    \begin{subfigure}{0.2\linewidth}
        \includegraphics[width=1\linewidth]{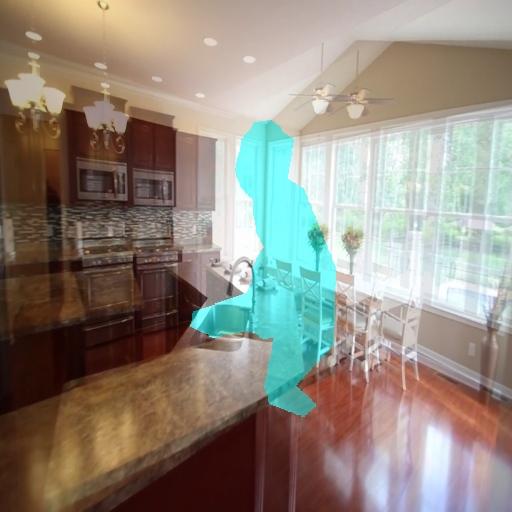}
    \end{subfigure}
    \begin{subfigure}{0.2\linewidth}
        \includegraphics[width=1\linewidth]{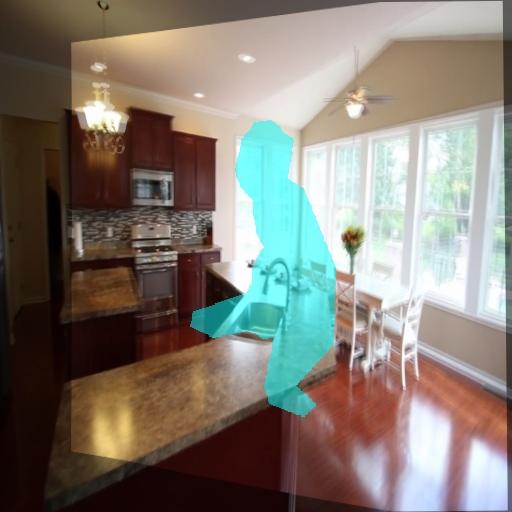}
    \end{subfigure}
    
    \centering
    \begin{subfigure}{0.2\linewidth}
        \includegraphics[width=1\linewidth]{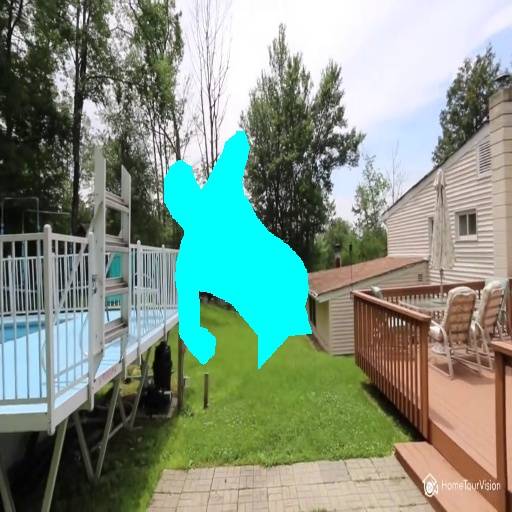}
    \end{subfigure}
    \begin{subfigure}{0.2\linewidth}
        \includegraphics[width=1\linewidth]{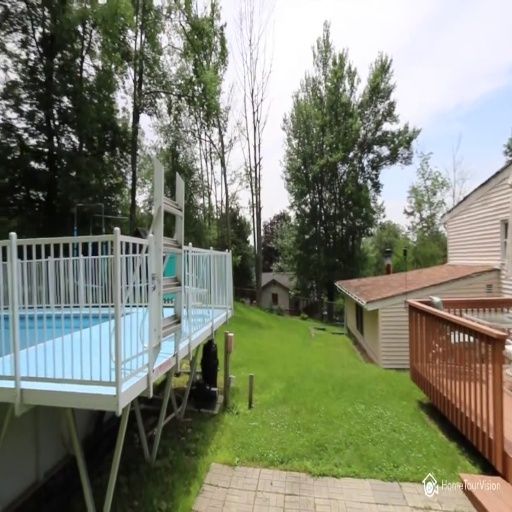}
    \end{subfigure}
    \begin{subfigure}{0.2\linewidth}
        \includegraphics[width=1\linewidth]{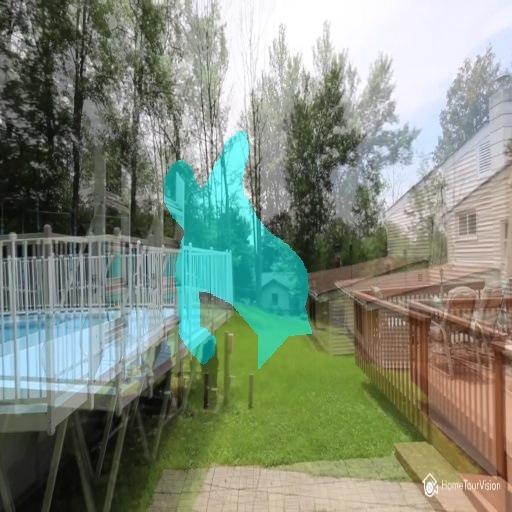}
    \end{subfigure}
    \begin{subfigure}{0.2\linewidth}
        \includegraphics[width=1\linewidth]{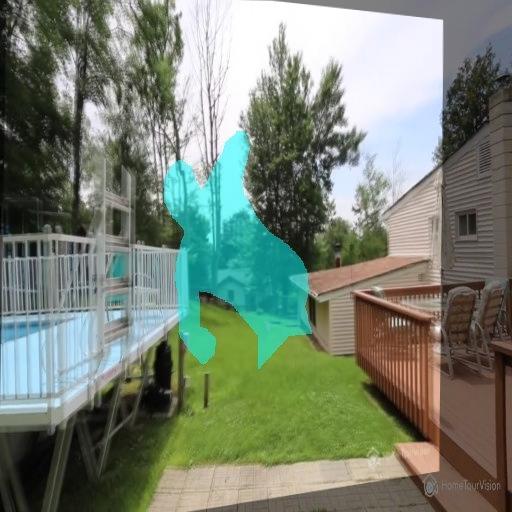}
    \end{subfigure}

    \centering
    \begin{subfigure}{0.2\linewidth}
        \includegraphics[width=1\linewidth]{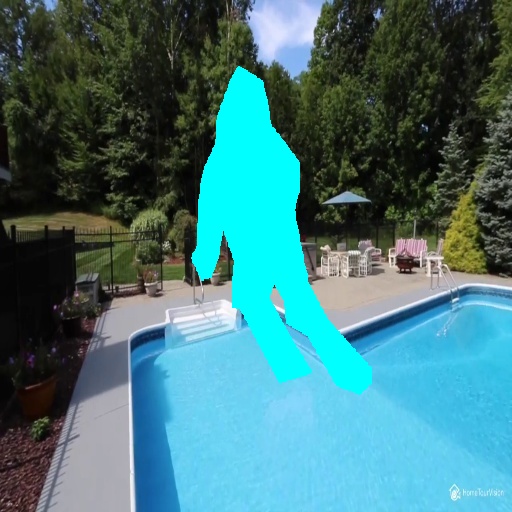}
    \end{subfigure}
    \begin{subfigure}{0.2\linewidth}
        \includegraphics[width=1\linewidth]{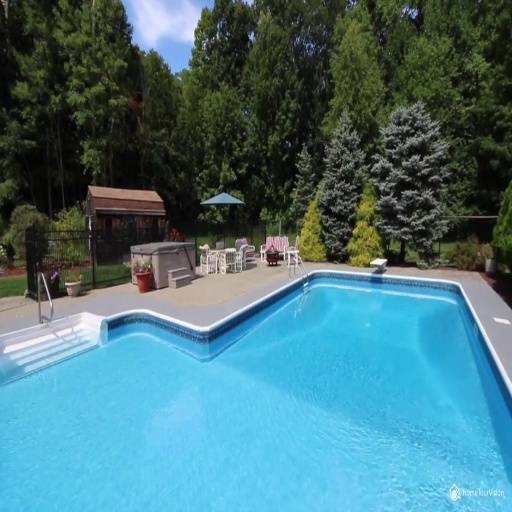}
    \end{subfigure}
    \begin{subfigure}{0.2\linewidth}
        \includegraphics[width=1\linewidth]{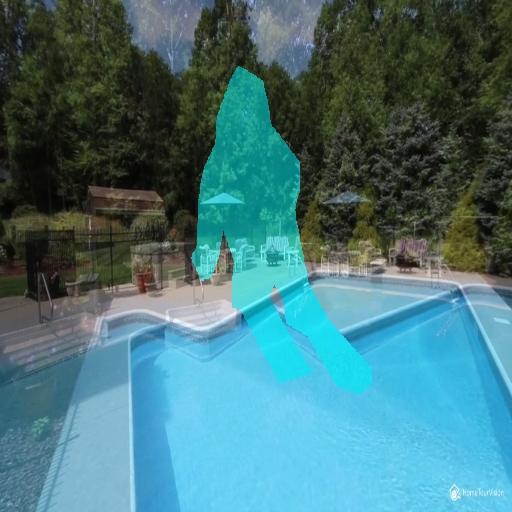}
    \end{subfigure}
    \begin{subfigure}{0.2\linewidth}
        \includegraphics[width=1\linewidth]{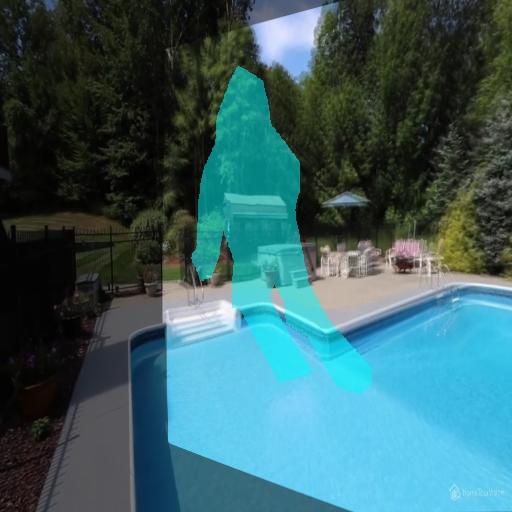}
    \end{subfigure}

    \centering
    \begin{subfigure}{0.2\linewidth}
        \includegraphics[width=1\linewidth]{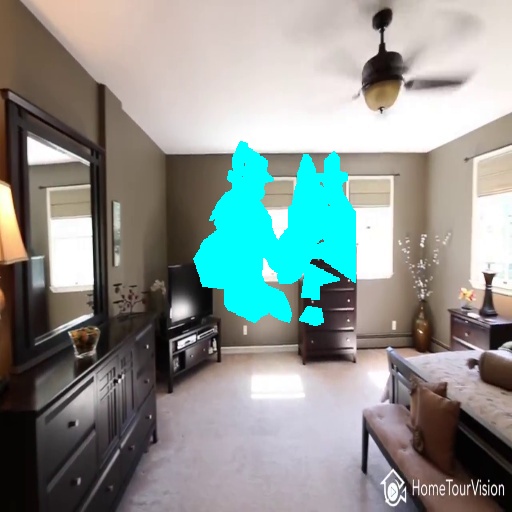}
    \end{subfigure}
    \begin{subfigure}{0.2\linewidth}
        \includegraphics[width=1\linewidth]{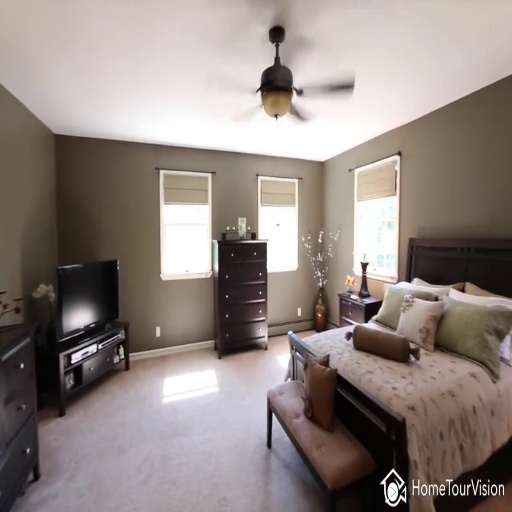}
    \end{subfigure}
    \begin{subfigure}{0.2\linewidth}
        \includegraphics[width=1\linewidth]{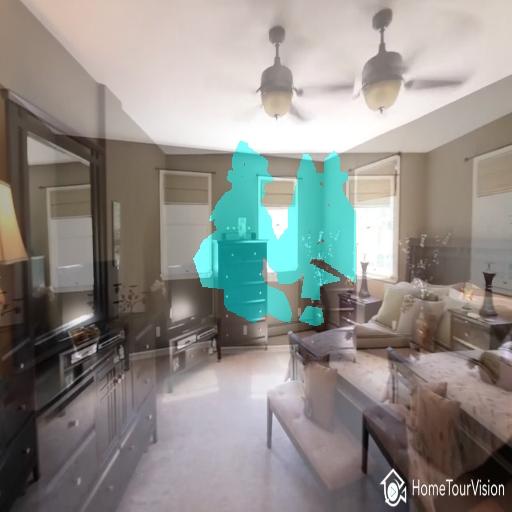}
    \end{subfigure}
    \begin{subfigure}{0.2\linewidth}
        \includegraphics[width=1\linewidth]{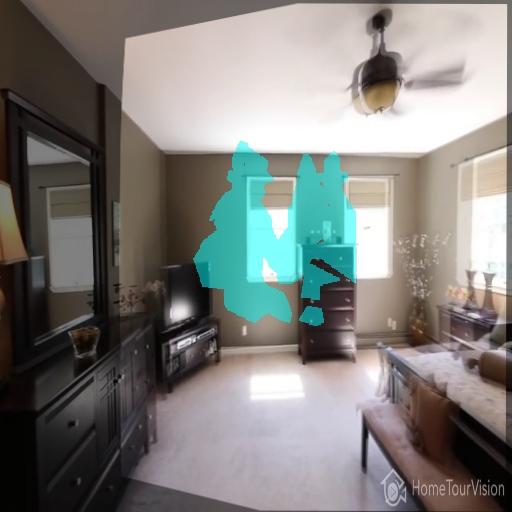}
    \end{subfigure}

    \centering
    \begin{subfigure}{0.2\linewidth}
        \includegraphics[width=1\linewidth]{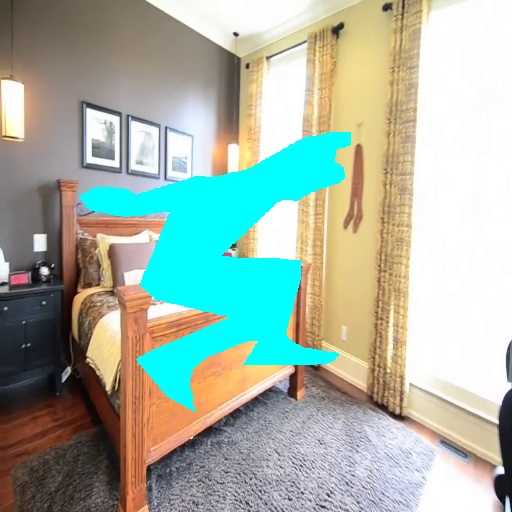}
        \caption{Target}
    \end{subfigure}
    \begin{subfigure}{0.2\linewidth}
        \includegraphics[width=1\linewidth]{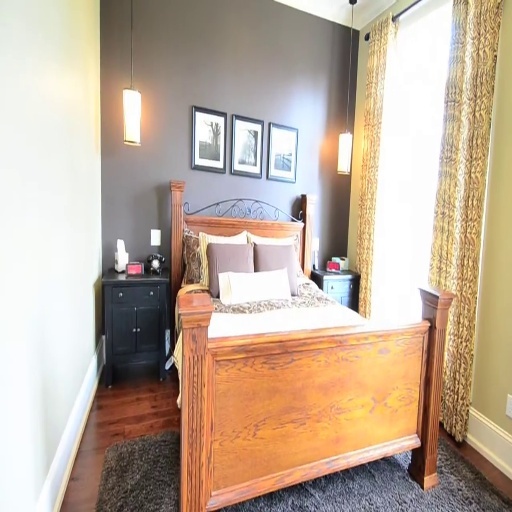}
        \caption{Reference}
    \end{subfigure}
    \begin{subfigure}{0.2\linewidth}
        \includegraphics[width=1\linewidth]{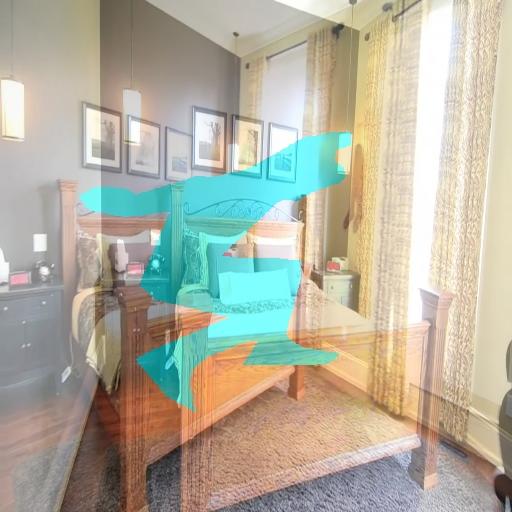}
        \caption{Before alignment}
    \end{subfigure}
    \begin{subfigure}{0.2\linewidth}
        \includegraphics[width=1\linewidth]{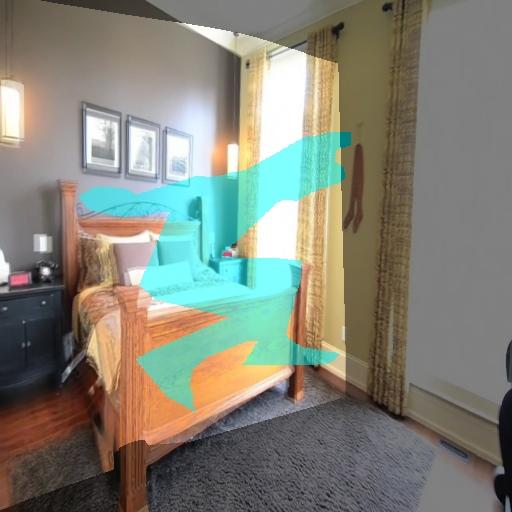}
        \caption{After alignment}
    \end{subfigure}
    
    \caption{Image alignment results.}
    \label{fig:alignment results}
\end{figure*}

\begin{figure*}[t]
    \centering
    \begin{subfigure}{0.16\linewidth}
        \includegraphics[width=1\linewidth]{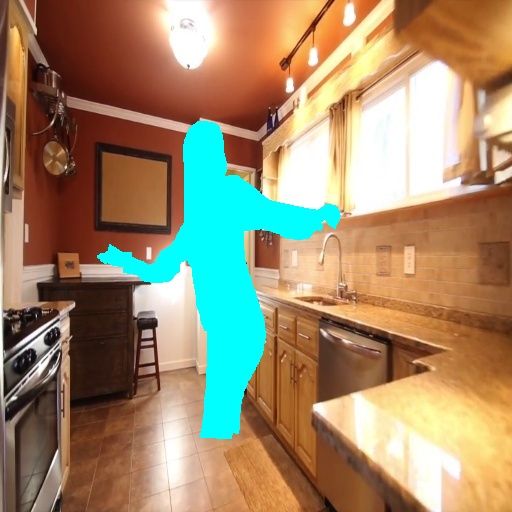}
    \end{subfigure}
    \begin{subfigure}{0.16\linewidth}
        \includegraphics[width=1\linewidth]{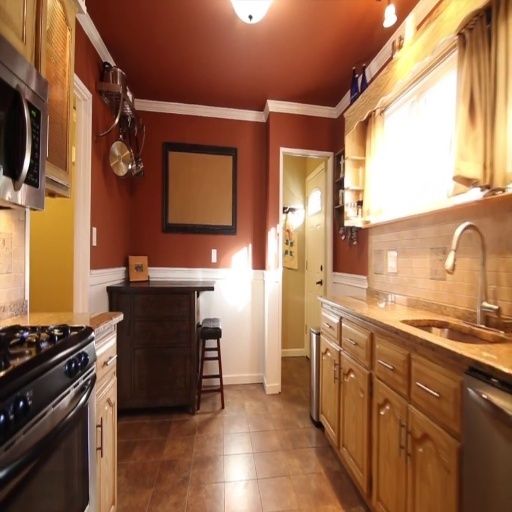}
    \end{subfigure}
    \hfill
    \begin{subfigure}{0.16\linewidth}
        \includegraphics[width=1\linewidth]{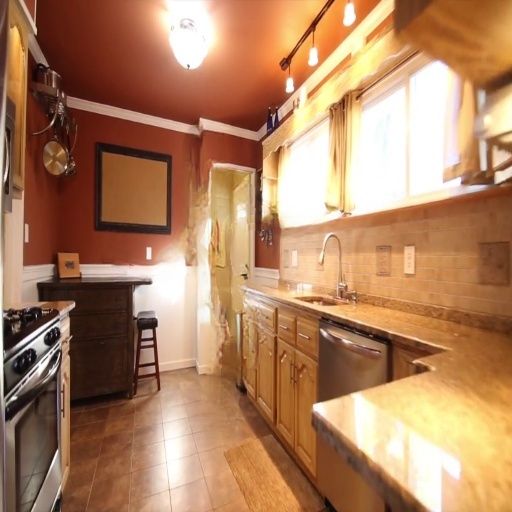}
    \end{subfigure}
    \begin{subfigure}{0.16\linewidth}
        \includegraphics[width=1\linewidth]{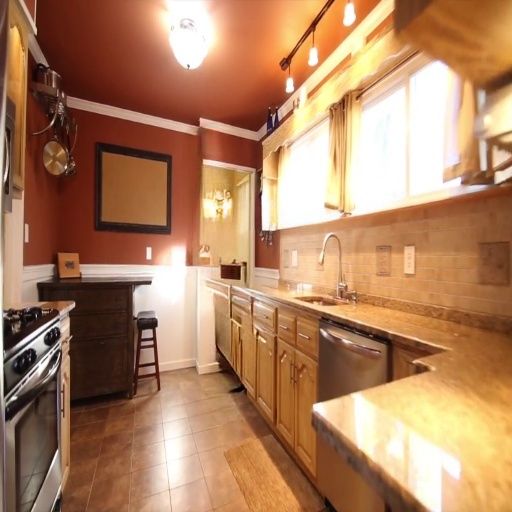}
    \end{subfigure}
    \begin{subfigure}{0.16\linewidth}
        \includegraphics[width=1\linewidth]{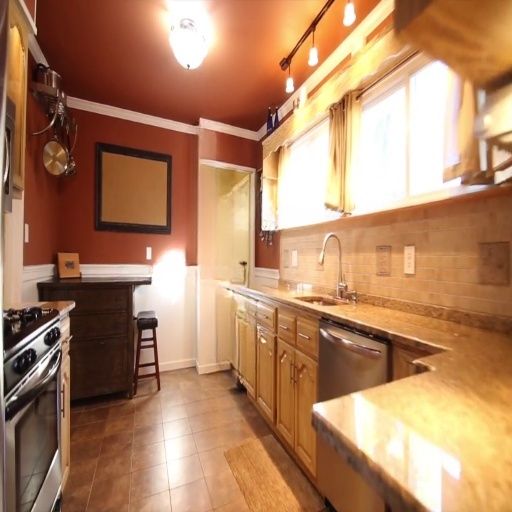}
    \end{subfigure}
    \begin{subfigure}{0.16\linewidth}
        \includegraphics[width=1\linewidth]{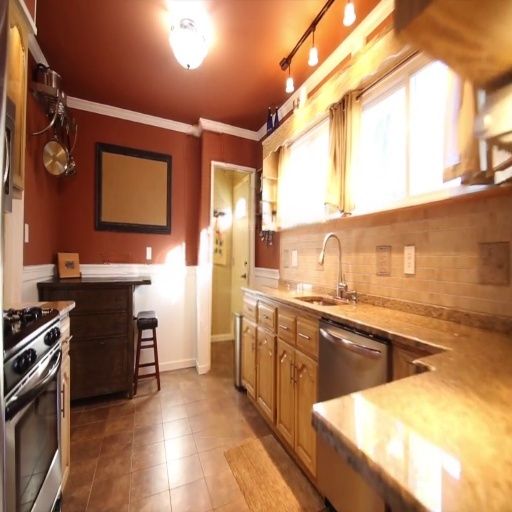}
    \end{subfigure}
    
    \centering
    \begin{subfigure}{0.16\linewidth}
        \includegraphics[width=1\linewidth]{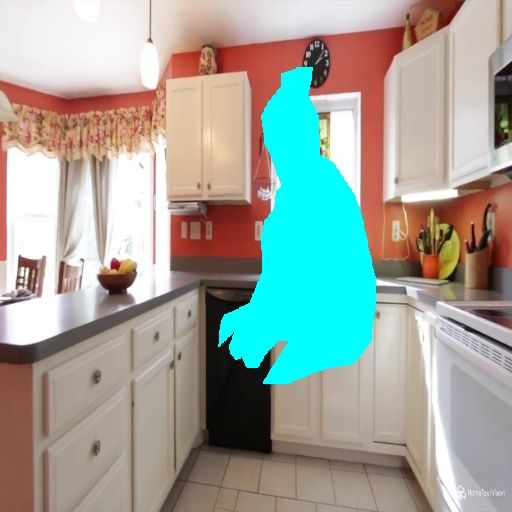}
    \end{subfigure}
    \begin{subfigure}{0.16\linewidth}
        \includegraphics[width=1\linewidth]{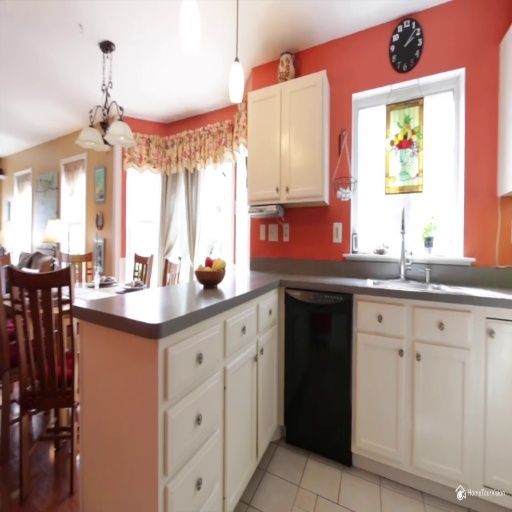}
    \end{subfigure}
    \hfill
    \begin{subfigure}{0.16\linewidth}
        \includegraphics[width=1\linewidth]{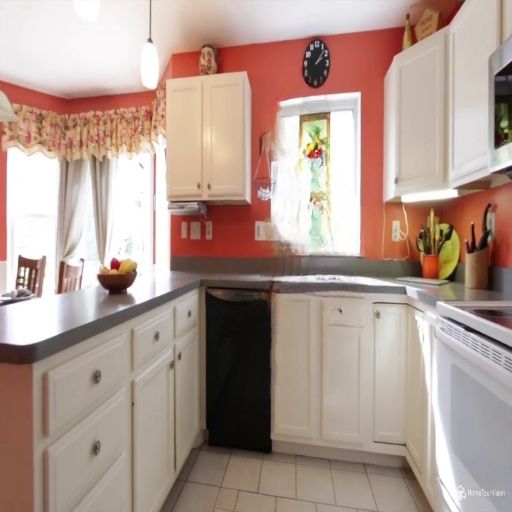}
    \end{subfigure}
    \begin{subfigure}{0.16\linewidth}
        \includegraphics[width=1\linewidth]{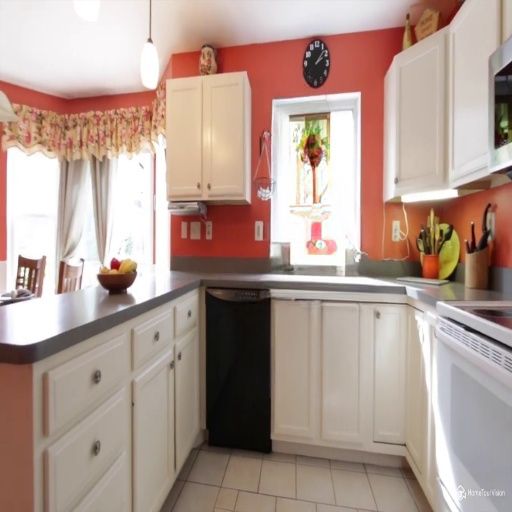}
    \end{subfigure}
    \begin{subfigure}{0.16\linewidth}
        \includegraphics[width=1\linewidth]{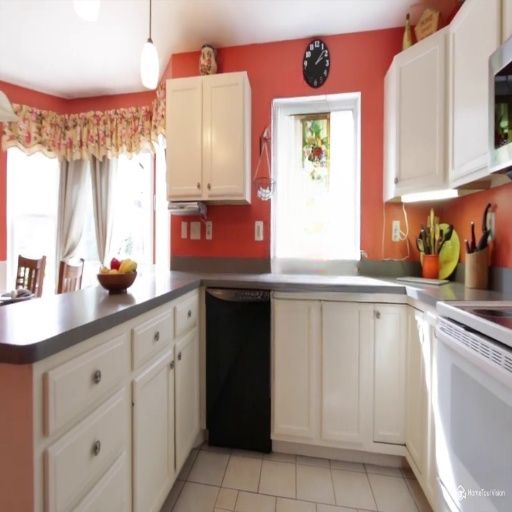}
    \end{subfigure}
    \begin{subfigure}{0.16\linewidth}
        \includegraphics[width=1\linewidth]{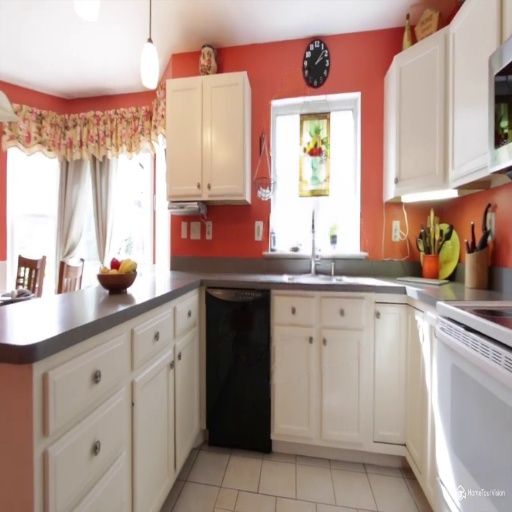}
    \end{subfigure}
    
    \centering
    \begin{subfigure}{0.16\linewidth}
        \includegraphics[width=1\linewidth]{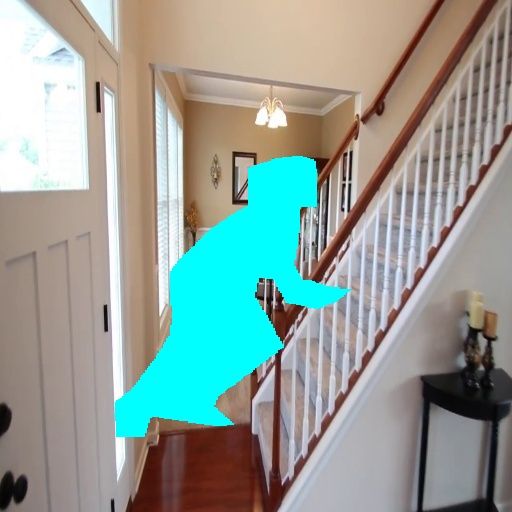}
    \end{subfigure}
    \begin{subfigure}{0.16\linewidth}
        \includegraphics[width=1\linewidth]{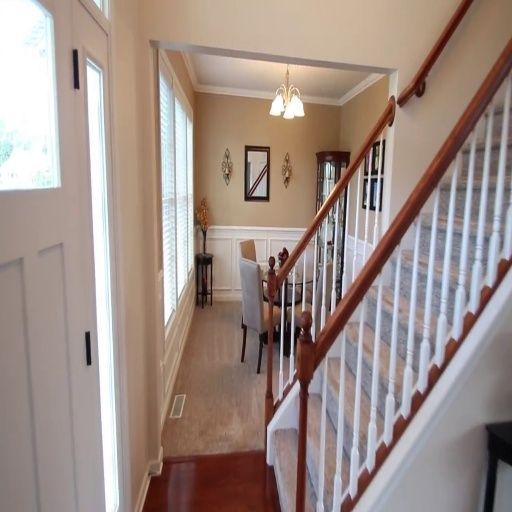}
    \end{subfigure}
    \hfill
    \begin{subfigure}{0.16\linewidth}
        \includegraphics[width=1\linewidth]{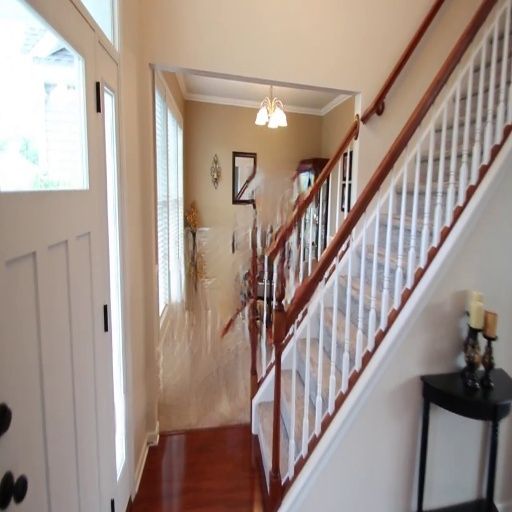}
    \end{subfigure}
    \begin{subfigure}{0.16\linewidth}
        \includegraphics[width=1\linewidth]{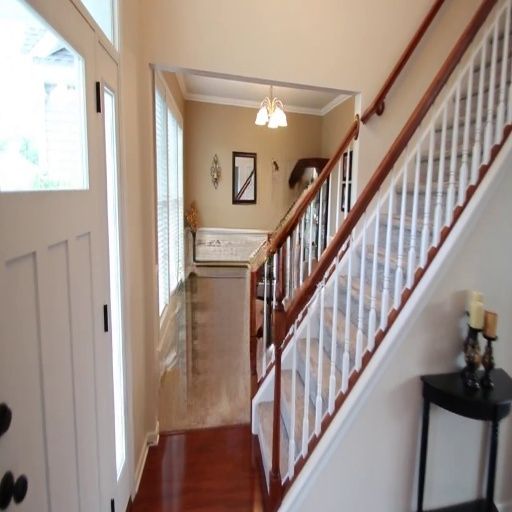}
    \end{subfigure}
    \begin{subfigure}{0.16\linewidth}
        \includegraphics[width=1\linewidth]{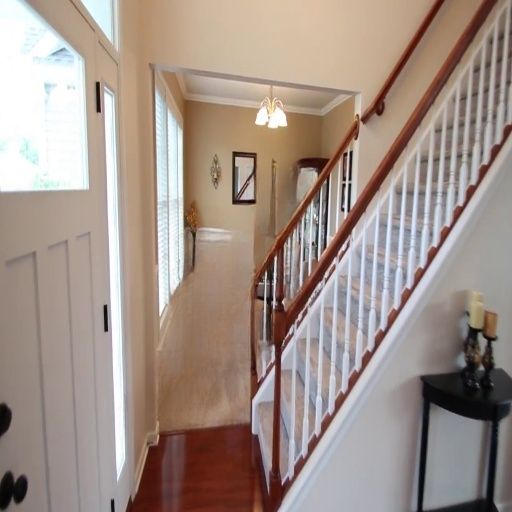}
    \end{subfigure}
    \begin{subfigure}{0.16\linewidth}
        \includegraphics[width=1\linewidth]{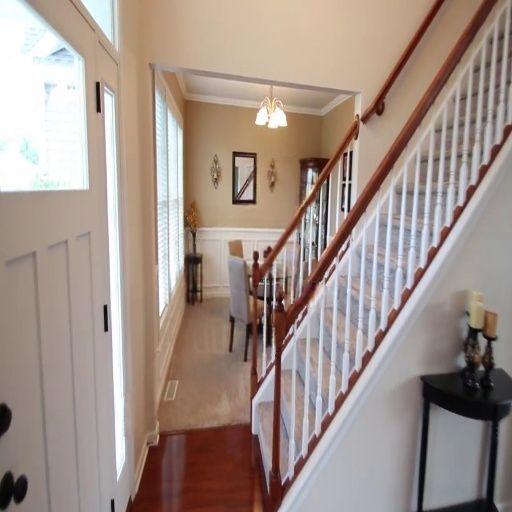}
    \end{subfigure}

    \centering
    \begin{subfigure}{0.16\linewidth}
        \includegraphics[width=1\linewidth]{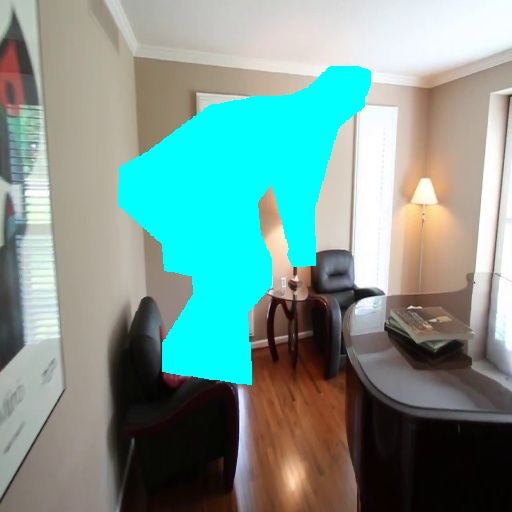}
    \end{subfigure}
    \begin{subfigure}{0.16\linewidth}
        \includegraphics[width=1\linewidth]{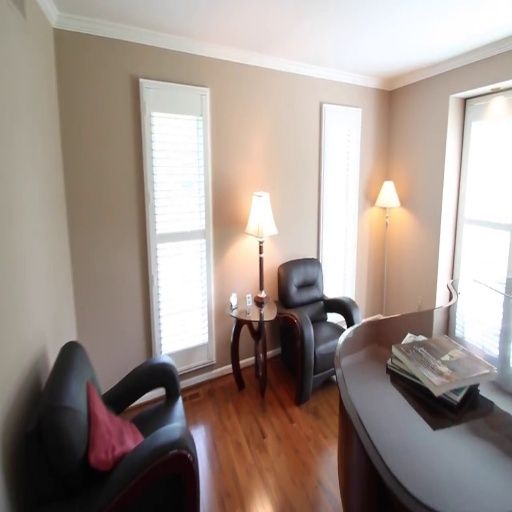}
    \end{subfigure}
    \hfill
    \begin{subfigure}{0.16\linewidth}
        \includegraphics[width=1\linewidth]{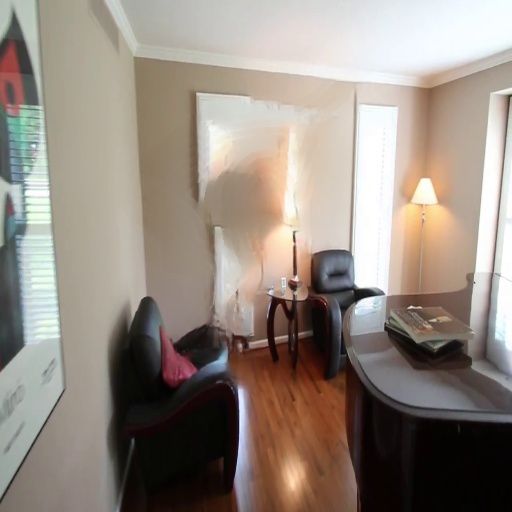}
    \end{subfigure}
    \begin{subfigure}{0.16\linewidth}
        \includegraphics[width=1\linewidth]{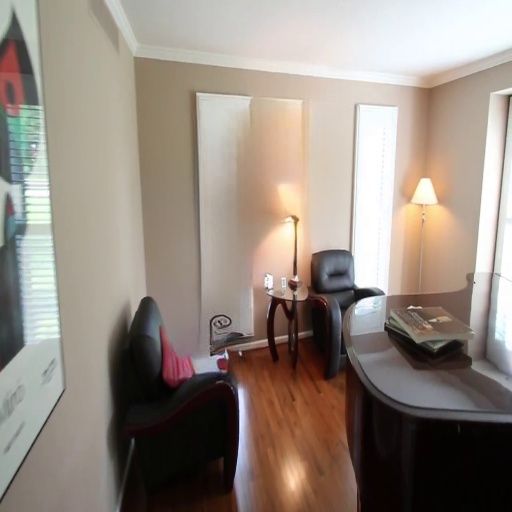}
    \end{subfigure}
    \begin{subfigure}{0.16\linewidth}
        \includegraphics[width=1\linewidth]{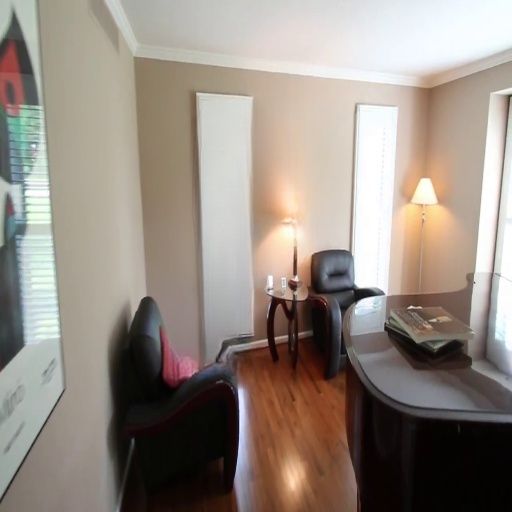}
    \end{subfigure}
    \begin{subfigure}{0.16\linewidth}
        \includegraphics[width=1\linewidth]{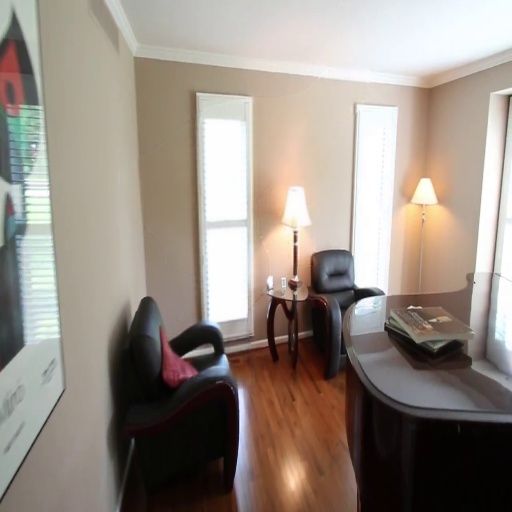}
    \end{subfigure}

    \centering
    \begin{subfigure}{0.16\linewidth}
        \includegraphics[width=1\linewidth]{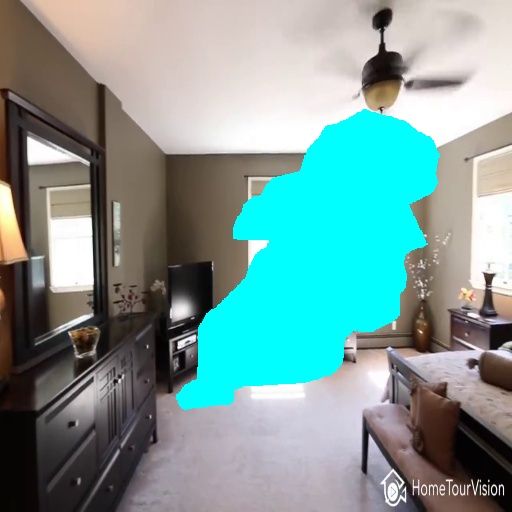}
    \end{subfigure}
    \begin{subfigure}{0.16\linewidth}
        \includegraphics[width=1\linewidth]{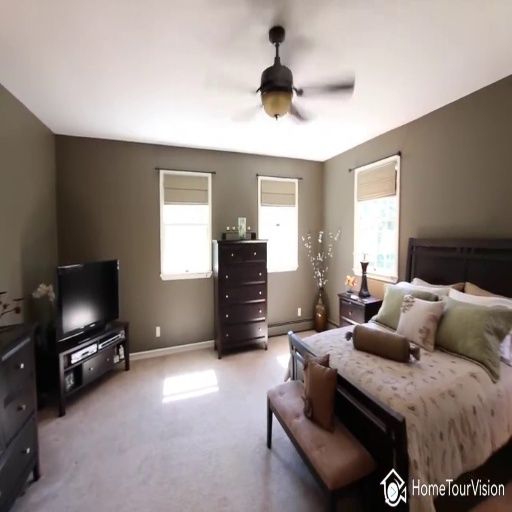}
    \end{subfigure}
    \hfill
    \begin{subfigure}{0.16\linewidth}
        \includegraphics[width=1\linewidth]{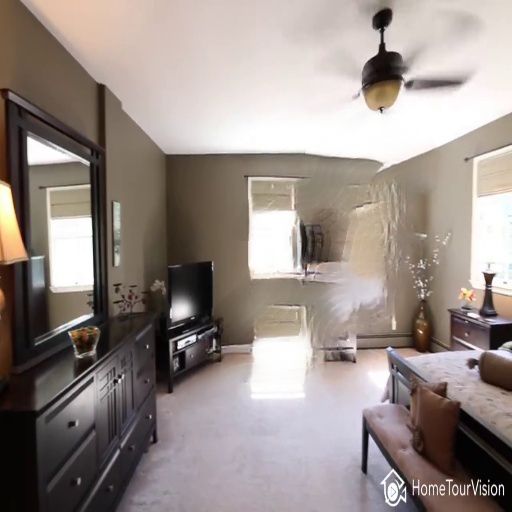}
    \end{subfigure}
    \begin{subfigure}{0.16\linewidth}
        \includegraphics[width=1\linewidth]{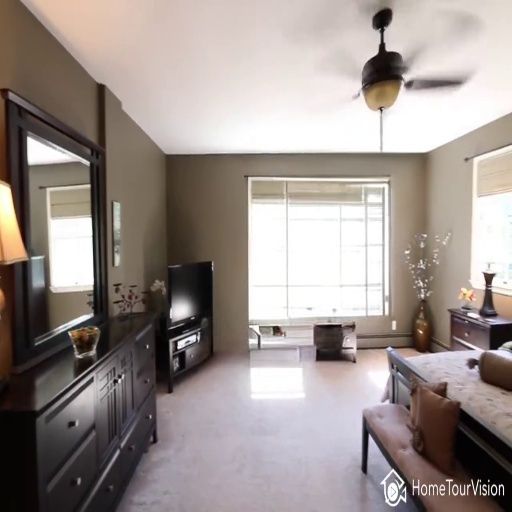}
    \end{subfigure}
    \begin{subfigure}{0.16\linewidth}
        \includegraphics[width=1\linewidth]{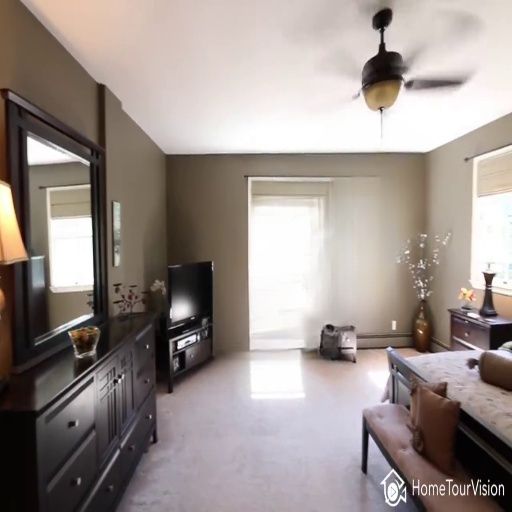}
    \end{subfigure}
    \begin{subfigure}{0.16\linewidth}
        \includegraphics[width=1\linewidth]{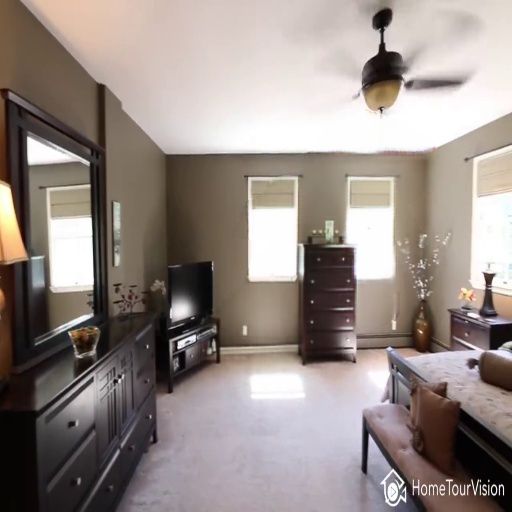}
    \end{subfigure}

    \centering
    \begin{subfigure}{0.16\linewidth}
        \includegraphics[width=1\linewidth]{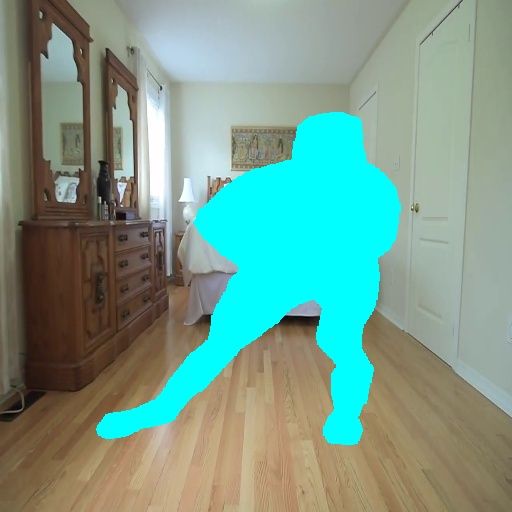}
        \caption{Target}
    \end{subfigure}
    \begin{subfigure}{0.16\linewidth}
        \includegraphics[width=1\linewidth]{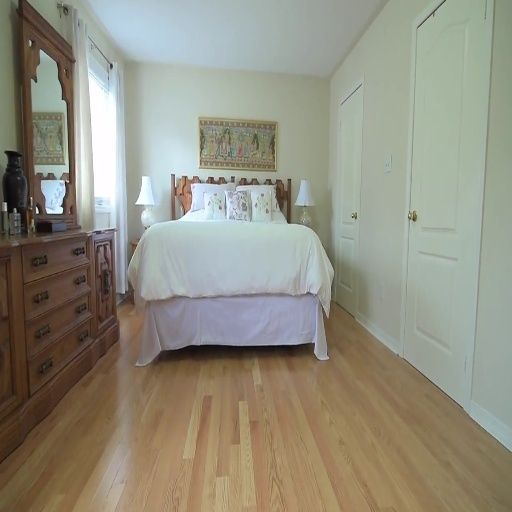}
        \caption{Reference}
    \end{subfigure}
    \hfill
    \begin{subfigure}{0.16\linewidth}
        \includegraphics[width=1\linewidth]{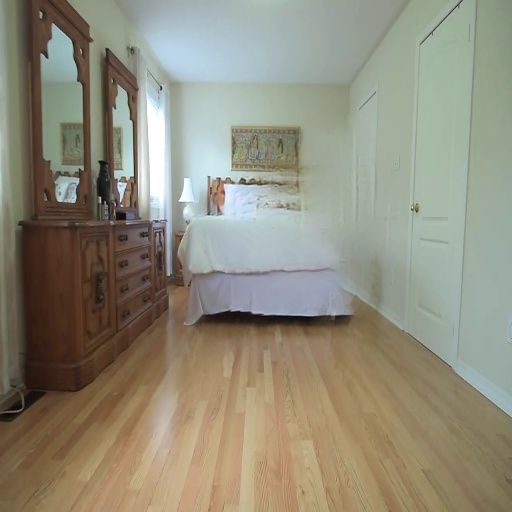}
        \caption{OPN}
    \end{subfigure}
    \begin{subfigure}{0.16\linewidth}
        \includegraphics[width=1\linewidth]{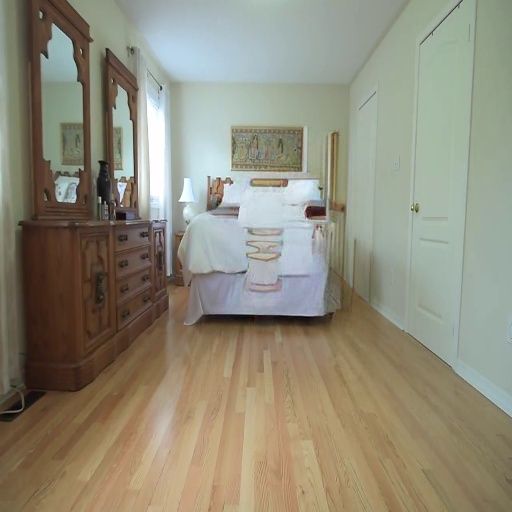}
        \caption{MAT}
    \end{subfigure}
    \begin{subfigure}{0.16\linewidth}
        \includegraphics[width=1\linewidth]{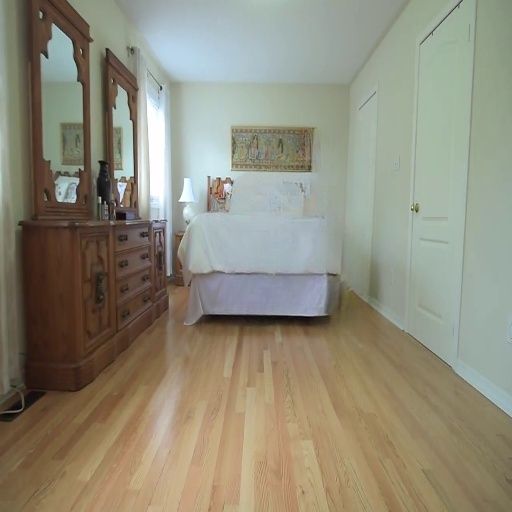}
        \caption{LaMa}
    \end{subfigure}
    \begin{subfigure}{0.16\linewidth}
        \includegraphics[width=1\linewidth]{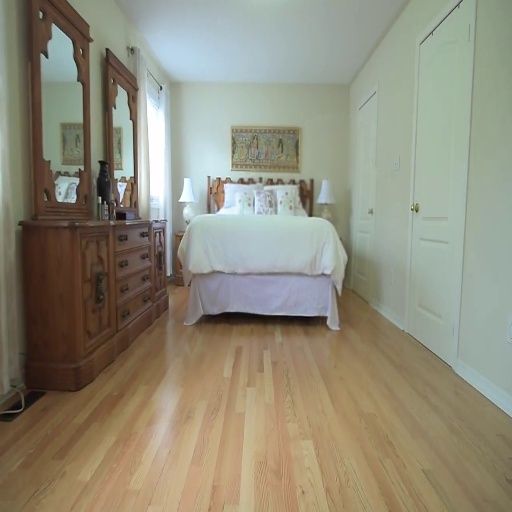}
        \caption{3DFill}
    \end{subfigure}
    
    \caption{Additional qualitative comparisons. For better visualization, please zoom in to see the details.}
    \label{fig:additional qualitative comparison}
\end{figure*}

The alignment results of 3DFill are shown in \cref{fig:alignment results}.
The additional qualitative comparisons are shown in \cref{fig:additional qualitative comparison}.

\section{Failure Cases}
\label{sec:fail}
\begin{figure*}[htbp]
    \centering
    \begin{subfigure}{0.19\linewidth}
        \includegraphics[width=1\linewidth]{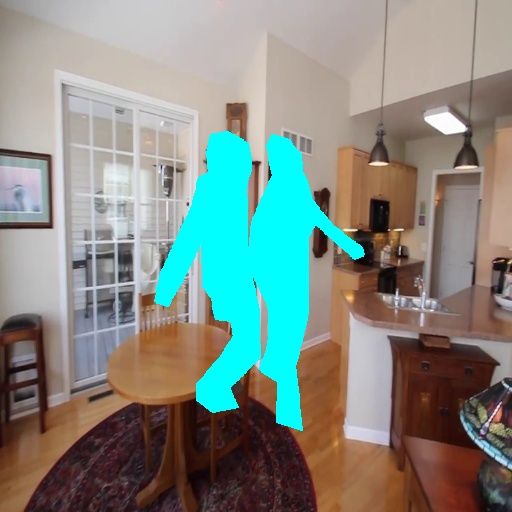}
        \caption{Target}
    \end{subfigure}
    \begin{subfigure}{0.19\linewidth}
        \includegraphics[width=1\linewidth]{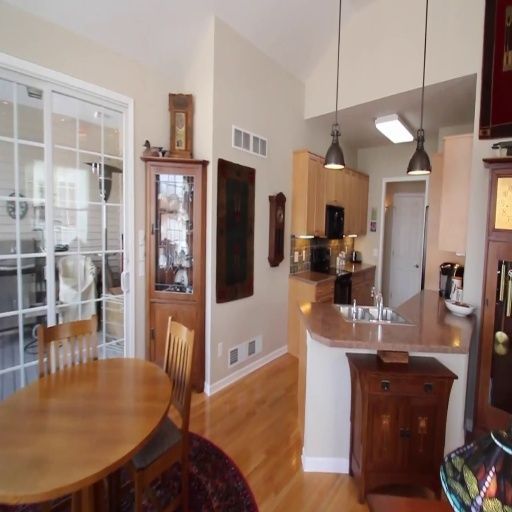}
        \caption{Reference}
    \end{subfigure}
    \begin{subfigure}{0.19\linewidth}
        \includegraphics[width=1\linewidth]{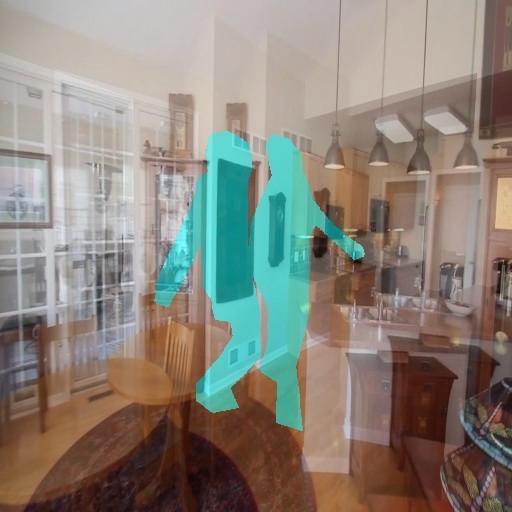}
        \caption{Before alignment}
    \end{subfigure}
    \begin{subfigure}{0.19\linewidth}
        \includegraphics[width=1\linewidth]{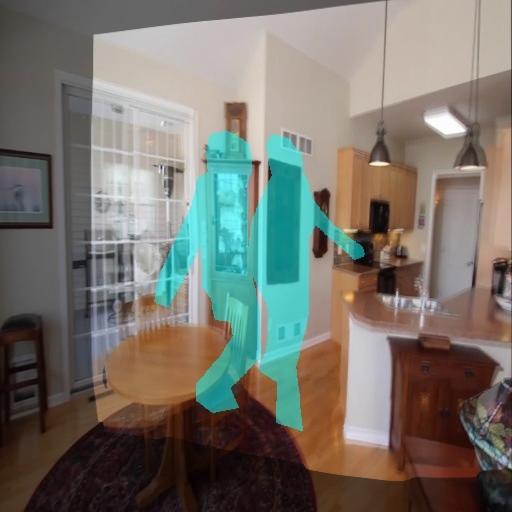}
        \caption{After alignment}
    \end{subfigure}
    \begin{subfigure}{0.19\linewidth}
        \includegraphics[width=1\linewidth]{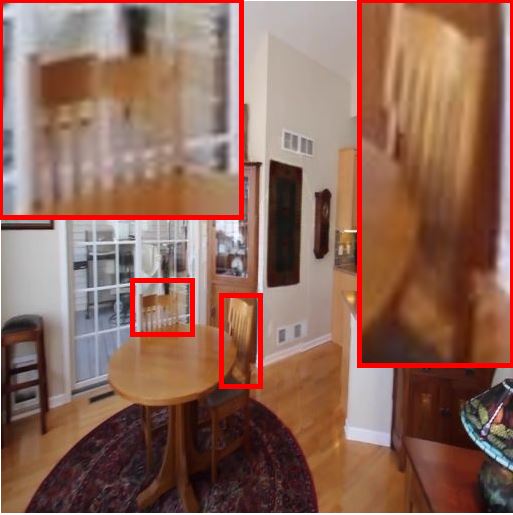}
        \caption{Result}
    \end{subfigure}   

    \caption{Failure case. In this case, the viewpoint of edge region (such as the tables and chairs at the bottom left of the image) has changed too much due to perspective and lens distortion, resulting in poor alignment.}
    \label{fig:failure cases}
\end{figure*}

Most failure cases of 3DFill are caused by holes that contain complex scenes at the edge of the image.
It is difficult to align the edge of the image as well as the center of the image due to perspective, errors in depth estimation and lens distortion, especially when some complex objects appears at the edge of the image.
Under this condition, although the overall structure in the hole regions can be restored, structural dislocation may occur at the edge of some hole regions as shown in \cref{fig:failure cases}.

The more realistic camera intrinsic parameters considering lens distortion parameters and the more accurate depth estimation model can alleviate the above problems.

\end{document}